\documentclass[preprint,12pt,authoryear]{elsarticle}

\usepackage{amssymb}
\usepackage{amsmath}
\usepackage[inline]{enumitem}
\usepackage{bbm}
\usepackage{booktabs}
\usepackage{multirow}
\usepackage{graphicx}
\usepackage{subcaption}
\usepackage[margin=1in]{geometry}
\usepackage{caption}
\usepackage{makecell}
\usepackage{placeins}

\journal{Journal of Choice Modelling}

\begin{document}
\sloppy

\begin{frontmatter}



\title{Functional effects models: Accounting for preference heterogeneity in panel data with machine learning} 


\author[inst1]{Nicolas Salvadé}
\ead{nicolas.salvade.22@ucl.ac.uk}
\affiliation[inst1]{organization={Department of Civil, Environmental and Geomatic Engineering, University College London},
           addressline={Gower Street}, 
            city={London},
            postcode={WC1E 6BT}, 
            country={United Kingdom}}

\author[inst1]{Tim Hillel\corref{cor1}}
\cortext[cor1]{Corresponding author}
\ead{tim.hillel@ucl.ac.uk}

\begin{abstract}

In this paper, we present a general specification for Functional Effects Models, which use Machine Learning (ML) methodologies to learn individual-specific preference parameters from socio-demographic characteristics, therefore accounting for inter-individual heterogeneity in panel choice data. This approach exploits the generalisation power of gradient-based ML regression techniques to account for inter-individual heterogeneity in sequential choices. We identify three specific advantages of the Functional Effects Model over traditional fixed, and random\slash mixed effects models:
\begin{enumerate*}[label=(\roman*)] 
    \item by mapping individual-specific effects as a function of socio-demographic variables, we can account for these effects when forecasting choices of previously unobserved individuals 
    \item the (approximate) maximum-likelihood estimation of functional effects avoids the incidental parameters problem of the fixed effects model, even when the number of observed choices per individual is small; and 
    \item we do not rely on the strong distributional assumptions of the random effects model, which may not match reality. 
\end{enumerate*}    
We learn functional intercept and functional slopes with powerful non-linear machine learning regressors for tabular data, namely gradient boosting decision trees and  deep neural networks.
We validate our proposed methodology on a synthetic experiment and three real-world panel case studies, demonstrating that the Functional Effects Model:
\begin{enumerate*}[label=(\roman*)] 
    \item can identify the true values of individual-specific effects when the data generation process is known;
    \item outperforms both state-of-the-art ML choice modelling techniques that omit individual heterogeneity in terms of predictive performance, as well as traditional static panel choice models in terms of learning inter-individual heterogeneity.
\end{enumerate*}
The results indicate that the FI-RUMBoost model, which combines the individual-specific constants of the Functional Effects Model with the complex, non-linear utilities of RUMBoost, performs marginally best on large-scale revealed preference panel data.
\end{abstract}


\begin{highlights}
\item Introduces the functional effects model, using Machine Learning methodologies to account for inter-individual heterogeneity.
\item Functional Effects Models learn individual-specific parameters\slash preferences from socio-demographic characteristics.
\item Verifies that true functional effects can be recovered on a synthetic experiment.
\item Provides a thorough comparison between Functional Effects Models and machine learning and statistical models.
\item Case study on the easySHARE, Swissmetro, and LPMC datasets.

\end{highlights}

\begin{keyword}
Panel Data \sep Machine Learning \sep Choice Modelling  \sep Mode Choice \sep Ordinal Regression


\end{keyword}

\end{frontmatter}


\section{Introduction}

Human choices are immensely complex and are inherently linked to observed variables, such as the cost of an alternative, and unobserved variables, such as life experiences and preferences. Consecutive choices made by the same individual share these unobserved variables and will be inevitably correlated. This violates the assumption of independence between observations, a common assumption in Machine Learning (ML) models and basic statistical models such as the Multinomial Logit (MNL) model. Therefore, these types of choice experiments, referred to as \textit{panel} or \textit{longitudinal} studies, require specific methodologies to address the correlation between observations. Common problems dealing with panel data include, for example, longitudinal health studies, household income variation over time studies, stated preference experiments where respondents are being asked to make several choices consecutively, or even time series analysis. 

Traditionally, practitioners model preference parameters with two types of choice models based on the random utility theory: 
\begin{enumerate*}[label=(\roman*)]
    \item static models; and
    \item dynamic models.
\end{enumerate*}
For a detailed discussion, see \citet{10.1093/oxfordhb/9780199940042.013.0006}.
These models differ in how they account for the serial correlation between the error terms of the same individual. The first type models preference parameters with either fixed effects, i.e., individual-specific constants (or intercepts), estimated from maximum likelihood of observed choices, or random, or mixed, effects, i.e., random parameters distributed across the population\footnote{Note, that the mixed effects model refers to each parameter being the combination of a fixed "population-level" parameter with an individual-specific random intercept (i.e., $\beta_m + u_{mn}$ where $u_{mn}$ is a random intercept), whereas the fixed effects model refers to individual-specific preference parameters (i.e., $\alpha_{in}$ which is learnt for each individual using maximum likelihood estimation). While mixed effects is a general term, in the context of panel data we distinguish between the random intercepts and random slopes models. }. However, the fixed effects model suffer from the incidental parameters problem, where the parameters are inconsistent when the number of observations per individual is small, and is typically only used in regression\slash ordinal tasks \citep{10.1093/oxfordhb/9780199940042.013.0006}. Therefore, in this work, we focus on the mixed effects model. Two special cases of mixed effects models are the random intercept model, with the intercept, or constant, assumed to be a random variable, and the random slopes model, where the coefficients related to a variable are assumed to be randomly distributed. This method has been first developed to identify the true values of estimated parameters, removing bias from not accounting for individual heterogeneity. However, in recent years, with the ever-increasing use of ML models, predictive tasks have become predominantly important, and the mixed effects model is limited when predicting preferences of unknown individuals. More specifically, the random effects fitted during training have to be averaged (i.e., use the population-level mean with the mean of the random effects distribution) \citep{krueger2021evaluating}, reducing predictive power. In addition, the random effects model rely on strong distributional assumptions that need to be specified by the modeller.

The second type of models is Markov chain models, where the last observed choice made by an individual is used to account for panel effects. When handling the first observed choice of an individual, dynamic models can be thought of as static models, since there are no previous choices to use as exogenous variables. This is known in the literature as the \textit{initial conditions problem} of dynamic models \citep{train2009discrete}. Therefore, dynamic models are also not suitable to predict preferences of unknown individuals, and there is a crucial need to develop specific methodologies to make static models suitable for forecasting. Note that we emphasise the parallels between dynamic models and recommender systems, especially in the \textit{cold start problem} \citep{panda2022approaches}, where static models could be used to generate recommendations for individuals without prior choice knowledge. Hereafter, since we are interested in panel data predictions for unobserved individuals, we focus on static models for panel data. Table \ref{glossary} provides a glossary of all the terminology used in this paper.

\begin{table}[htbp] 
\centering
\footnotesize
\caption{Glossary of inputs, model types, and parameters.}\label{glossary}
\begin{tabular}{p{0.3\linewidth} p{0.65\linewidth}}
\toprule
\textbf{Term} & \textbf{Definition} \\
\midrule
\multicolumn{2}{l}{\textit{1. Inputs / Data}} \\
Panel / longitudinal data & Data containing repeated choices from individuals. \\
Target variable ($y_i$) & The variable to be modelled or predicted. \\
Explanatory variables\slash features ($x_{imnt}$) & Observed variables indexed by $i$ (alternative), $m$ (variable), $n$ (individual), and $t$ (time). \\
Socio-demographic characteristics ($s_n$) & Characteristics of individual $n$, such as demographic or economic background. \\ \\

\multicolumn{2}{l}{\textit{2. Model Types}} \\
Static models & Models accounting preferences without prior choices knowledge. \\
Dynamic models & Models including prior choices as variables to model preferences. \\
Fixed effects model & Model preferences with individual-specific intercepts. \\
Mixed / random effects model & Model preferences with random variables drawn from a distribution. \\
Random intercept model & A mixed effects model where only the intercept varies randomly across individuals. \\
Random slope model & A mixed effects model where slopes of features vary randomly across individuals. \\
Random intercept and slope model & A mixed effects model with both random intercepts and random slopes. \\
Functional effect model & Model where preferences are learnt from $s_n$. \\
Functional intercept model with linear coefficients & Functional intercept depending on $s_n$ while slopes remain constant (linear). \\
Functional intercept model with non-linear coefficients & Functional intercept depending on $s_n$ while slopes are a non-linear function of the features. \\
Functional slopes model & Functional slopes depending on $s_n$ while intercepts remain constant. \\
Functional intercept and slopes model & Both intercept and slopes depend on $s_n$. \\ \\
\multicolumn{2}{l}{\textit{3. Parameters}} \\
Intercepts ($\alpha_{in}$) & Value at which a function intersects the y-axis. \\
Linear coefficients ($\beta_{im}$) & Constant marginal effects of features. \\
Non-linear coefficients ($f_{im}(x_{imnt})$) & Non-linear functions of features learnt with a gradient-based ML regressor. \\
Functional intercept ($g_{i,0}(s_n)$) & Intercept learnt from $s_n$ with a gradient-based ML regressor. \\
Functional slopes ($g_{im}(s_n)$) & Slope of variable $m$ learnt from $s_n$ with a gradient-based ML regressor. \\
\bottomrule
\end{tabular}
\end{table}

There have been several attempts to adapt static models, such as the mixed effects model, with Machine Learning (ML) methodologies. The main idea is to learn the population-level mean with an out-of-the-box ML regressor and estimate random effects as in a linear mixed effects model. For example, this has been done for regression trees \citep{hajjem2017generalized, sela2012re}, linear trees \citep{fokkema2018detecting}, Random Forests (RF) \citep{hajjem2014mixed}, Gradient Boosting Decision Trees (GBDT) \citep{9759834}, Neural Networks (NN) \citep{mandel2023neural}, and Convolutional Neural Networks (CNN) \citep{xiong2019mixed}. All these papers use some sort of Expectation-Maximisation (EM) algorithm to iteratively estimate the population-level mean and random effects. This framework has also been generalised by \cite{ngufor2019mixed} and \cite{kilian2023mixed}. However, these models face two main problems when applied in choice modelling with panel data:
\begin{enumerate*}[label=(\roman*)]
    \item the random effects are not suited for predictions, which means that they need to be averaged, dropped, or re-estimated at inference time at the cost of an expensive computational procedure; and
    \item they mostly rely on out-of-the-box ML models, which are not interpretable.
\end{enumerate*}
Note that in \cite{xiong2019mixed}, the authors decompose the output of the model as fixed and random effects, where both effects depend on input features. However, by doing so, the random effects of their model are \textit{effectively} fixed effects and do not have any random component, falling back to an out-of-the-box CNN. Finally, it is worth mentioning that there is very little existing research into dynamic ML models in a choice modelling context. 

In this paper, we use existing gradient-based ML methodologies to learn individual-specific parameters as a function of socio-demographic characteristics. We learn two types of functional effects: 
\begin{enumerate*}[label=(\roman*)]
    \item functional intercept, effectively emulating the random intercept model; and
    \item functional slopes, effectively emulating the random slopes model.
\end{enumerate*}
At a high level, we are using unrestricted ML regressors to impute an individual-specific constant from the socio-demographic characteristics. This constant mimics the random effect in traditional mixed effects models, but since it is a function of the socio-demographic characteristics, it can be easily used for inference. We do so with the two most popular ML regressors for tabular data, Gradient Boosting Decision Trees (GBDTs) and Deep Neural Networks (DNNs). It is worth noting that, whilst the primary contribution of the paper is a general framework for capturing individual effects for panel data, this framework incorporates and builds on several existing works in the literature, specifically: L-MNL \citep{sifringer2020enhancing}, TasteNet-MNL \citep{han2022neural}, and RUMBoost \citep{salvade2025rumboost}. We further note that all of the models presented in this paper could similarly be used to account for individual heterogeneity in cross-sectional data, though this is not the focus of this paper. We systematically evaluate our methodology on a synthetic experiment and three real-world panel case studies.

The main strengths of our proposed methodology are:
\begin{enumerate}
    \item incorporates individual-specific preferences in powerful ML models, resulting in improved real-world predictive performance compared to existing state-of-the-art ML-based choice models that assume homogenous preferences;
    \item accounts for individual-specific effects when forecasting choices of previously unobserved individuals, which is essential for counterfactual analysis; and
    \item compared to traditional static choice models, avoids the incidental parameters problem of the fixed effects model and the strong distributional assumptions of the random effects models.
\end{enumerate}

All the code used in this paper, including the implementation of a generalised functional effects model learnt with GBDTs and DNNs for different datasets and choice situations, is open source and freely available on Github\footnote{https://github.com/big-ucl/functional-effects-model}.

The rest of the paper is structured as follows: Section \ref{theoretical background} provides the theoretical background, and Section \ref{methodology} introduces the methods used in the paper. We validate the methodology with a synthetic experiment in Section \ref{synthetic} and provide a thorough benchmark with three real-world case studies in Section \ref{case studies}. Finally, Section \ref{conclusion} concludes the paper.

\section{Theoretical background} \label{theoretical background}

\subsection{Choice models with homogenous preferences} \label{multiclass}

Choice models based on the random utility theory assume that choice makers are rational and will choose the alternative that maximises their utility, a latent representation of their preferences. The utility function is typically composed of a deterministic part and a random part, i.e.:

\begin{equation}\label{utility}
    U_{in} = V_{in} + \epsilon_{in}
\end{equation}
where:
\begin{itemize}
    \item $U_{in}$ is the utility function for an individual $n$, and alternative $i$;
    \item $V_{in}$ is the deterministic utility for an individual $n$, and alternative $i$; and
    \item $\epsilon_{in}$ is the error term, capturing unobserved informations.
\end{itemize}

The deterministic utility is widely represented as a \textit{linear-in-parameter} function of the variables, i.e.;
\begin{equation}\label{deterministic utility}
    V_{in} = \alpha_{i} + \sum_{m=1}^{M_i}\beta_{im}x_{inm}
\end{equation}
where:
\begin{itemize}
    \item $M_i$ is the number of variables for alternative $i$;
    \item $x_{inm}$ is the variable $m$ for alternative $i$ and observation $n$;
    \item $\beta_{im}$ are homogenous parameters to be estimated from the data $\forall i, m$; and
    \item $\alpha_i$ is the intercept, or constant, for alternative $i$.
\end{itemize}

Interpretable non-linear ML utility models (e.g. RUMBoost \citep{salvade2025rumboost}) extends the deterministic utility function specification of Eq. \ref{deterministic utility} by replacing $\beta_{im}x_{inm}$ with the output of a gradient-based ML regressor:
\begin{equation}\label{ml utility}
    V_{in} = \alpha_{i} + \sum_{m=1}^{M_i}f_{im}(x_{inm})
\end{equation}
where:
\begin{itemize}
    \item $f_{im}$ is a non-parametric non-linear function learnt from a gradient-based ML regressor $\forall i, m$
\end{itemize}

Note that some researchers have attempted to interpret the output of an ML regressor as the deterministic utility:
\begin{equation}\label{out-of-the-box ml utility}
    V_{in} = f_i(\mathbf{x}_{in})
\end{equation}
where:
\begin{itemize}
    \item $f_{i}$ is a non-parametric non-linear function learnt from a gradient-based ML regressor $\forall i$, where all variables can interact without restrictions.
\end{itemize}
However, $f_i$ is usually not consistent with the random utility theory and is not interpretable.

In this work, we assume the error term to be \textit{independent and identically distributed} (\textit{i.i.d.}), such that the probability $P_{in}$ of an individual $n$ to choose alternative $i$ with $J$ alternatives can be derived as in an MNL model, that is:

\begin{equation}
    P_{in} = \frac{e^{V_{in}}}{\sum_{j=1}^{J} e^{V_{jn}}}
\end{equation}

The parameters are chosen to minimise the negative Cross-Entropy Loss (CEL) function, akin to Maximum Log-Likelihood Estimation (MLE):

\begin{equation}
    \mathcal{L} = \sum_{n=1}^{N}\sum_{j=1}^{J-1} \mathbbm{1}(j=y_n) \ln(P_{jn})
\end{equation}
where:
\begin{itemize}
    \item $\mathbbm{1}(j=y_n)$ is $1$ if $j$ equals to the observed chosen alternative $y_n$, $0$ otherwise.
\end{itemize}

\subsection{Static models for panel data} \label{static models}

Panel data have more than one observation per individual. Therefore, we extend Equation \ref{utility} as follows:

\begin{equation}\label{panel utility}
    U_{int} = V_{int} + \epsilon_{int}
\end{equation}
where $t$ represents the $t^{th}$ choice of an individual $n$. Inevitably, the error term $\epsilon_{int}$ is not \textit{i.i.d.} across $t$ and specific methodologies are required to deal with this intrinsic correlation, i.e., dynamic models. However, we omit the dynamic correlations here, as we do not address these in the current methodology, instead leaving this to further work. In order to reduce the parameter bias by accounting for inter-individual heterogeneity, the fixed effects model extends Eq. \ref{deterministic utility} as follows:
\begin{equation}\label{fixed effects utility}
    V_{int} = \alpha_{in} + \sum_{m=1}^{M_i}\beta_{inm}x_{inmt}
\end{equation}
where the model parameters $\alpha_{in}$ and $\beta_{inm}$ are now individual-specific. A fixed intercept model is a model with only $\alpha_{in}$ being individual-specific, and a fixed slopes model is a model with only the slopes, or coefficients, $\beta_{inm}$ being individual-specific.

On the other hand, the mixed\slash random effects model incorporates inter-individual heterogeneity with random variables, i.e.:
\begin{equation}\label{random effects utility}
    V_{int} = \alpha_{i} +u_{in0} + \sum_{m=1}^{M_i}(\beta_{im} + u_{inm})x_{inmt}
\end{equation}
where $\alpha_{i}$ and $\beta_{im}$ are population-level parameters and $u_{inm}$ is a random variable (typically normally distributed) with 0 mean and standard deviation to be estimated from the data. A mixed\slash random intercept model is a model with only $u_{in0}$ being randomly distributed, and a mixed\slash random slopes model is a model with only the slopes, or coefficients, $u_{inm}$ being individual-specific.

\section{Methodology} \label{methodology}

\subsection{Functional effects models} \label{functional effect model metho}
In this study, we assume that the correlation of the error term across $t$ can be captured by individual-specific parameters learnt from the socio-demographic characteristics, such that the error term $\epsilon_{int}$ can be assumed to be \textit{i.i.d.} and the methodology described in Section \ref{multiclass} is still applicable.

More formally, given a set of $M$ variables $\mathbf{x}_{int} \in \mathbb{R}^{M}$ and a set of $Q$ socio-demographic characteristics $\mathbf{s}_n \in \mathbb{R}^Q$ for individual $n$ and alternative $i$, we define the deterministic utility function as follows:

\begin{equation} \label{functional effects utility}
    V_{int} =g_{i0}(\mathbf{s}_{n}) +\sum_{m=1}^{M_i}g_{im}(\mathbf{s}_n)x_{intm} 
\end{equation}

where:
\begin{itemize}
    \item $g_{im}(\mathbf{s}_n)$ is the output of a gradient-based ML regressor using the socio-demographic characteristics as input data for all $M_i$ variables, $J$ classes, and individuals $n$ that does not depend on the choice dimension $t$.
    \item $g_{i0}(\mathbf{s}_n)$ is the output of an ML regressor using the socio-demographic characteristics as input data for the intercept for alternative $i$ and individual $n$.
\end{itemize}

\begin{figure}
    \centering
    \begin{subfigure}[b]{0.47\textwidth}
        \includegraphics[width=\textwidth]{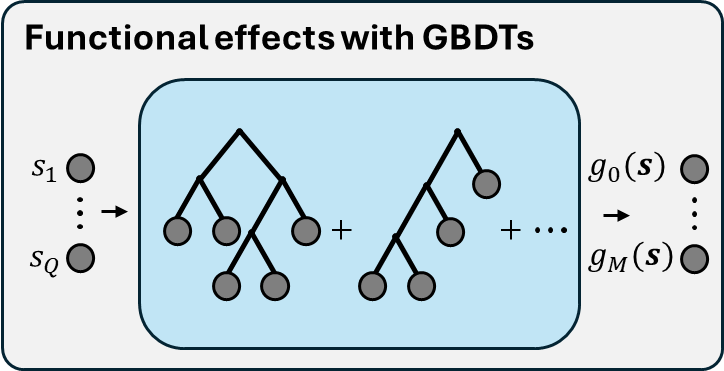}
        \subcaption{Functional effects learnt with GBDTs.}
    \end{subfigure}
    \begin{subfigure}[b]{0.468\textwidth}
        \includegraphics[width=\textwidth]{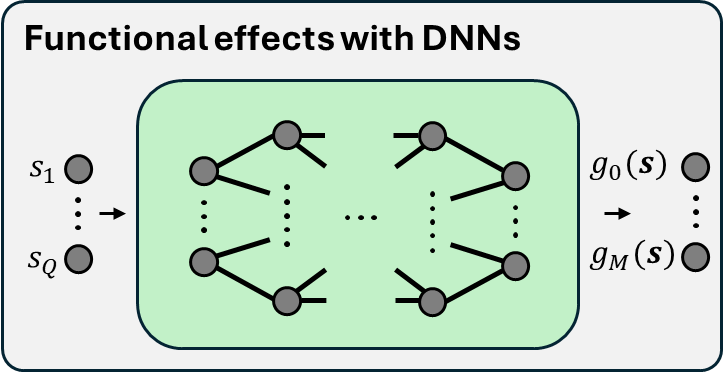}
        \subcaption{Functional effects learnt with DNNs.}
    \end{subfigure}
    \begin{subfigure}[c]{.944\textwidth}
        \includegraphics[width=\textwidth]{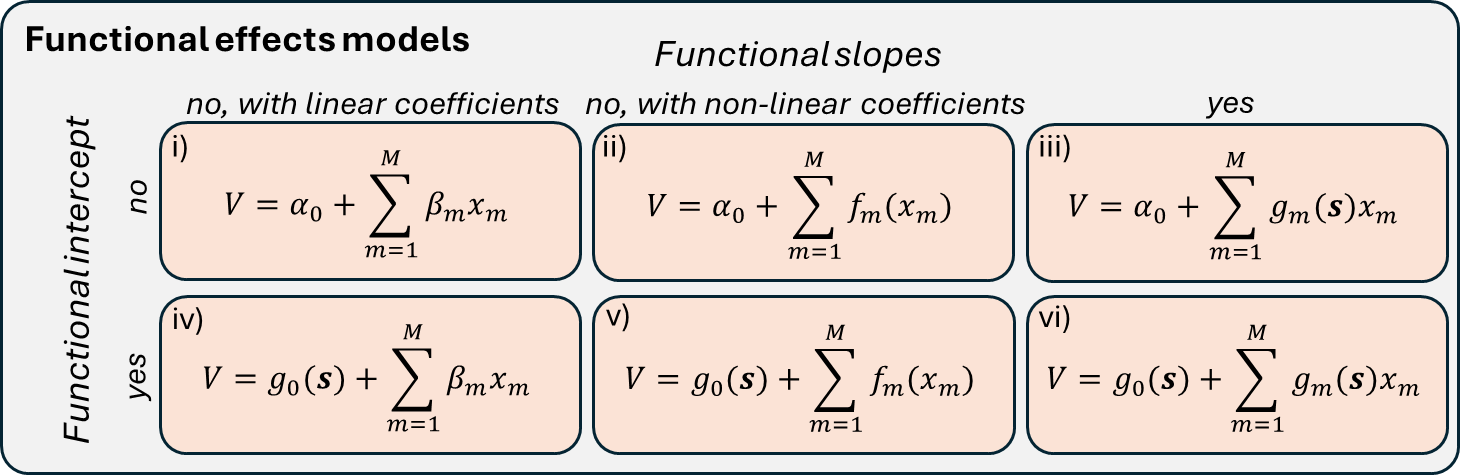}
        \subcaption{Types of functional effects models. They can be with or without functional intercept, slopes, and with linear or non-linear coefficients. Model i) is equivalent to an MNL.}
    \end{subfigure}
    \caption{Overview of the methodology.}
    \label{Methodology overview}
\end{figure}

We define more precisely the \textit{Functional Intercept} (FI) model, where only $g_{i0}(\mathbf{s}_n)$ is the output of a gradient-based ML regressor, and the \textit{Functional Slopes} (FS) model, where only $g_{im}(\mathbf{s}_n), \forall m, i$ are the outputs of a gradient-based ML regressor. Finally, the \textit{Functional Intercept and Slopes} (FIS) model combines both functional intercept and slopes. If not learnt from the socio-demographic characteristics, the parameters can be either estimated linearly as in Eq. \ref{deterministic utility} or non-linearly imputed from a gradient-based ML regressor as in \ref{ml utility}. Figure \ref{Methodology overview} provides an overview of all possible \textit{functional effects} models. We note that keeping the same \textit{linear-in-parameter} utility function as in a traditional MNL model allows for the functional effects model to maintain significant interpretability.

\subsection{ML Regressors}

The functional intercept or slopes are the output of a gradient-based ML regressor. Note that this approach requires the models to be fit to the gradients of the loss function with respect to the individual-specific parameters, and so can be adapted for any gradient-based ML algorithm. 

In this paper, we present a comparison between the two most popular ML regressors for tabular data: GBDTs and DNNs.

\subsubsection{Functional effects with GBDTs}

The first ML regressor used in this paper is GBDT \citep{friedman2001greedy, chen2016xgboost}, which uses ensembles of regression trees to learn functional effects. \citet{salvade2025rumboost} extend this concept to a parametric utility context with RUMBoost, where, for a model with $R$ iterations, the GBDT predictive function is an additive function of the form:

\begin{equation}
    f_{im}(x_{inm}) = \sum_{r=1}^{R} w_{inmr}
\end{equation}
where $w_{inmr}$ is the leaf value (a constant) of a tree $r$, where the individual $n$ has been partitioned to. At each iteration, the leaf values of a new tree are chosen to directly minimise the second-order Taylor expansion of the loss function, such that:

\begin{equation} \label{leaf value}
    w_{inmr} =  - \frac{(\sum_{n \in l}g_{inm})}{(\sum_{n \in l}h_{inm})}
\end{equation}

where:
\begin{itemize}
    \item $g_{inm} = \partial\mathcal{L}/\partial V_{in}$ the first derivative of the loss function with respect to $V_{in}$; 
    \item and $h_{inm} = \partial^2\mathcal{L}/\partial^2V_{in}$ is the second derivative of the loss function with respect to $V_{in}$; and
    \item $l$ is the subset of observations that has been partitioned to the corresponding leaf value.
\end{itemize}  

In this paper, we extend this concept to functional effects being a function of socio-demographic characteristics $\mathbf{s}_n$. We have:
\begin{equation}
    g_{im}(\mathbf{s}_{n}) = \sum_{r=1}^{R} w_{inmr}
\end{equation}
where the leaf values are computed as in Eq. \ref{leaf value}, but in the \textit{parameter space}, that is $g_{inm} = \partial\mathcal{L}/\partial\beta_{inm} = \partial\mathcal{L}/\partial V_{in} \cdot \partial V_{in}/\partial\beta_{inm} $ the first derivative of the loss function with respect to $\beta_{inm}$ and $h_{inm} = \partial^2\mathcal{L}/\partial^2\beta_{inm} = \partial^2\mathcal{L}/\partial^2 V_{in} \cdot (\partial V_{in}/\partial\beta_{inm})^2$ is the second derivative of the loss function with respect to $\beta_{inm}$. Since the utility function is linear-in-parameter, the first-order derivatives are scaled by $x_{inmt}$ and the second-order derivatives are scaled by $x_{inmt}^2$ compared to a RUMBoost model.

Note that, for the FI model, we combine RUMBoost to find generic non-linear coefficients (i.e., $f_{im}(x_{inm})$) with individual-specific intercept (i.e., $g_{i0}(\mathbf{s}_n)$) to capture preferences.

\subsubsection{Functional effects with DNNs}
The second ML regressor that we consider to learn functional effects is \textit{feed-forward} DNNs \citep{goodfellow2016deep}. More formally, the feed-forward DNN predictive function takes the following form:

\begin{equation} \label{DNN}
    g_{im}(\mathbf{s}_n) = h_O \circ \cdots \circ h_1(\mathbf{s}_n)
\end{equation}
where:
\begin{itemize}
    \item $h_o = a(w_ox + b_o)\quad \forall o = 1, \cdots, O$;
    \item $w_o$ and $b_o$ are the weight and biases of the hidden layer $o$;
    \item $a(x)$ is an activation function (e.g., ReLU, sigmoid, leaky ReLU, tanh, softplus, ...); and
    \item $O$ is the number of hidden layers in the DNN.
\end{itemize}

For training stability, it is common to split the datasets into batches of data. After each batch, the parameters of the models are updated with some variant of the gradient descent with first-order derivatives of the loss function (see e.g. \citet{kingma2014adam}), with the gradients being efficiently computed through back-propagation (an algorithm efficiently applying the chain rule). 

Note that the FI model falls back to the L-MNL proposed in \citet{sifringer2020enhancing} and the FIS model falls back to the TasteNet-MNL proposed in \citet{han2022neural}. We refer the reader to the original papers for complete explanations of the underlying methodology.

\subsubsection{Monotonicity constraints} \label{monotonic constraints}

For both models, we can constrain the sign of the functional effects with a Rectified Linear Unit (ReLU) activation function:

\begin{equation}
    g_{im}^{c}(\mathbf{s}_n) = c \cdot \text{ReLU}(c g_{im}(\mathbf{s}_n)) = c \cdot \text{max}(0, c g_{im}(\mathbf{s}_n))
\end{equation}

where $c = 1$ for a positive monotonic constraint and $c=-1$ for a negative monotonic constraint.

\subsection{Benchmarked models and relationship with prior work} \label{model benchamrked}
Table \ref{functional effects models definition} enumerates all 11 potential functional effects models, with the corresponding model numbers from Figure \ref{Methodology overview}. 

\begin{table}[htbp]
\centering
\caption{All potential functional effects model. The model number is linked to the one from Figure \ref{Methodology overview}. FI stands for functional intercept, FS stands for functional slopes, and FIS stands for functional intercept and slopes. FI-DNN is equivalent to the L-MNL model, and FIS-DNN is equivalent to the TasteNet-MNL model.}
\label{functional effects models definition}
\begin{tabular}{rllll} \toprule
 &  &       & \multicolumn{2}{c}{\textbf{Model names}} \\
   \textbf{Model}                    &    \textbf{Intercept}                        &    \textbf{Slopes}                          & \multicolumn{1}{c}{\textbf{GBDT-based}}           & \multicolumn{1}{c}{\textbf{DNN-based}}           \\ \midrule
i)                     & $\alpha_0$    & $\beta_m x_m$ & \multicolumn{2}{c}{(MNL\slash Ordinal Logit)}         \\
ii)                    & $\alpha_0$    & $f_m(x_m)$                  & \multicolumn{1}{r}{RUMBoost}        & \multicolumn{1}{r}{N/A}           \\
iii)                   & $g_0(\textbf{s})$                & $\beta_m x_m$ & \multicolumn{1}{r}{N/A}             & \multicolumn{1}{r}{FI-DNN}        \\
iv)                    & $g_0(\textbf{s})$                    & $f_m(x_m)$                   & \multicolumn{1}{r}{FI-RUMBoost}         & \multicolumn{1}{r}{N/A}           \\
v)                     & $\alpha_0$    & $g_m(\textbf{s}) x_m$                 & \multicolumn{1}{r}{FS-GBDT}         & \multicolumn{1}{r}{FS-DNN}        \\
vi)                    & $g_0(\textbf{s})$                    & $g_m(\textbf{s}) x_m$                 & \multicolumn{1}{r}{FIS-GBDT}      & \multicolumn{1}{r}{FIS-DNN}    \\ \bottomrule
\end{tabular}
\end{table}

Note that, whilst this paper presents a unified modelling framework for capturing functional effects in panel data, all models in Table~\ref{functional effects models definition} can also be applied to cross-sectional data. The proposed framework incorporates several existing models in the literature:
\begin{itemize}
    \item the FI-DNN model is equivalent to the L-MNL model proposed by \citet{sifringer2020enhancing}, 
    \item the FIS-DNN model is equivalent to the TasteNet-MNL model proposed by \citet{han2022neural}, and 
    \item the FI-RUMBoost model was first proposed as an extension to the RUMBoost model by \citet{salvade2025rumboost}. 
\end{itemize}
The FS-GBDT, FIS-GBDT, and FS-DNN models are all new to this paper. 

Note we use the names FI-DNN and FIS-DNN as:
\begin{enumerate*}[label=(\roman*)] 
    \item they clearly indicate how each model relates to the others in the framework, and 
    \item these models have been re-implemented within the Functional Effects framework\footnote{https://github.com/big-ucl/functional-effects-model}, and may therefore be different from the original implementation.
\end{enumerate*}
Note further that our naming convention distinguishes between FI-RUMBoost and FS-\slash FIS-GBDT, where the former uses the functional non linear coefficients $f_m(x_m)$ from \citet{salvade2025rumboost} whilst the latter use linear coefficients. Both FI-RUMBoost and FIS-GBDT use a single GBDT ensemble for the functional intercept. 

We further highlight that there are three model combinations in the framework that are not evaluated in this paper. 
To the best of our knowledge, there is no existing methodology that enables the estimation of interpretable non-linear functional coefficients $f_m(x_m)$ with neural networks, hence we not present a DNN equivalent to the RUMBoost and FI-RUMBoost models. Similarly, due to the complexity of estimating generic parameters within gradient boosting, we do not evaluate the FI-GBDT model with linear coefficients. 

\subsection{Overview of experiments}

Table \ref{tab:dataset_overview} provides an overview of the datasets used in the experiments.
For all experiments, we tune the hyperparameters with the Python library Optuna \citep{akiba2019optuna}, using the Tree-structured Parzen Estimator (TPE) algorithm with 100 trials. Table \ref{tab:search space} summarises the hyperparameters tuned and the search space. We keep the hyperparameters that have optimal values on the validation set. Note that for all functional effects models with GBDTs, the learning rate is fixed and calculated as $\min(0.1, 1/\min(\mathbf{M}))$ where $\min(\mathbf{M})$ is the smallest number of variables in a utility function.

\begin{table}[htbp]
\centering
\caption{Overview of the datasets used in the experiments. SP means Stated Preference and RP means Revealed Preference}
\label{tab:dataset_overview}
\begin{tabular}{lrrrr}
\toprule
& \textbf{Synthetic} & \textbf{Swissmetro} & \textbf{LPMC} & \textbf{easySHARE} \\
\midrule
Data type & Synthetic & SP & RP & RP \\
Target type & Multinomial & Multinomial & Multinomial & Ordinal \\
$N_\text{classes}$ & 4 & 3 & 4 & 13 \\
$N_\text{individuals}$ & 10000 & 1192 & 31954 & 130620 \\
$N_\text{observations}$ & 100000 & 10692 & 81086 & 281975 \\
Avg. obs. per individual & 10 & 8.97 & 2.54 & 2.15 \\
\bottomrule
\end{tabular}
\end{table}

\section{Synthetic experiment} \label{synthetic}

The only true way of assessing the behavioural quality of the functional effects models is through a synthetic experiment. Indeed, in a control setting, we can generate known true functional effects and verify the ability of the functional effects models to recover them. We do so for FI-RUMBoost, FI-DNN and a Randnom Intercept model (i.e. a mixed logit model with distributed alternative specific constants) on a fully synthetic dataset of 100000 observations from 10000 individuals. We simulate a multinomial discrete choice problem with 4 alternatives. We generate 4 socio-demographic characteristics ($x_1$-$x_4$) and 4 alternative-specific variables ($x_5$-$x_8$), drawn from a continuous uniform distribution between 0 and 1 such that $x_k \sim \mathcal{U}_{[0, 1]}, \forall k = 1,\cdots ,8$. We draw 10000 unique individuals as independent draws of $x_1$-$x_4$. For each individual, we then draw 10 different choice scenarios, as separate independent draws of $x_5$-$x_8$, repeating the socio-demographic characteristics for each scenario. The complete model specification is the following:

\begin{align}
    V_\text{1} = e^{x_1 + x_2 + x_3 + x_4} - 1 \cdot x_5 \\
    V_\text{2} = (x_1 + x_2 + x_3 + x_4)^2 - 1 \cdot x_6 \\
    V_\text{3} = -\ln{(x_1x_2x_3x_4)} - 1 \cdot x_7 \\
    V_\text{4} =  - 1 \cdot x_8
\end{align}
We assume a Type-1 Extreme Value i.i.d. error term, and so obtain choice probabilities from the logit\slash softmax function. Discrete choices are then sampled from the probabilities using a Monte-Carlo simulation. The functional effects are normalised to 0 in the third utility function, as the functional effects would not be recoverable because of the overspecification of the logit\slash softmax function. We repeat the process to obtain a test set of 20000 observations from 2000 previously unobserved individuals.
Table \ref{tab:synthetic hyp search} of \ref{app:hyperparameter search} summarises the optimal hyperparameter values from the hyperparameter search. We implement the Random Intercept model with Biogeme \citep{bierlaire2023short}, making the assumption that the random intercepts are normally distributed, and using the mean (i.e. without Monte-Carlo Simulation) for forecasting on the test set. We follow \citet{krueger2021evaluating} to simulate the maximum likelihood estimation with 500 draws generated from the Modified Latin Hypercube sampling approach \citep{hess2006use}.

Table~\ref{tab:synthetic_benchmark} shows the Mean Absolute Error (MAE) between recovered and true functional intercepts, the negative Cross-Entropy Loss (CEL) on both train and test set and the computational time to train the models. We observe that FI-DNN marginally outperforms FI-RUMBoost on the MAE and CEL, while FI-RUMBoost is about 4 times faster. The good performance of the two models with functional effects contrasts with the MAE of the Random Intercept model, which is bound to a normal distribution on the train set, and can only use the mean of this distribution on the test set. 
\begin{table}[htbp]
\centering
\small
\caption{MAE between recovered and true functional intercepts, negative cross-entropy loss on train and test set, and computational time. The MAE is averaged over the three class intercepts.}
\label{tab:synthetic_benchmark}
\begin{tabular}{lrrrrr}
\toprule
 & \multicolumn{2}{c}{\textbf{MAE}} & \multicolumn{2}{c}{\textbf{CEL}} &  \\
 & \textbf{Train} & \textbf{Test} & \textbf{Train} &\textbf{Test} & \textbf{Comput. time [s]} \\
\midrule
\textbf{FI-RUMBoost} & 0.044 & 0.043 & 1.343 & 1.344 & 3.100 \\
\textbf{FI-DNN} & 0.038 & 0.037 & 1.346 & 1.343 & 11.820 \\
\textbf{Random Intercept} & 0.289 & 0.112 & 1.35 & 1.349 & 47.640 \\
\bottomrule
\end{tabular}
\end{table}

\begin{figure}[htbp]
    \begin{subfigure}[b]{0.95\textwidth}
        \includegraphics[width=\linewidth]{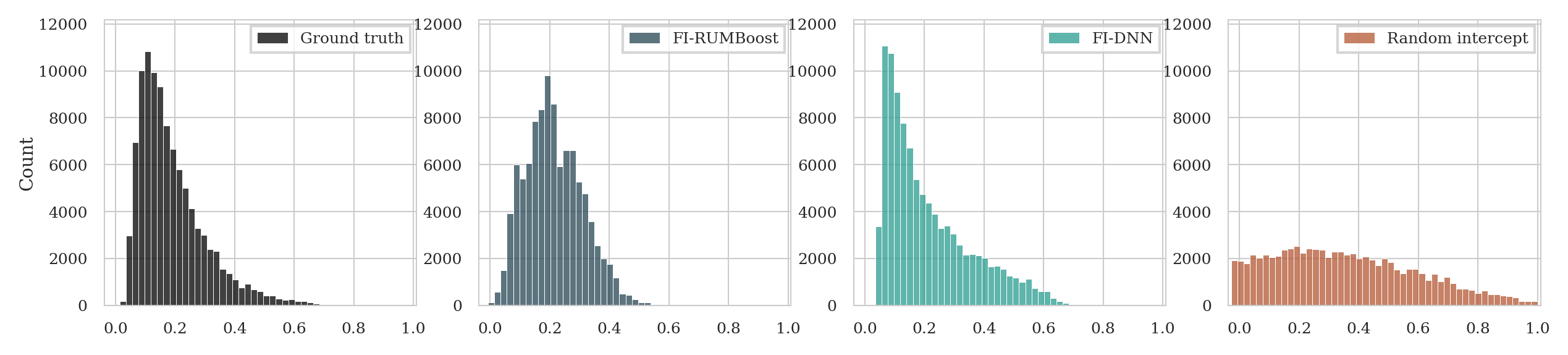}
        \caption{Intercept for alternative 1 - train set}
    \end{subfigure}
    \begin{subfigure}[b]{0.95\textwidth}
        \includegraphics[width=\linewidth]{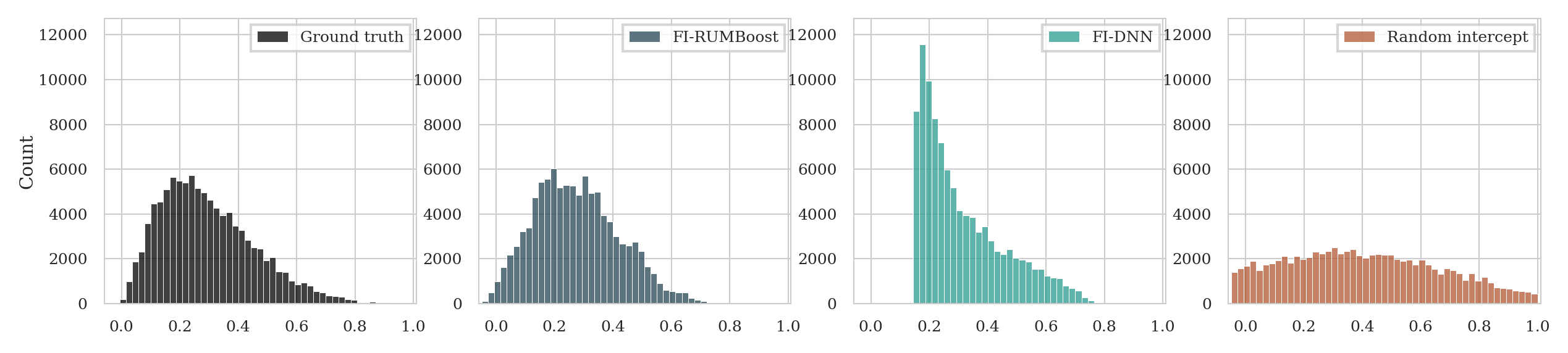}
        \caption{Intercept for alternative 2 - train set}
    \end{subfigure}
    \begin{subfigure}[b]{0.95\textwidth}
        \includegraphics[width=\linewidth]{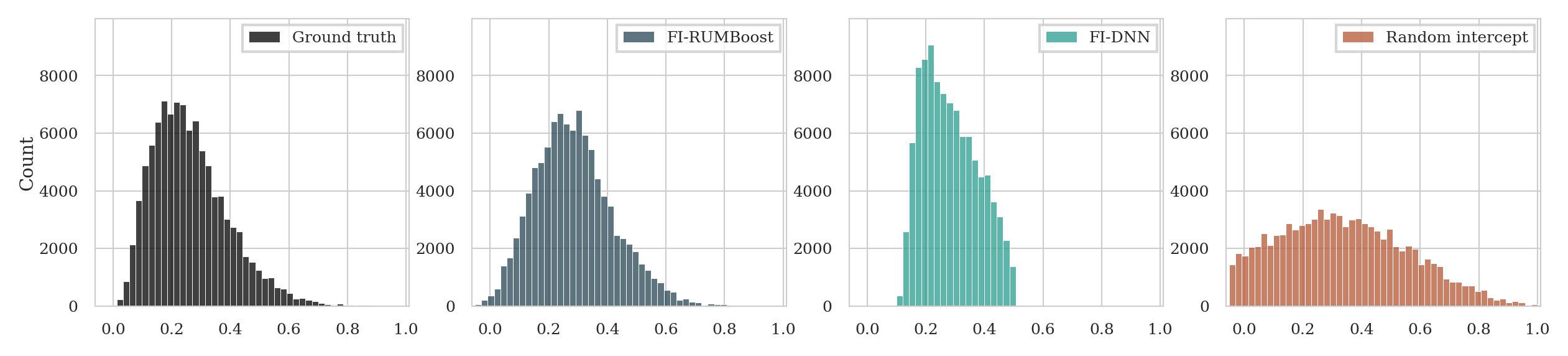}
        \caption{Functional intercept for alternative 3 - train set}
    \end{subfigure}
    \caption{Distribution of intercepts for the ground truth, FI-RUMBoost, FI-DNN, and Random Intercept model on synthetic train dataset. Note the plot for the Random Intercept model is truncated to [0,1] for comparison.}
    \label{functional effects synthetic}
\end{figure}

Figure \ref{functional effects synthetic} shows the recovered functional effects and parameters on the train set. The distributions of the functional effects are overall well-recovered by FI-RUMBoost and FI-DNN. We note that the distributions of the functional effects learnt with DNN are more concentrated around the mean for alternative 1 and 2. However, the functional effects learnt with GBDT seem to match more closely the true distribution. The Random Intercept model, on the other hand, fails to recover the true distribution, because of the \textit{a priori} assumption of normality. 

\begin{figure}[htbp]
    \begin{subfigure}[b]{0.95\textwidth}
        \includegraphics[width=\linewidth]{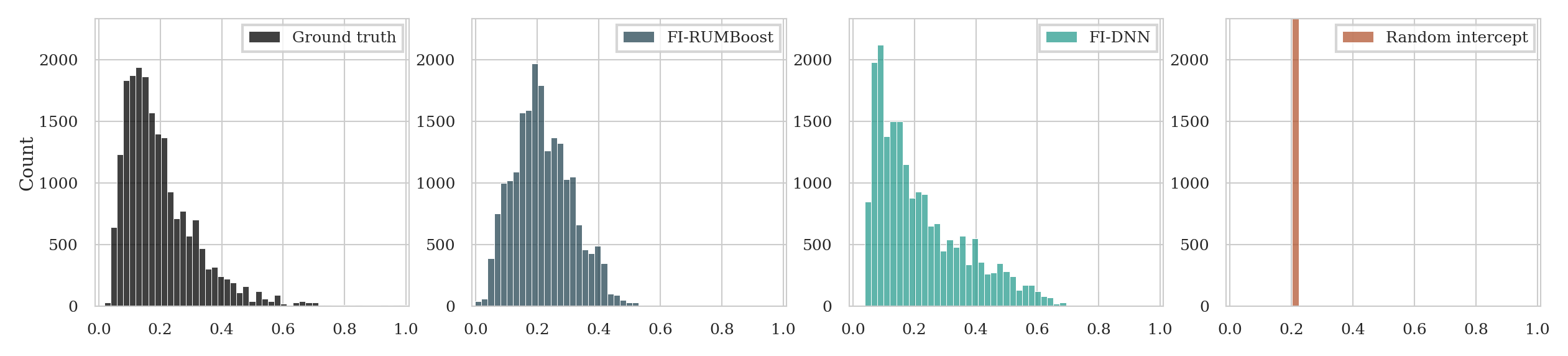}
        \caption{Intercept for alternative 1 - test set}
    \end{subfigure}
    \begin{subfigure}[b]{0.95\textwidth}
        \includegraphics[width=\linewidth]{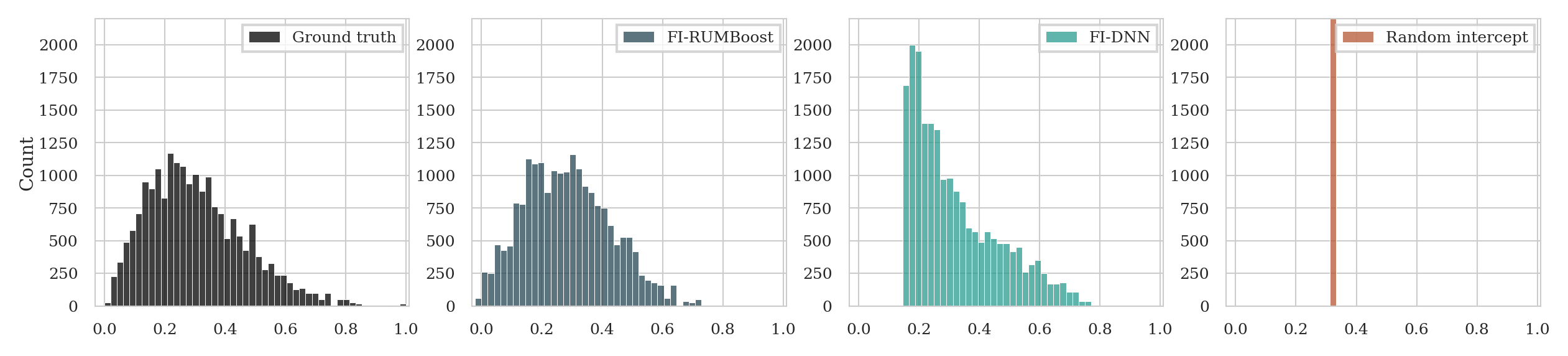}
        \caption{Intercept for alternative 2 - test set}
    \end{subfigure}
    \begin{subfigure}[b]{0.95\textwidth}
        \includegraphics[width=\linewidth]{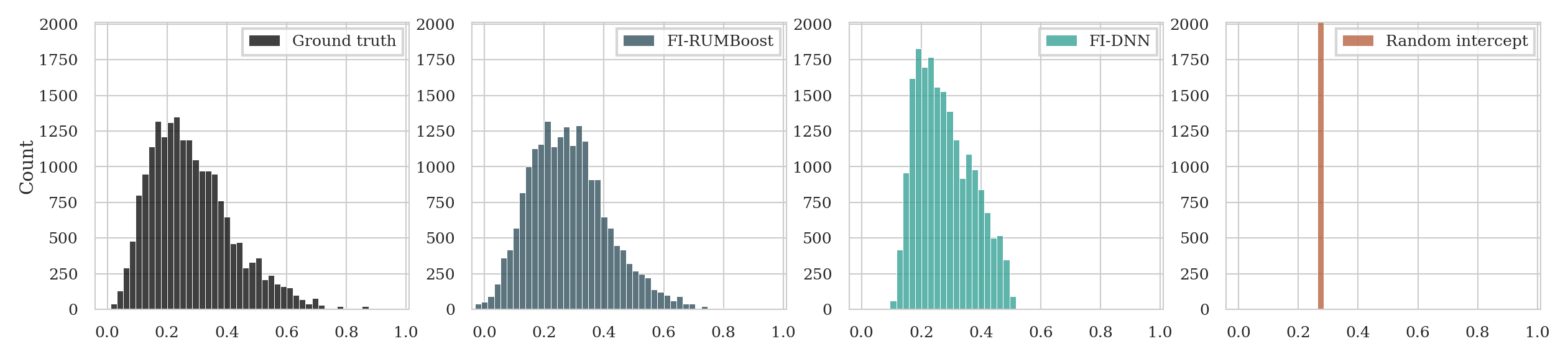}
        \caption{Intercept for alternative 3 - test set}
    \end{subfigure}
    \caption{Distribution of intercepts for the ground truth, FI-RUMBoost, FI-DNN, and Random Intercept model on synthetic test dataset. Note the Random Intercept uses the mean intercept values for forecasting and is truncated for comparison.}
    \label{functional effects synthetic test}
\end{figure}

Figure \ref{functional effects synthetic test} displays the recovered functional intercepts on the holdout test set. FI-RUMBoost and FI-DNN are able to recover the functional effects even on unseen data. These results indicate that the functional effects models are suitable for inference, and contrast with the limitations of the Random Intercept model, which can only output the mean of the random effects estimated during training. 

\begin{figure}[htbp]
    \begin{subfigure}[b]{0.24\textwidth}
        \includegraphics[width=\linewidth]{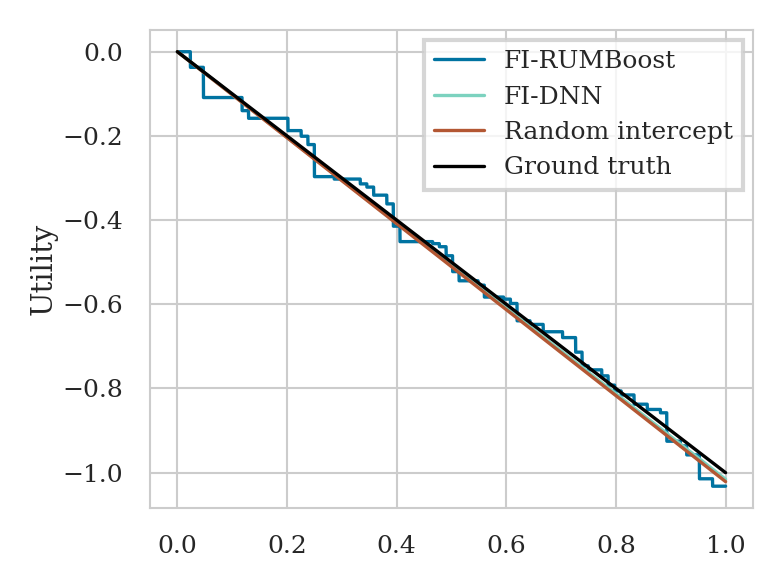}
        \caption{Variable 5.}
    \end{subfigure}
    \begin{subfigure}[b]{0.24\textwidth}
        \includegraphics[width=\linewidth]{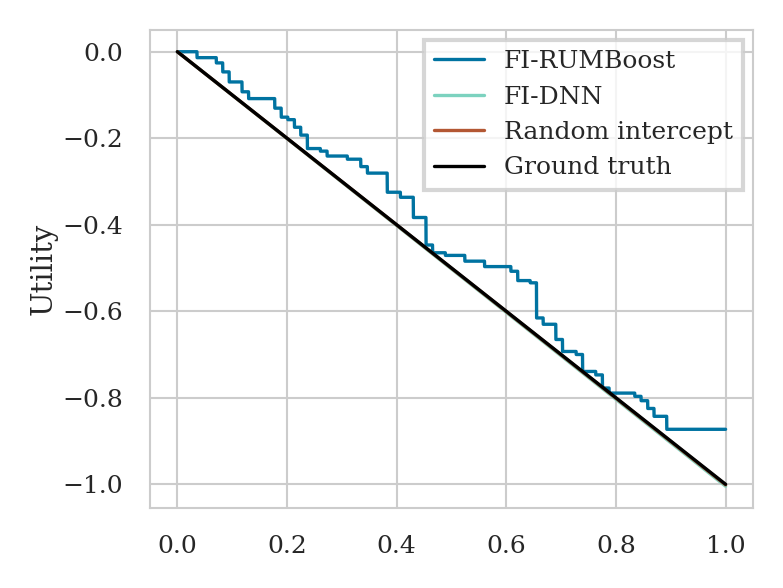}
        \caption{Variable 6.}
    \end{subfigure}
    \begin{subfigure}[b]{0.24\textwidth}
        \includegraphics[width=\linewidth]{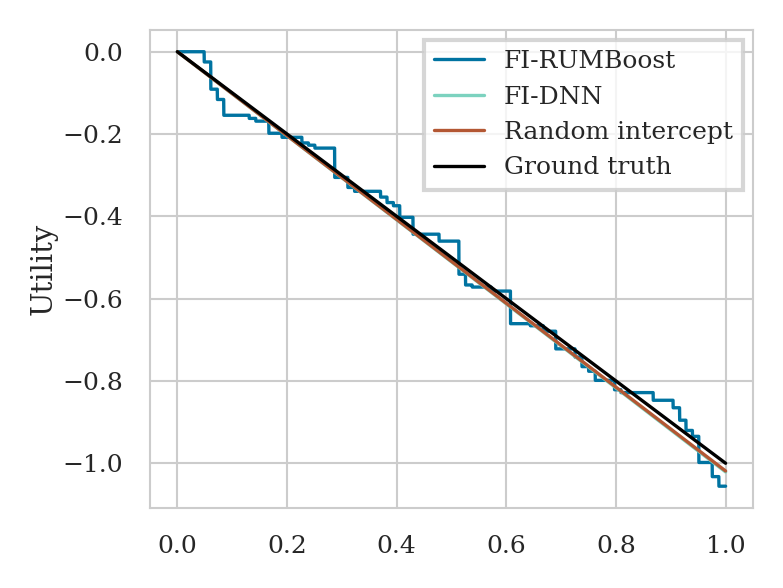}
        \caption{Variable 7.}
    \end{subfigure}
        \begin{subfigure}[b]{0.24\textwidth}
        \includegraphics[width=\linewidth]{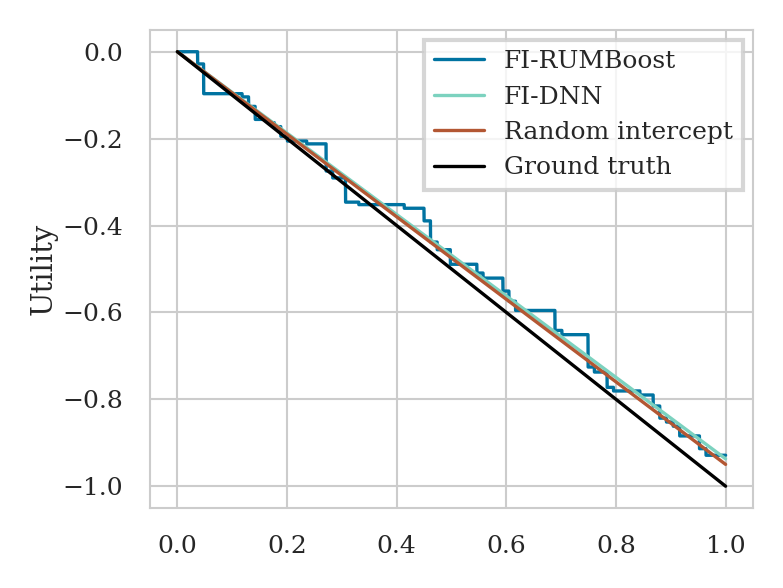}
        \caption{Variable 8.}
    \end{subfigure}
    \caption{Coefficients for the ground-truth, FI-RUMBoost, FI-DNN, and Random Intercept model. Note FI-RUMBoost uses non-linear coefficients $f_m(x_m)$, where the other models use linear coefficients $\beta_m x_m$.}
    \label{coefficients synthetic}
\end{figure}

Finally, we also observe in Figure~\ref{coefficients synthetic} that all models are able to recover the true linear parameters for all variables. This special linear case highlights the proficiency of \textit{linear-in-parameters} models to recover linear functions, but it is reassuring to notice that FI-RUMBoost, developed to discover non-linear coefficients, is still able to recover the linear functions. 

\section{Real-world case studies} \label{case studies}

\subsection{Mode choice datasets}\label{swissmetro}

\subsubsection{Datasets and model specifications}
We use the open-source Swissmetro \citep{bierlaire2001acceptance} and London Passenger Mode Choice (LPMC) \citep{hillel2018recreating} datasets for the benchmarks. The first dataset is a stated preference datasets where respondents are asked to choose between Swissmetro (a new and hypothetical underground mag-lev transportation mode in Switzerland), car, or train. We follow the same data preprocessing and model specification as in \cite{han2022neural}. We keep observations where the choice is known. After this preprocessing, we have $10692$ observations from $1192$ individuals (about 9 observations per individual) that we split into 70 \% for training, 15 \% for validation, and 15 \% for testing.

The second dataset is a revealed preference dataset obtained from the London Travel Demand Survey (LTDS) travel diary dataset, enriched with alternative-specific travel times obtained from the Google Directions API and a bespoke cost model. Possible alternatives are walking, cycling, public transport, and driving. We follow the same data preprocessing and model specification as in \cite{salvade2025rumboost}. The dataset contains 81086 trips made by 31954 individuals (about 2.5 trips per individual). The first two years of observation are used as a training set that we split at the household level into 80\% for training and 20\% for validation. 

Table \ref{vars in swissmetro and lpmc} summarises which variables are included in the models for the Swissmetro and LPMC datasets, as well as monotonicity constraints. Note that we assume that all alternatives are available for every individual. Table \ref{tab:swissmetro hyp search} and \ref{tab:lpmc hyp search} in \ref{app:hyperparameter search} summarise optimal hyperparameter values obtained from the hyperparameter search for all models.

\begin{table}[htbp]
\centering
\caption{Variables used in Swissmetro and LPMC datasets. A negative monotonic constraint indicates that an increase in the variable must decrease the utility function.}
\begin{tabular}{llll} \label{vars in swissmetro and lpmc}\\
\toprule
\textbf{Variable} & \textbf{Swissmetro} & \textbf{LPMC} & \textbf{Monotonic constr.} \\ \midrule
\textit{Alternatives} & Swissmetro, & Walking, & -- \\
 & Train, & Cycling,  & -- \\
 & Driving & PT,& -- \\
&  &  Driving & -- \\
\\
\multicolumn{4}{l}{\textit{Socio-demographics (inputs for functional effects)}} \\ 
Age & Yes & Yes & -- \\
Income & Yes & -- & -- \\
Gender & Yes & Yes & -- \\
Purpose of trip & Yes & Yes & -- \\
Has luggage & Yes & -- & -- \\
Pays for trip & Yes & -- & -- \\
Swiss seasonal ticket & Yes & -- & -- \\
First class & Yes & -- & -- \\
Driving license & -- & Yes & -- \\
N. of cars in household & -- & Yes & -- \\
Fuel type & -- & Yes & -- \\
\\

\multicolumn{4}{l}{\textit{Trip variables (alternative-specific variables)}} \\
Travel times & Yes & Yes & Negative \\
Costs & Yes & PT and Driving & Negative \\
Headway & Train and Swissmetro & -- & Negative \\
Seat type & Swissmetro only & -- & -- \\
Trip day & -- & Yes & -- \\
Start time & -- & Yes & -- \\
Distance & -- & Yes & Negative \\
Degree of congestion & -- & Driving only & Negative \\ \bottomrule

\end{tabular}
\end{table}

\subsubsection{Predictive performance}

We compare the different models with their negative cross-entropy loss on a holdout test set. We include in the benchmarks two out-of-the-box ML classifiers where the utility is computed as in Eq. \ref{out-of-the-box ml utility}: GBDT, using the LightGBM python library \citep{ke2017lightgbm}; and DNN, using the PyTorch python library \citep{10.1145/3620665.3640366}. These models provide state-of-the-art predictive performance without accounting for inter-heterogeneity, therefore being useful to measure the impact of accounting for preference heterogeneity. The results are shown in Table \ref{benchmark swissmetro}. Overall, all functional effects models learnt with GBDTs exhibit better predictive performance than the functional effects models learnt with DNNs on the Swissmetro dataset. FIS-GBDT performs the best, while the interpretable baselines perform the least well. We note that GBDT, a blackbox baseline, is the second best performing model, but the model has full trip feature interaction, as opposed to the functional effects models and interpretable baselines, which have no trip feature interaction. Looking at the computational time, FS-DNN and FIS-DNN are faster than the FS-GBDT and FIS-GBDT, whereas FI-RUMBoost and RUMBoost are slightly faster than FI-DNN and MNL.
On the LPMC dataset, FI-RUMBoost performs best. On this dataset, FS-DNN and FIS-DNN perform better than their GBDT counterpart, whereas the baseline models perform the least well. We deduce from these results that non-linear coefficients and inter-individual heterogeneity are of greater importance in this dataset. Finally, the functional effects models learnt with GBDTs are faster than the ones learnt with DNNs, and all models are faster proportionally to the number of functional effects in the model.

\begin{table}[htbp]
\centering
\caption{Benchmarks on the holdout test set. The models are evaluated on the negative cross-entropy loss (lower the better). We also report the training time with optimal hyperparameters in seconds. Best results are highlighted in bold. FE means functional effects. Baseline models are trained without socio-demographic characteristics.}
\label{benchmark swissmetro}
\begin{tabular}{llrrrr}
\toprule
 &  & \multicolumn{2}{c}{\textbf{Swissmetro}} & \multicolumn{2}{c}{\textbf{LPMC}} \\
 & \textbf{Model} & \textbf{CEL} & \textbf{Time [s]} & \textbf{CEL} & \textbf{Time [s]} \\
\midrule
\multirow[c]{3}{*}{\textbf{GBDT-based FE}} & FIS-GBDT & \textbf{0.614} & 9.02 & 0.699 & 15.40 \\
 & FS-GBDT & 0.679 & 7.32 & 0.724 & 12.87 \\
 & FI-RUMBoost & 0.630 & 1.76 & \textbf{0.673} & 3.40 \\
\cline{2-6}
\multirow[c]{3}{*}{\textbf{DNN-based FE}} & FIS-DNN & 0.670 & 6.04 & 0.691 & 69.19 \\
 & FS-DNN & 0.720 & 3.04 & 0.714 & 66.27 \\
 & FI-DNN & 0.779 & 12.18 & 0.705 & 52.21 \\
 \cline{2-6}
 \multirow[c]{2}{*}{\textbf{Baselines - Interpretable}}
 & RUMBoost* & 0.786 & 1.54 & 0.824 & 3.36 \\
 & MNL* & 0.854 & 9.74 & 0.841 & 29.00 \\
 \cline{2-6}
  \multirow[c]{2}{*}{\textbf{Baselines - Blackbox}}
 & GBDT* & 0.622 & 18.92 & 0.805 & 15.30 \\
 & DNN* & 0.746 & 1.76 & 0.840 & 29.51 \\ \bottomrule
  \multicolumn{6}{r}{*Baseline models are trained without socio-demographic characteristics.} \\

\end{tabular}
\end{table}

\subsubsection{Model parameters}

Due to the large number of parameters and model combinations across multiple alternatives for the Swissmetro and LPMC, we do not present a structured overview of the model parameters for each model in this paper. We direct the reader to \citet{salvade2025rumboost} for an in-depth discussion of the FI-RUMBoost parameters on the LPMC data. For the remaining models and DNN-based models, all results are available on GitHub\footnote{https://github.com/big-ucl/functional-effects-model}. 

We summarise key findings for the model coefficients and functional effects as follows: 

\begin{itemize}
    \item The models with non-linear coefficients show the biggest differences with models with linear coefficients on continuous variables such as travel time or trip starting time;
    \item The functional effects are more scattered on the LPMC dataset compared to the Swissmetro dataset;
    \item The functional effects learnt with GBDT are more likely to experience extreme negative values; and
    \item The monotonic constraints push some functional slopes towards zero for most individuals.
\end{itemize}

\subsection{EasySHARE: modelling the mental health of elder people in Europe} \label{case study}
\subsubsection{Dataset}

For this case study, we use data from the Survey of Health, Ageing and Retirement in Europe (SHARE), a longitudinal study of older people's health across 28 European countries. We use the simplified easySHARE dataset \citep{easyshare}, which has been preprocessed to combine the data from the 9 waves of the study in a long table format. For this study, we follow \citet{mendorf2023prospective} where we model the EURO-D measure of depressive symptoms, a scale composed of 12 depressive symptoms. A measurement of 0 indicates that the patient has no depressive symptoms and a score of 12 means that the patient exhibits all 12 depressive symptoms. This is an ordinal target variable, meaning that we need to adapt the generic multiclass methodology described in \ref{multiclass}. This is done using the CORAL methodology \citep{shi2023deep}, with complete derivations in \ref{app:ordinal}.

We preprocess the dataset to remove missing values for the target variable and drop variables with more than 10\% missing values. We further remove observations that would still have missing values. We drop wave 3 and wave 7 observations as the survey differs from other waves. We encode all categorical variables with dummy variables, where we normalise one category to 0. After preprocessing, we have 281975 observations from 130620 individuals (about 2.15 observations per individual). We split the data at the individual level to avoid data leakage, keeping 20\% for the hold-out test set and 80\% for training. We further split (also at the individual level) the training set into 80\% for training and 20\% for validation, to perform a hyperparameter search for both functional effects models with GBDTs and DNNs. Table \ref{tab:easyshare hyp search} in \ref{app:hyperparameter search} summarises the optimal hyperparameter values from the hyperparameter search. Table \ref{variable description easySHARE} summarises and provides a brief description of all variables used in the model. 

\begin{table}[htbp]
\centering
\caption{easySHARE dataset variable descriptions and types}
\footnotesize
\label{variable description easySHARE}
\begin{tabular}{lll}
\toprule
\textbf{Variable} & \textbf{Description} & \textbf{Type} \\
\midrule
\textit{Target} & & \\
eurod & Depression scale (0–12) & Ordinal \\
\midrule
\multicolumn{2}{l}{\textit{Socio-demographics coefficients (inputs for functional effects):}} & \\
age & Age in years at interview & Continuous \\
female & Gender (1 - female, 0 - male) & Binary \\
country & Country of residence & Nominal \\
mar\_stat & Marital status & Nominal \\
dn004\_mod & Respondent born in interview country (yes/no) & Binary \\
isced1997\_r & Education level (1-6) & Ordinal \\
thinc\_m & Household net income & Continuous \\
ch001\_ & Number of children & Discrete \\
hhsize & Household size & Discrete \\
partnerinhh & Partner lives in the same household (yes/no) & Binary \\
mother\_alive & Mother alive (yes/no) & Binary \\
father\_alive & Father alive (yes/no) & Binary \\
sp002\_mod & Receives help from outside household (yes/no) & Binary \\
smoking & Respondent smokes (yes/no) & Binary \\
ever\_smoked & Respondent ever smoked (yes/no) & Binary \\
br015\_ & Frequency of vigorous activities & Ordinal \\
ep005\_ & Job situation & Nominal \\
co007\_ & Household meets end & Ordinal \\
has\_citizenship & Respondent has citizenship (yes/no) & Binary \\
\midrule
\textit{Situational variables:} & & \\
\textit{Health indices \& conditions} & & \\
bmi & Body mass index & Continuous \\
sphus & Self-perceived health (1 - excellent to 5 - poor) & Ordinal \\
chronic\_mod & Number of chronic diseases & Discrete \\
maxgrip & Maximum grip strength (kg) & Continuous \\ 
\textit{Functional \& mobility scores} & & \\
adla  & Sum of difficulty in daily tasks (0–5) & Ordinal \\
iadlza  & Instrumental activities difficulties (0–5) & Ordinal \\
mobilityind & Mobility difficulties (0–4) & Ordinal \\
lgmuscle & Large muscle functioning difficulties (0–4) & Ordinal \\
grossmotor & Gross motor skills difficulties (0–4) & Ordinal \\ 
finemotor & Fine motor skills difficulties (0–3) & Ordinal \\ 
\textit{Cognitive measures} & &\\
recall\_1 & Immediate word recall score (0–10) & Discrete \\
recall\_2 & Delayed word recall score (0–10) & Discrete \\
orienti & Orientation score (0–4) & Discrete \\
numeracy\_1 & Numeracy test score & Discrete \\ 
\textit{Healthcare usage} & & \\
hc002\_ & Number of doctor visits last 12 months & Discrete \\
hc012\_ & Hospital stay in last 12 months (yes/no) & Binary \\
hc029\_ & Nursing home in last 12 months (perm./temp./no) & Ordinal \\
\bottomrule
\end{tabular}
\end{table}

\subsubsection{Predictive performance}

We report the performance of the 8 different models described in \ref{model benchamrked} with respect to the MAE, EMAE, and MCEL on the holdout test set (see \ref{app:ordinal} for complete derivations of the metrics). Table \ref{benchmarks table} shows all these results.

\begin{table}[htbp]  
\centering
\caption{Benchmarks on the holdout test set. The models are evaluated on the mean absolute error, expected mean absolute error, and multi-label cross entropy loss. All metrics are lower the better. We also report the training time with optimal hyperparameters in seconds. Best results are highlighted in bold. FE means functional effects. Baseline models are trained without socio-demographic characteristics.}
\label{benchmarks table}
\begin{tabular}{llrrrr}
\toprule
 & \textbf{Model} & \textbf{MAE} & \textbf{EMAE} & \textbf{MCEL} & \textbf{Time [s]} \\
\midrule
 \multirow[c]{3}{*}{\textbf{GBDT-based FE}} & FIS-GBDT & 1.369 & \textbf{0.146} & \textbf{0.251} & 169 \\
 & FS-GBDT & 1.37 & \textbf{0.146} & \textbf{0.251} & 178 \\
  &  FI-RUMBoost & \textbf{1.368} & \textbf{0.146} & \textbf{0.251} & 241 \\
  \cline{2-6}
\multirow[c]{3}{*}{\textbf{DNN-based FE}} & FIS-DNN & 1.373 & 0.148 & 0.252 & 61 \\
 & FS-DNN & 1.371 & \textbf{0.146} &\textbf{0.251} & 65 \\
 & FI-DNN & 1.38 & 0.148 & 0.253 & 44 \\
 \cline{2-6}
\multirow[c]{2}{*}{\textbf{Baselines}} & RUMBoost* & 1.414 & 0.151 & 0.26 & 807 \\
 & Ordinal Logit* & 1.421 & 0.152 & 0.261 & 50 \\
\bottomrule
\multicolumn{6}{r}{*Baseline models are trained without socio-demographic characteristics.} \\
\end{tabular}
\end{table}

On average, all models predict the number of depressive symptoms between $1.368$ to $1.421$ away from the observed measurement. We observe that all models accounting for preferences perform better than models without any inter-individual heterogeneity. FI-RUMBoost has the best MAE, whereas FI-RUMBoost, FS-GBDT, FIS-GBDT, and FS-DNN exhibit the best MCEL and EMAE. We remark that the metric difference between models with functional effects is minimal. Since we perform only a single train--validate--test split for computational reasons, it is difficult to conclude that one model is better than another. In terms of computational time, learning functional effects with DNNs is about 3 to 16 times faster than with GBDTs. Finally, we also compare the estimation of the ordinal thresholds for all models in \ref{app:thresholds}. It is reassuring to observe that all values are in the same range, with minor differences.

\subsubsection{Model parameters}
The primary advantage of using ML regressors in a constrained setting, such as functional effects models, compared to out-of-the-box ML classification models, is that they are interpretable. When trained without functional slopes, we can observe the traditional parameters as in an Ordinal Logit from a functional effects model with linear coefficients and the non-linear utility function from functional effects models with non-linear coefficients. We can also observe the distribution of the individual-specific functional intercept and functional slope values as histograms. 

Figure \ref{fig:health_measures_3x7} shows the linear and non-linear effects from the four models without functional slopes. Overall, we observe that the linear and non-linear utility output from the functional effects models exhibit the same trends. An increase in the number of conditions, number of doctor visits, fine motor difficulties, large muscle difficulties, mobility difficulties, daily activity difficulties, instrumental activity difficulties, if the respondent has been hospitalised or temporarily in a nursing home, and if the self-perceived health is not excellent (reference category) increases the likelihood of having more depressive symptoms. On the other hand, a higher BMI, better gross motor difficulties, being able to recall more words, having a better max grip strength, and living permanently in a nursing home decrease the likelihood of having more depressive symptoms. We also observe that the non-linearity provides benefits for: the number of doctor visits, which is logarithmic; the BMI, which exhibits a plateau; and the mobility difficulties, which is a parabola. 

\begin{figure}[htbp]
    \centering

    \begin{subfigure}[b]{0.3\textwidth}
        \includegraphics[width=\linewidth]{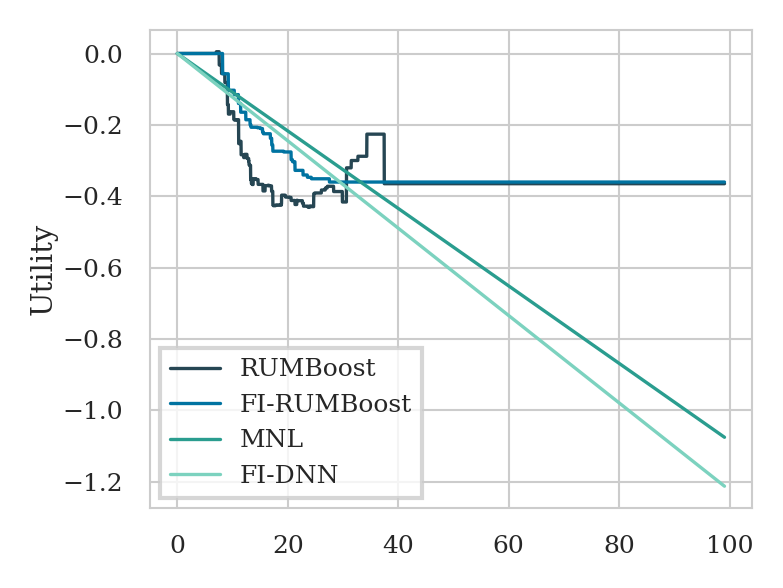}
        \caption{BMI}
    \end{subfigure}
    \begin{subfigure}[b]{0.3\textwidth}
        \includegraphics[width=\linewidth]{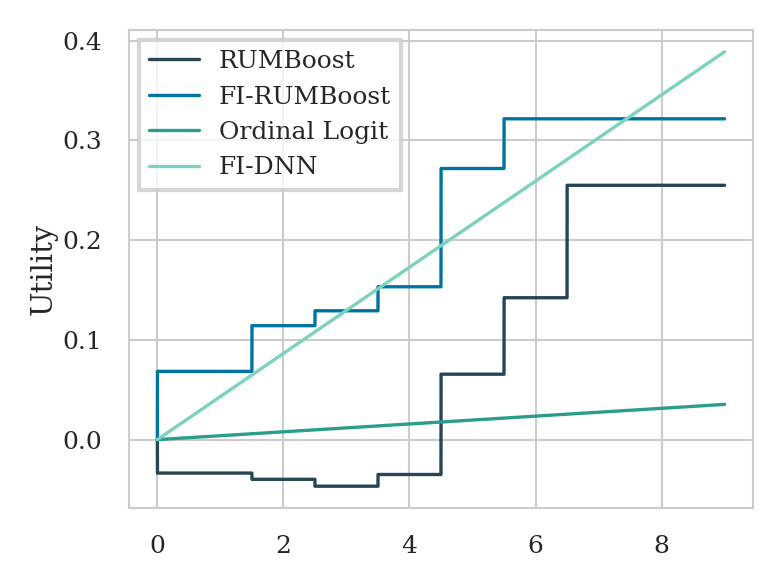}
        \caption{Number of chronic conditions}
    \end{subfigure}
    \begin{subfigure}[b]{0.3\textwidth}
        \includegraphics[width=\linewidth]{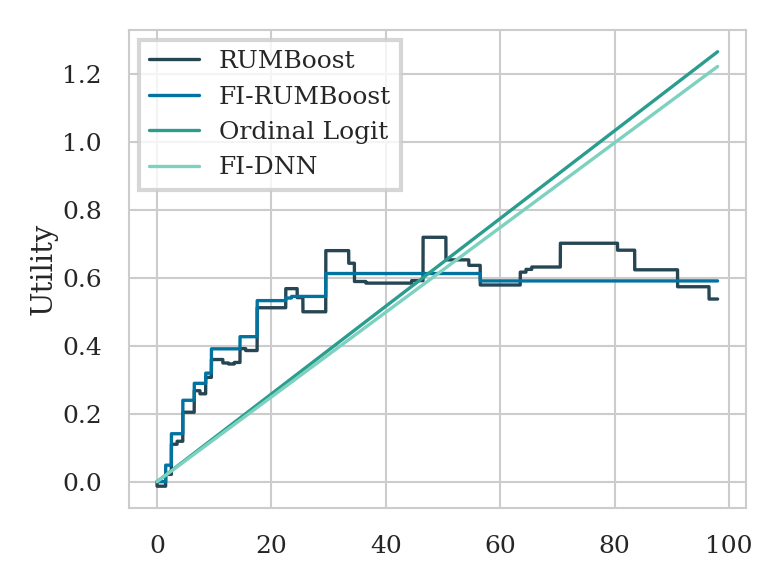}
        \caption{Number of doctor visits}
    \end{subfigure}

    \begin{subfigure}[b]{0.3\textwidth}
        \includegraphics[width=\linewidth]{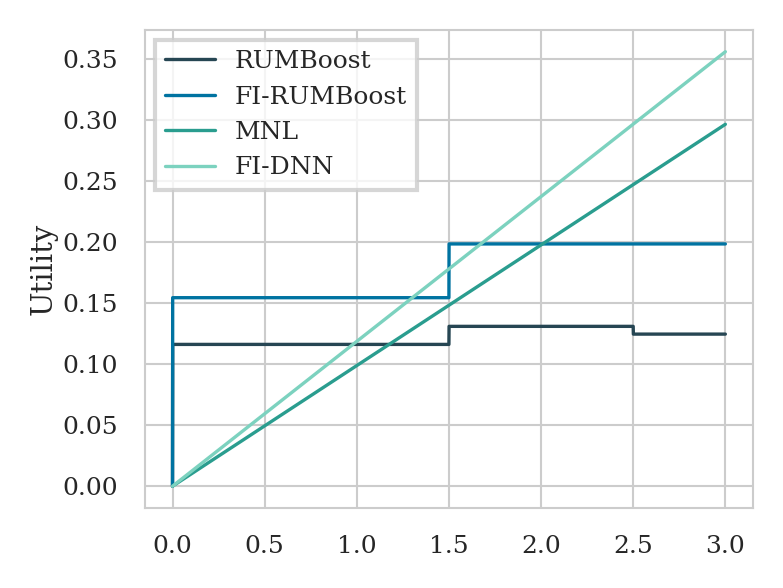}
        \caption{Fine motor skills}
    \end{subfigure}
    \begin{subfigure}[b]{0.3\textwidth}
        \includegraphics[width=\linewidth]{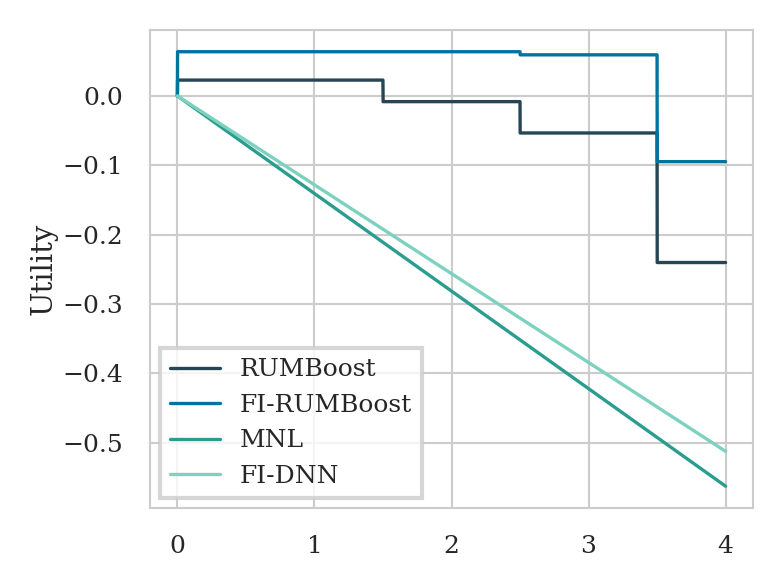}
        \caption{Gross motor skills}
    \end{subfigure}
    \begin{subfigure}[b]{0.3\textwidth}
        \includegraphics[width=\linewidth]{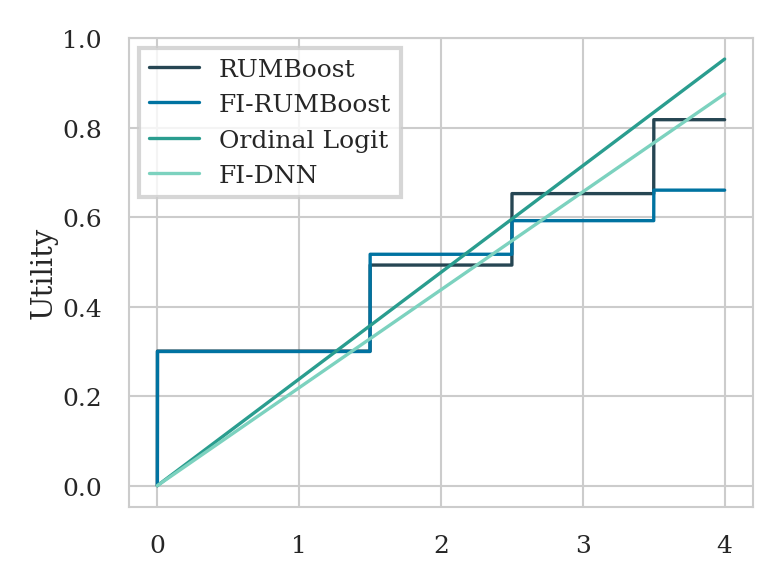}
        \caption{Large muscle skills}
    \end{subfigure}

    \begin{subfigure}[b]{0.3\textwidth}
        \includegraphics[width=\linewidth]{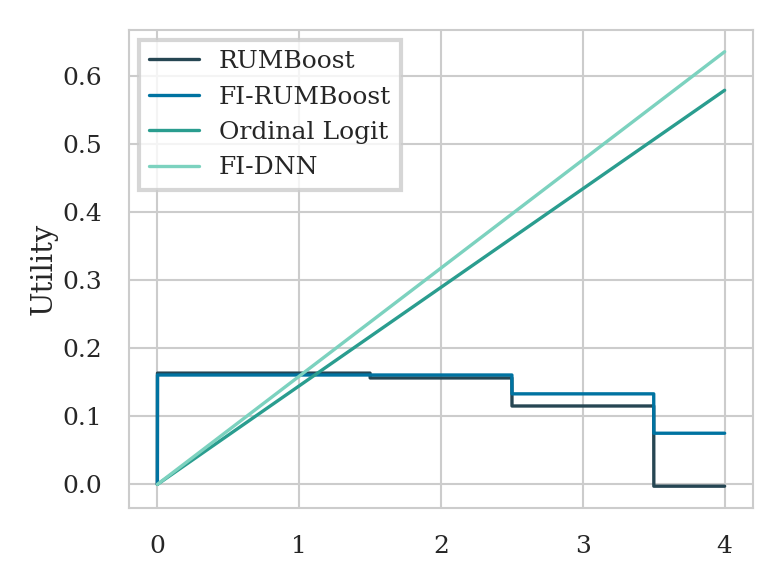}
        \caption{Mobility index}
    \end{subfigure}    
    \begin{subfigure}[b]{0.3\textwidth}
        \includegraphics[width=\linewidth]{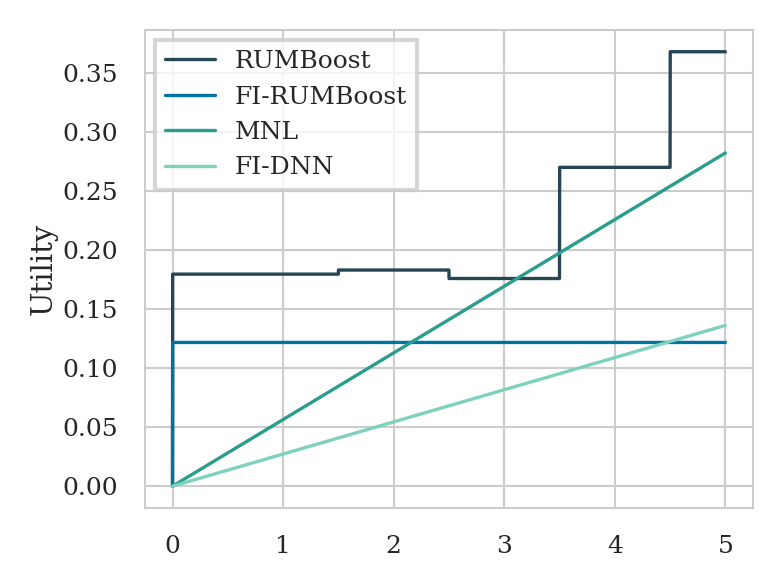}
        \caption{Daily activities index}
    \end{subfigure}
    \begin{subfigure}[b]{0.3\textwidth}
        \includegraphics[width=\linewidth]{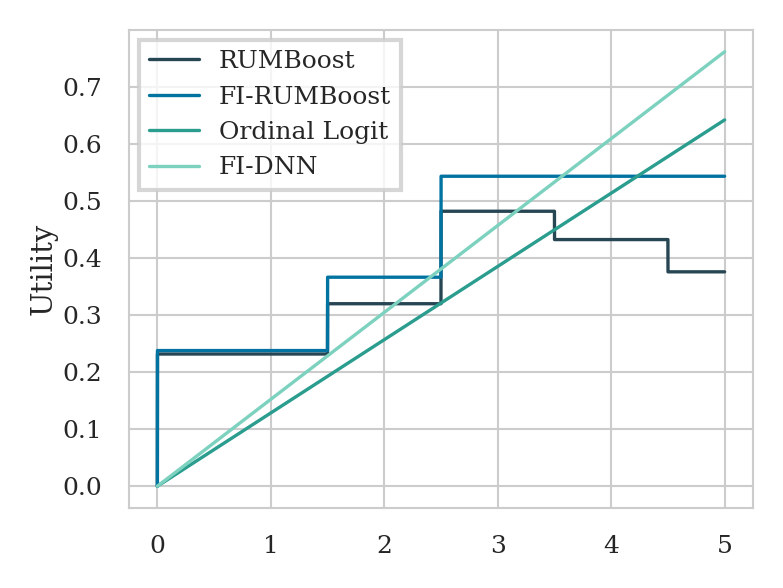}
        \caption{Instrumental activities index}
    \end{subfigure}

        
    \begin{subfigure}[b]{0.3\textwidth}
        \includegraphics[width=\linewidth]{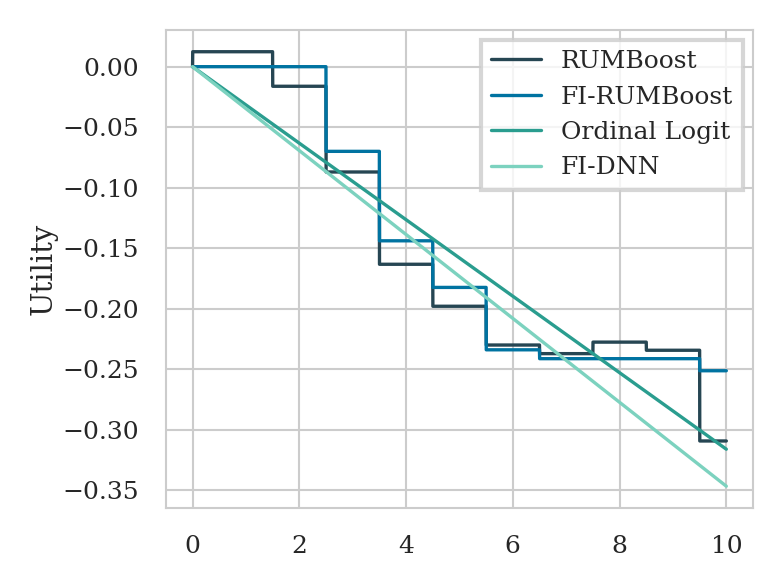}
        \caption{Recall 1}
    \end{subfigure}
    \begin{subfigure}[b]{0.3\textwidth}
        \includegraphics[width=\linewidth]{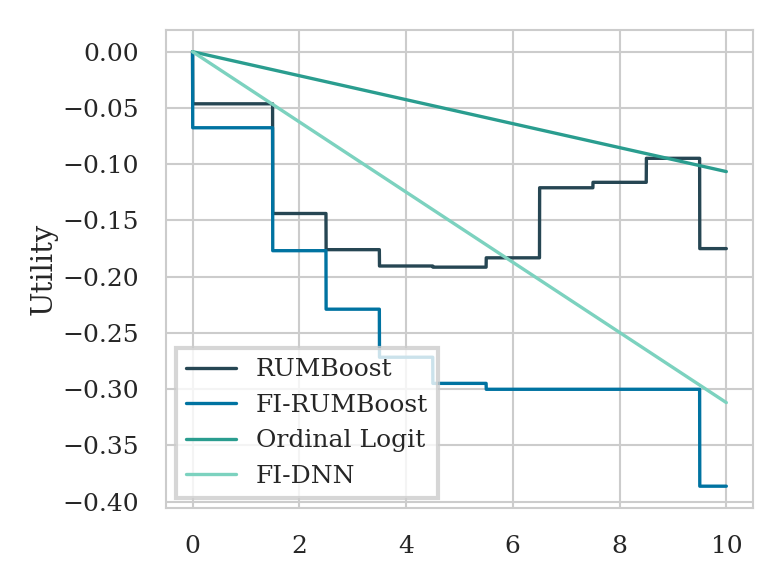}
        \caption{Recall 2}
    \end{subfigure}
    \begin{subfigure}[b]{0.3\textwidth}
        \includegraphics[width=\linewidth]{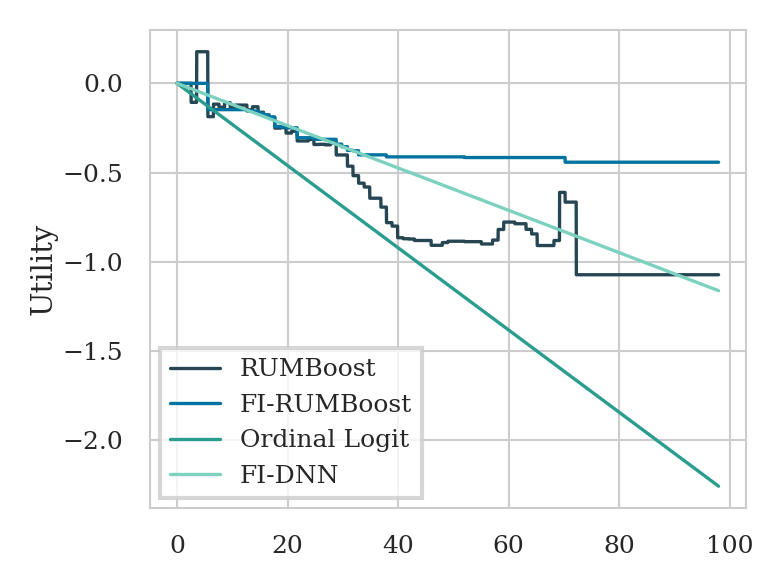}
        \caption{Max grip strength}
    \end{subfigure}

    \begin{subfigure}[b]{0.3\textwidth}
        \includegraphics[width=\linewidth]{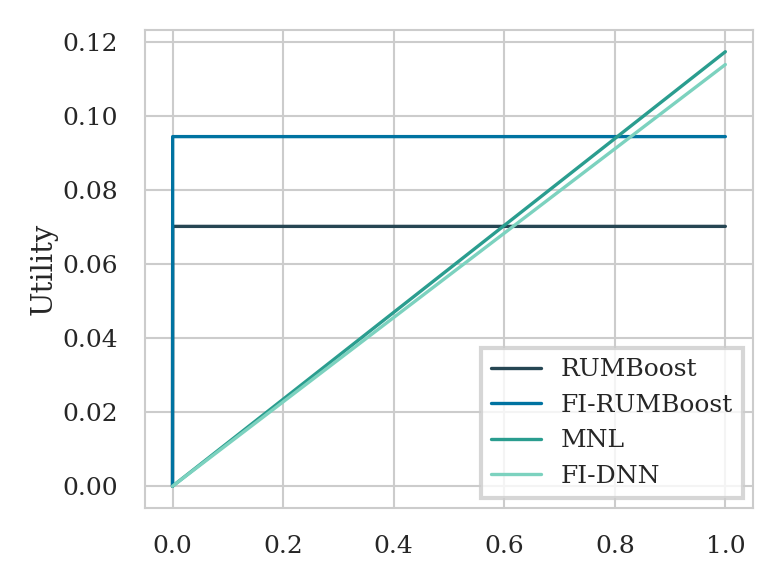}
        \caption{Hospitalised last year}
    \end{subfigure}
    \begin{subfigure}[b]{0.3\textwidth}
        \includegraphics[width=\linewidth]{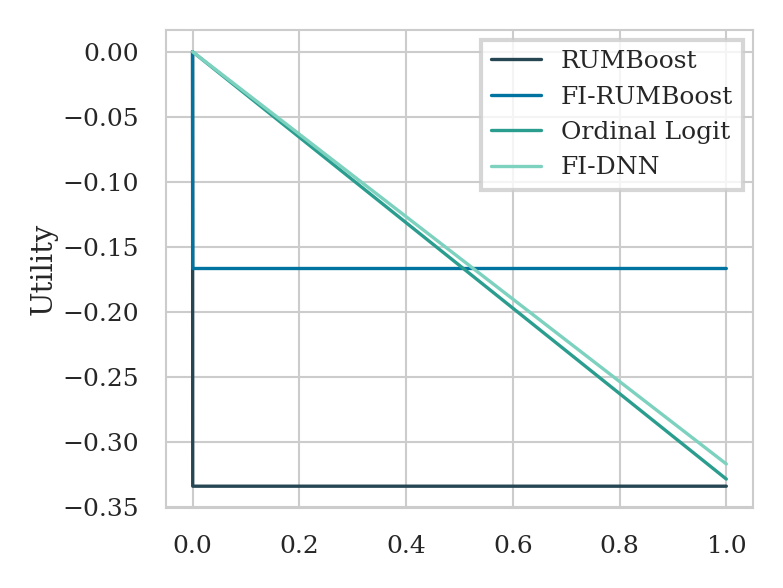}
        \caption{Nursing home (permanently)}
    \end{subfigure}
    \begin{subfigure}[b]{0.3\textwidth}
        \includegraphics[width=\linewidth]{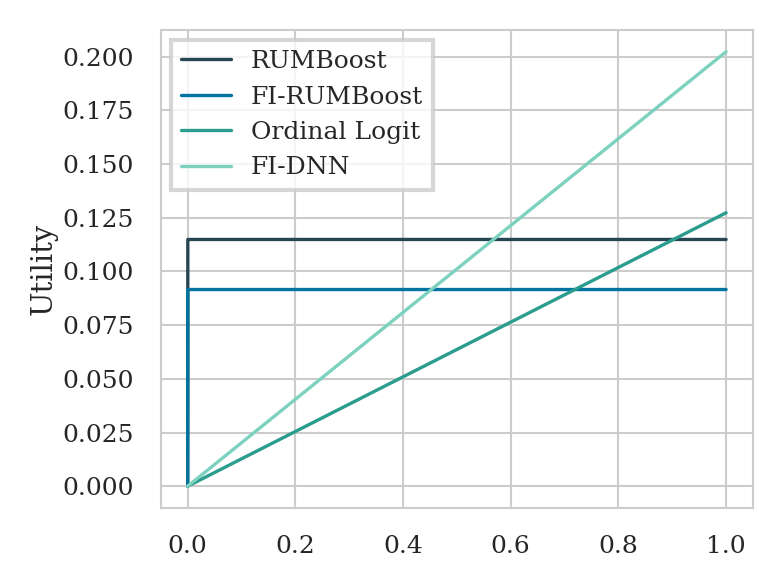}
        \caption{Nursing home (temporarily)}
    \end{subfigure}

\end{figure}

\begin{figure}[htbp]\ContinuedFloat
    \centering

    \begin{subfigure}[b]{0.3\textwidth}
        \includegraphics[width=\linewidth]{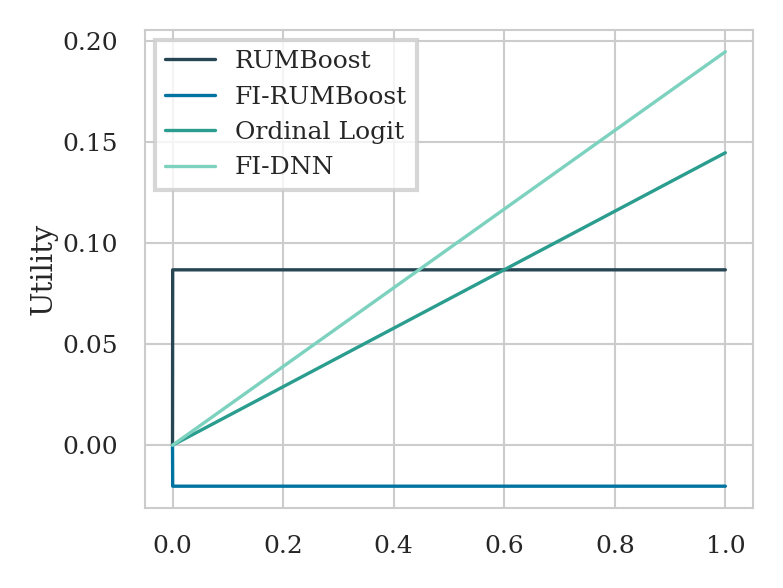}
        \caption{Self-perceived health -- very good}
    \end{subfigure}
    \begin{subfigure}[b]{0.3\textwidth}
        \includegraphics[width=\linewidth]{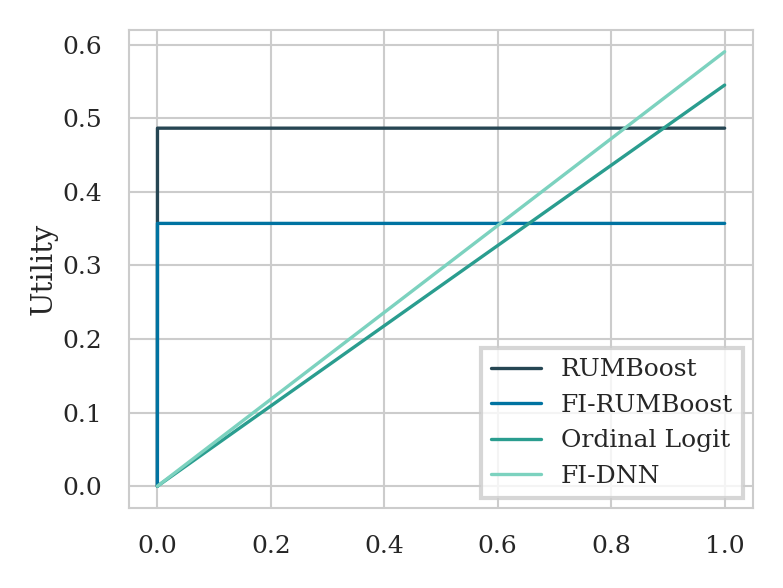}
        \caption{Self-perceived health -- good}
    \end{subfigure}   
    \begin{subfigure}[b]{0.3\textwidth}
        \includegraphics[width=\linewidth]{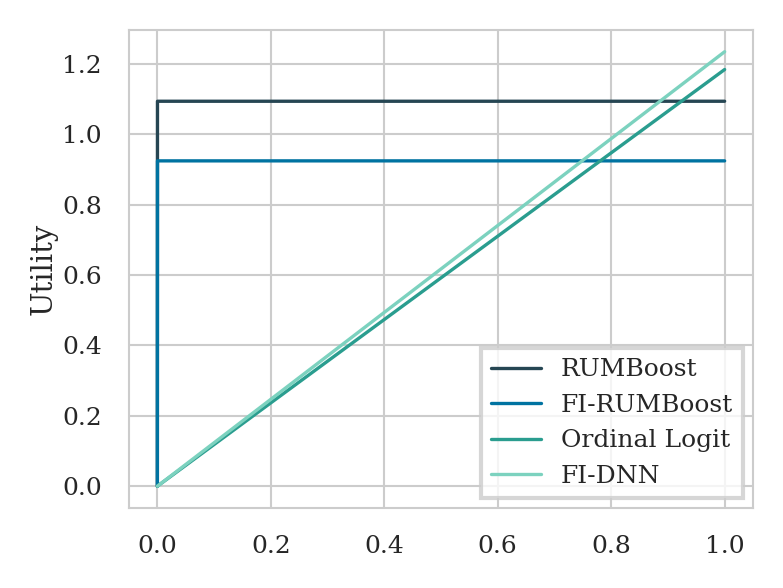}
        \caption{Self-perceived health -- fair}
    \end{subfigure}

    \begin{subfigure}[b]{0.3\textwidth}
        \includegraphics[width=\linewidth]{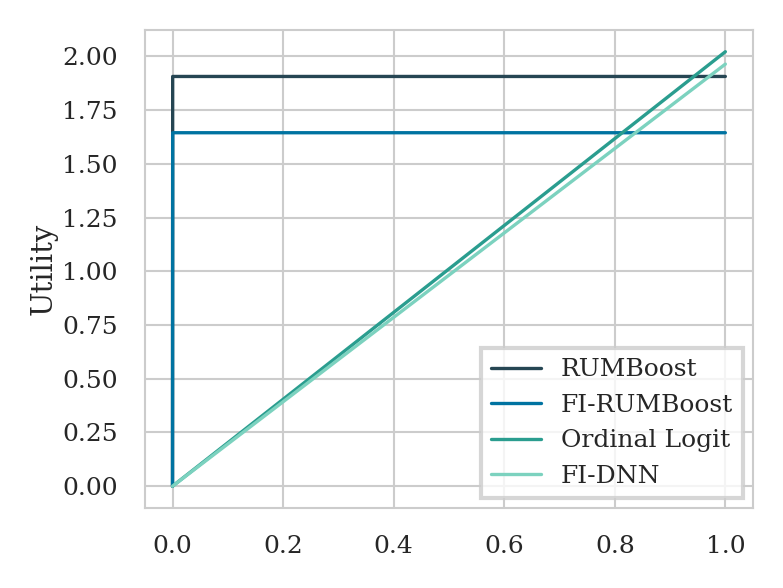}
        \caption{Self-perceived health -- poor}
    \end{subfigure} 

    \caption{Non-linear and linear coefficients of health-related variables on the EURO D depression scale measurement.}
    \label{fig:health_measures_3x7}
\end{figure}

\begin{figure}[htbp]
    \begin{subfigure}[b]{0.45\textwidth}
        \includegraphics[width=\linewidth]{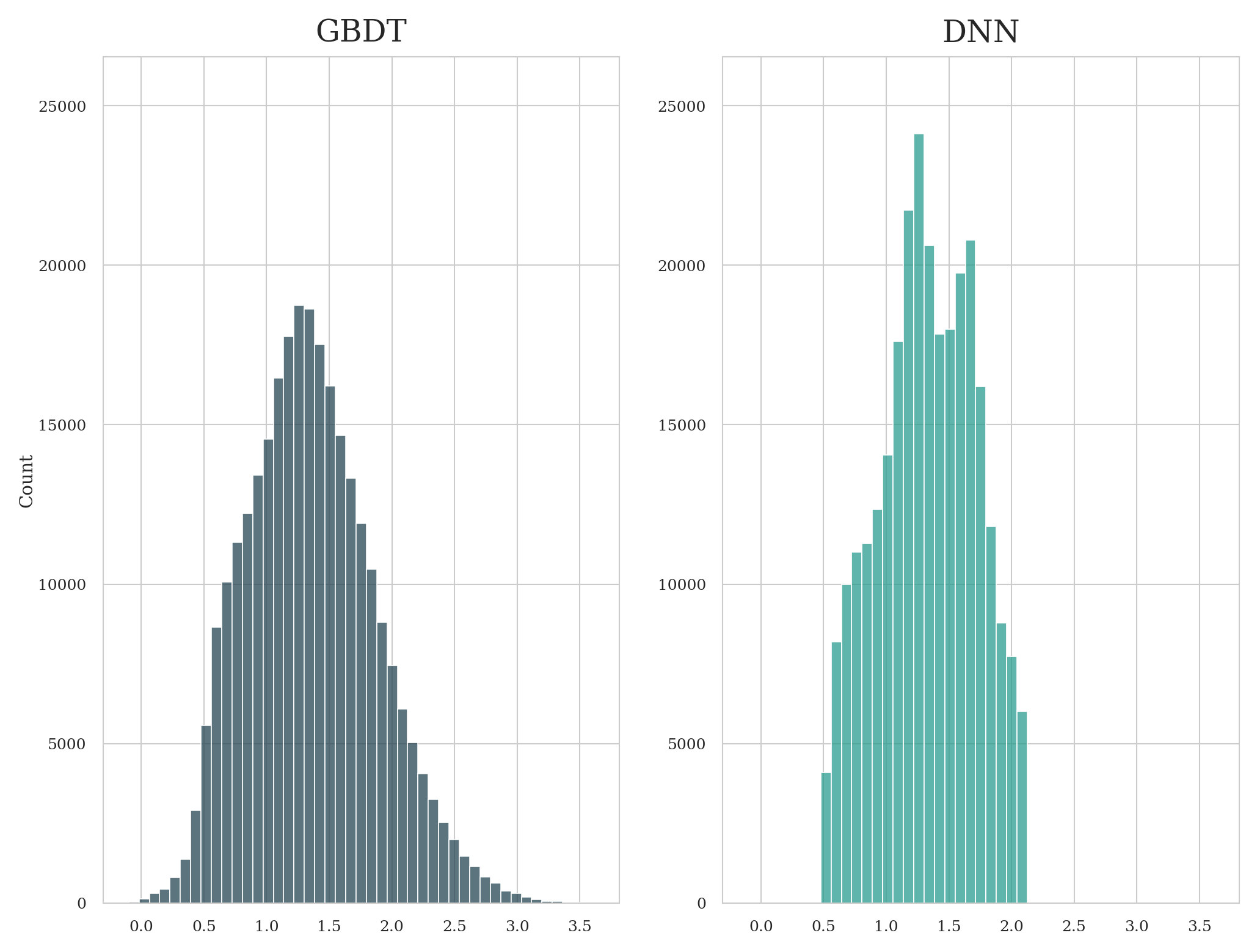}
        \caption{Functional intercept}
    \end{subfigure}
    \begin{subfigure}[b]{0.45\textwidth}
        \includegraphics[width=\linewidth]{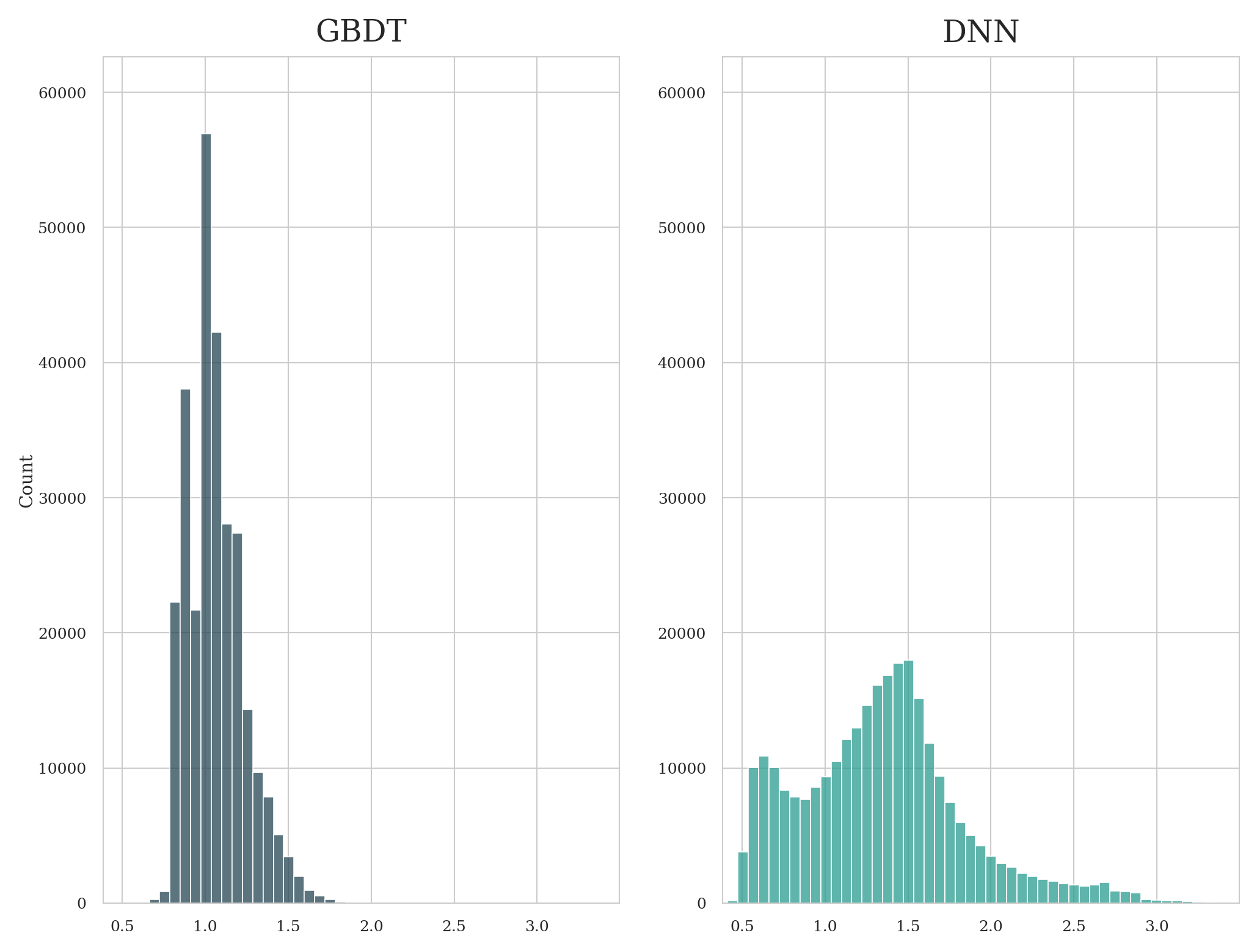}
        \caption{Functional intercept - models with functional slopes}
    \end{subfigure}
    \caption{Functional intercepts learnt with GBDT and DNN for FI-RUMBoost and FI-DNN (a) and FIS-GBDT and FIS-DNN (b).}
    \label{functional intercept}
\end{figure}

\begin{figure}[htbp]
    \begin{subfigure}[b]{0.45\textwidth}
        \includegraphics[width=\linewidth]{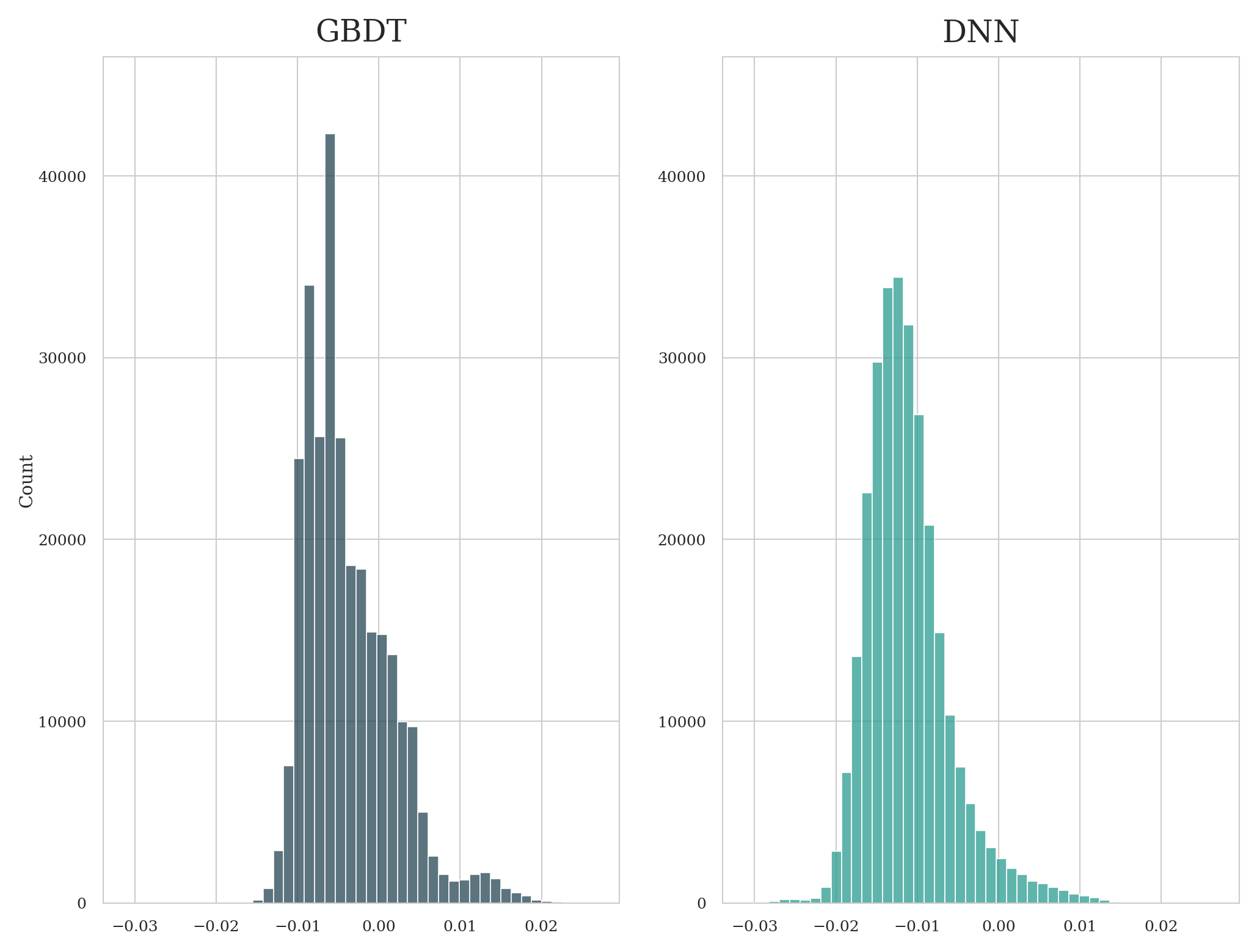}
        \caption{Max grip strength}
    \end{subfigure}
    \begin{subfigure}[b]{0.45\textwidth}
        \includegraphics[width=\linewidth]{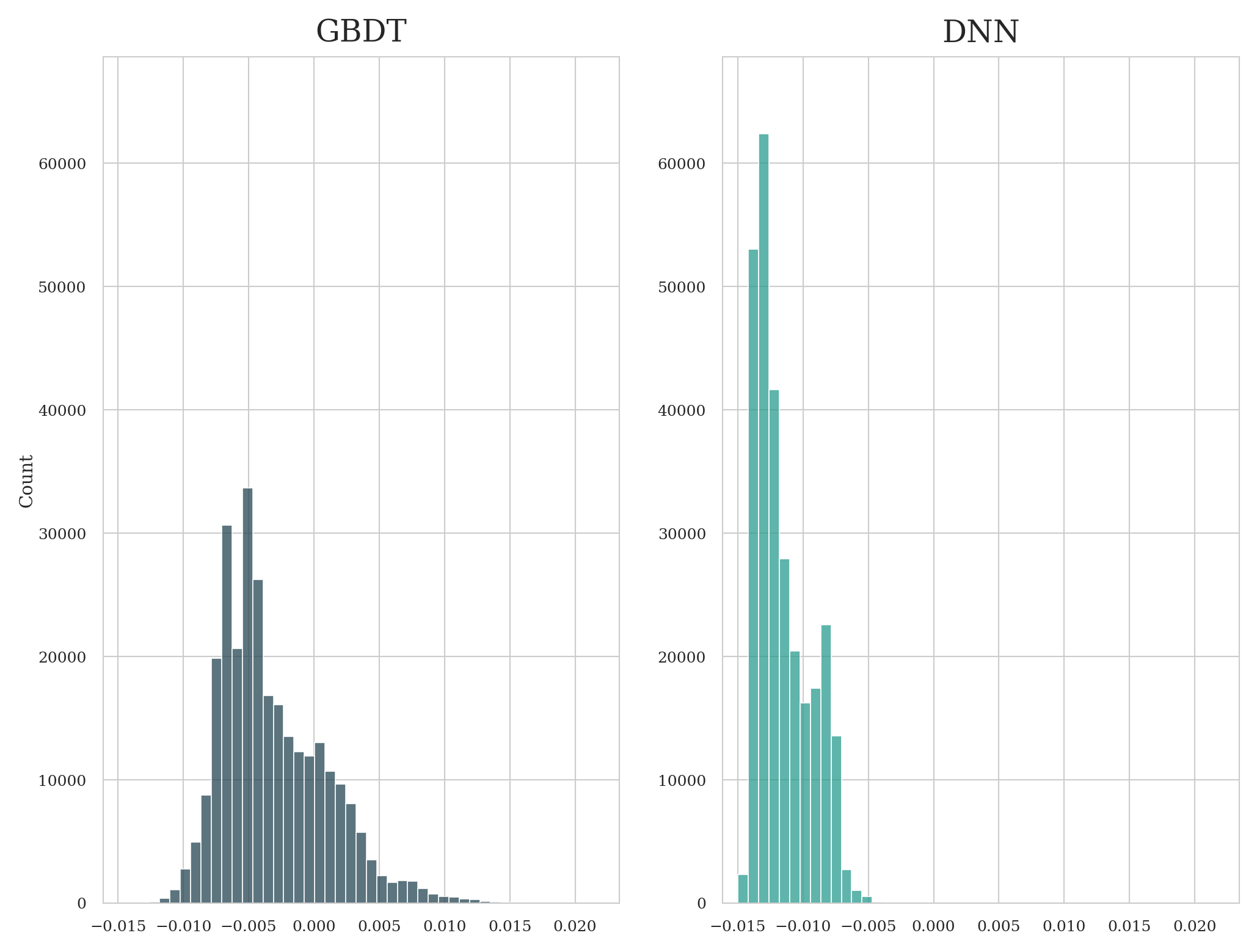}
        \caption{Max grip strength - models with functional intercept}
    \end{subfigure}
    \caption{Functional slopes learnt with GBDT and DNN for the max grip strength variable for FS-GBDT and FS-DNN (a) and FIS-GBDT and FIS-DNN (b).}
    \label{functional slopes for max grip strength}
\end{figure}

We can also visualise histograms of the functional intercept and slopes of the six models having functional effects. We show the functional intercept in Figure \ref{functional intercept}, and we select the functional slopes of the max grip strength variable to analyse in Figure \ref{functional slopes for max grip strength}. Other figures are in \ref{app:functional effects full}. We observe that all functional intercepts learnt with GBDT are unimodal, mostly normally distributed, whereas the functional intercepts learnt with DNN seem bimodal. It is interesting to see how the functional intercept learnt with DNN has a slightly bigger standard deviation than the one learnt with GBDT when the model is also trained with functional slopes. We also observe that all distributions have a positive mean. For the max grip strength functional slopes distribution, we mostly observe an extreme value distribution with a negative mean for all models. For most individuals, an increase in max grip strength is negatively correlated with a higher number of depressive symptoms. It is similar to what was observed for the models without functional slopes, but with more nuances, since we can observe that some individuals have a positive slope, meaning that an increase in the max grip strength is positively correlated with more depressive symptoms.

\section{Conclusion} \label{conclusion}
In this paper, we have adapted and applied existing methodologies to panel data, effectively learning functional effects from socio-demographic characteristics. This allows for modelling inter-heterogeneity in an interpretable ML framework, outperforming out-of-the-box and interpretable ML choice models that would not account for preferences. In addition, we have extensively benchmarked 8 out of the 11 possible functional effects models, showing how they can account for preferences of unobserved individuals, a limitation of traditional choice models such as the fixed and mixed effects models. By doing so, we make the implicit assumption that the choices and tastes of individuals are inherently linked with their socio-demographic characteristics. We note that it is important to review the socio-demographic characteristics included in the model to stay on ethical grounds and be sure that no specific strata of the population are being discriminated against. Interestingly, our approach captures the relationship of individuals at a given point in time. This means that socio-demographic characteristics can change, also adapting the functional effects that these individuals are experiencing. This is likely to be the case for real-life situations, where, for example, having a higher monthly income can change the life experience and, therefore, unobserved factors influencing the decision process. In addition, we have applied the methodology to an ordinal choice problem and multiclass classification problems, but the methodology is not problem-specific and can easily be applied to regression tasks and more complex choice models. 
Further work includes applying eXplainable Artificial Intelligence (XAI) techniques to the functional effects distributions to map heterogeneity to socio-demographics, e.g., certain demographic groups having preferences to alternatives or being more time sensitive, and extending the parametric ML approach to dynamic effects for panel data.

\section*{CRediT authorship contribution statement}

\textbf{Nicolas Salvadé}: Writing – original draft, Visualization, Software, Methodology, Conceptualization, Formal analysis, Investigation. \textbf{Tim Hillel}: Writing – review \& editing, Supervision, Methodology, Conceptualization, Project administration.

\section*{Declaration of competing interest}
The authors declare that they have no known competing financial interests or personal relationships that could have appeared to influence the work reported in this paper.

\section*{Acknowledgments}
This research is supported by a Chadwick PhD Scholarship awarded to Nicolas Salvadé by the UCL Department of Civil, Environmental, and Geomatic Engineering.

This paper uses data from SHARE Waves 1, 2, 3, 4, 5, 6, 7, 8 and 9  (DOIs:  10.6103/SHARE.w1.900, 10.6103/SHARE.w2.900, 10.6103/SHARE.w3.900, 10.6103/SHARE.w4.900, 10.6103/SHARE.w5.900, 10.6103/SHARE.w6.900, 10.6103/SHARE.w6.DBS.100, 10.6103/SHARE.w7.900, 10.6103/SHARE.w8.900, 10.6103/SHARE.w8ca.900, 10.6103/SHARE.w9.900, 10.6103/SHARE.w9ca900, 10.6103/SHARE.HCAP1.100) see \cite{borsch2013data} for methodological details.(1)
The SHARE data collection has been funded by the European Commission, DG RTD through FP5 (QLK6-CT-2001-00360), FP6 (SHARE-I3: RII-CT-2006-062193, COMPARE: CIT5-CT-2005-028857, SHARELIFE: CIT4-CT-2006-028812), FP7 (SHARE-PREP: GA N°211909, SHARE-LEAP: GA N°227822, SHARE M4: GA N°261982, DASISH: GA N°283646) and Horizon 2020 (SHARE-DEV3: GA N°676536, SHARE-COHESION: GA N°870628, SERISS: GA N°654221, SSHOC: GA N°823782, SHARE-COVID19: GA N°101015924) and by DG Employment, Social Affairs \& Inclusion through VS 2015/0195, VS 2016/0135, VS 2018/0285, VS 2019/0332, VS 2020/0313, SHARE-EUCOV: GA N°101052589 and EUCOVII: GA N°101102412. Additional funding from the German Federal Ministry of Education and Research (01UW1301, 01UW1801, 01UW2202), the Max Planck Society for the Advancement of Science, the U.S. National Institute on Aging (U01\_AG09740-13S2, P01\_AG005842, P01\_AG08291, P30\_AG12815, R21\_AG025169, Y1-AG-4553-01, IAG\_BSR06-11, OGHA\_04-064, BSR12-04, R01\_AG052527-02, R01\_AG056329-02, R01\_AG063944, HHSN271201300071C, RAG052527A) and from various national funding sources is gratefully acknowledged (see www.share-eric.eu).
This paper uses data from the generated easySHARE data set (DOI: 10.6103/SHARE.easy.900), see \citep{gruber2014generating} for methodological details. The easySHARE release 8.8.0 is based on SHARE Waves 1, 2, 3, 4, 5, 6, 7,8 and 9 (DOIs:      10.6103/SHARE.w1.900, 10.6103/SHARE.w2.900, 10.6103/SHARE.w3.900, 10.6103/SHARE.w4.900, 10.6103/SHARE.w5.900, 10.6103/SHARE.w6.900,     10.6103/SHARE.w7.900, 10.6103/SHARE.w8.900, 10.6103/SHARE.w9.900).

\appendix
\setcounter{table}{0}
\setcounter{figure}{0}

\section{Ordinal data - CORAL}\label{app:ordinal}

The CORAL methodology, introduced in \cite{shi2023deep}, has been developed to model ordinal target variables. It is a popular methodology in the ML community and exhibits close similarities with the Ordinal Logit model. CORAL can be derived from a latent regression model, where we have: 

\begin{equation}
    U_n = \hat{y}_n = V_n + \epsilon_n
\end{equation}
where:
\begin{itemize}
    \item $\hat{y}_n$ is the unobserved latent response for an individual $n$ interpreted as the utility function $U_n$. Note that this is a simple regression value, being the same for all alternatives;
    \item $V_n$ is the deterministic utility defined in Equation \ref{deterministic utility}; and
    \item $\epsilon_n$ is the error term, capturing unobserved information.
\end{itemize}

The continuous space of the deterministic utility can be separated into $J$ categories with $J-1$ thresholds $\tau_j$ such that the predicted dependent variable $y'_n$ is:

\begin{equation}
    y'_n = 
    \begin{cases}
        1, & \text{if $\hat{y}_n \leq \tau_1$,} \\
        2, & \text{if $\tau_1 < \hat{y}_n \leq \tau_2$,} \\
        \vdots \\
        J, & \text{if $\hat{y}_n > \tau_{J-1}$} \\
    \end{cases}
\end{equation}

If we assume that the error term follows a standard logistic distribution, the probability that the latent scalar output of the model is greater than a threshold $\tau_j$ is the sigmoid function:

\begin{equation}
    P_{jn} = P(\hat{y}_n > \tau_j) = \sigma(V_n-\tau_j) = \frac{1}{1 + \exp^{-(V_n-\tau_j)}} \quad \forall j = 1, \dots, J-1
\end{equation}

Then, the probability of choosing a class can be reconstructed from the binary decomposition, i.e.:

\begin{equation}
    P(\hat{y}_n = 1) = 1 - P(\hat{y}_n > \tau_1)
\end{equation}
\begin{equation}
    P(\hat{y}_n = j) = P(\hat{y}_n > \tau_{j-1}) - P(\hat{y}_n > \tau_j) \quad \forall j = 2, \dots, J-1 \\    
\end{equation}
\begin{equation}
    P(\hat{y}_n = J) = P(\hat{y}_n > \tau_{J-1})
\end{equation}

In CORAL, and this is where the methodology differs from a traditional Ordinal Logit model, the thresholds and parameters are chosen to maximise the Multi-label Cross-Entropy Loss (MCEL) function of the binary sub-problems (as opposed to the negative CEL function of the probabilities in the Ordinal Logit model):

\begin{equation}
    \mathcal{L} = \sum_{n=1}^{N}\sum_{j=1}^{J-1} \mathbbm{1}(j > y_n)\ln(P(\hat{y}_n > \tau_j)) + (1-\mathbbm{1}(j > y_n))\ln(1-P(\hat{y}_n > \tau_j))
\end{equation}

where:
\begin{equation}
    \mathbbm{1}(j>y_n) = 
    \begin{cases}
        1, & \text{if $j>y_n$ the chosen class,} \\
        0, & \text{otherwise}
    \end{cases}
\end{equation}

This equation has the advantage of including all thresholds in the loss function, therefore also penalising thresholds that are not directly precedent or subsequent to the chosen class.

For model evaluation, in addition to the MCEL, we also report two additional metrics:
\begin{enumerate}
    \item the \textit{mean absolute error} (MAE) a discrete metric; and
    \item and the \textit{expected mean absolute error} (EMAE), a probabilistic metric which considers distances between classes.
\end{enumerate}

The MAE is defined as:

\begin{equation}
    MAE = \frac{1}{N}\sum_{n=1}^{N}|y_n - l_n|
\end{equation}
and the EMAE as:

\begin{equation}
    EMAE = \frac{1}{N}\sum_{n=1}^{N}\sum_{j=1}^{J} P(\hat{y}_n = j) \cdot (j - y_n)^2
\end{equation}

where:
\begin{itemize}
    \item $N$ is the number of observations;
    \item $y_n$ is the observed choice for individual $n$;
    \item $P(\hat{y}_n = j)$ is the predicted probability that individual $n$ will choose alternative $j$; and
    \item $l_n = \sum_{j=1}^{J-1}\mathbbm{1}{(\hat{y}_n > \tau_j)}$ is the discrete class prediction of the model for individual $n$.
\end{itemize}

\FloatBarrier
\section{Functional effects}\label{app:functional effects full}

Figure \ref{fig:ind_spec_const_all} shows all functional slopes for the models of Section \ref{case study} with the easySHARE dataset.

\begin{figure}[htbp]
    \centering


    \begin{subfigure}[b]{0.45\textwidth}
        \includegraphics[width=\linewidth]{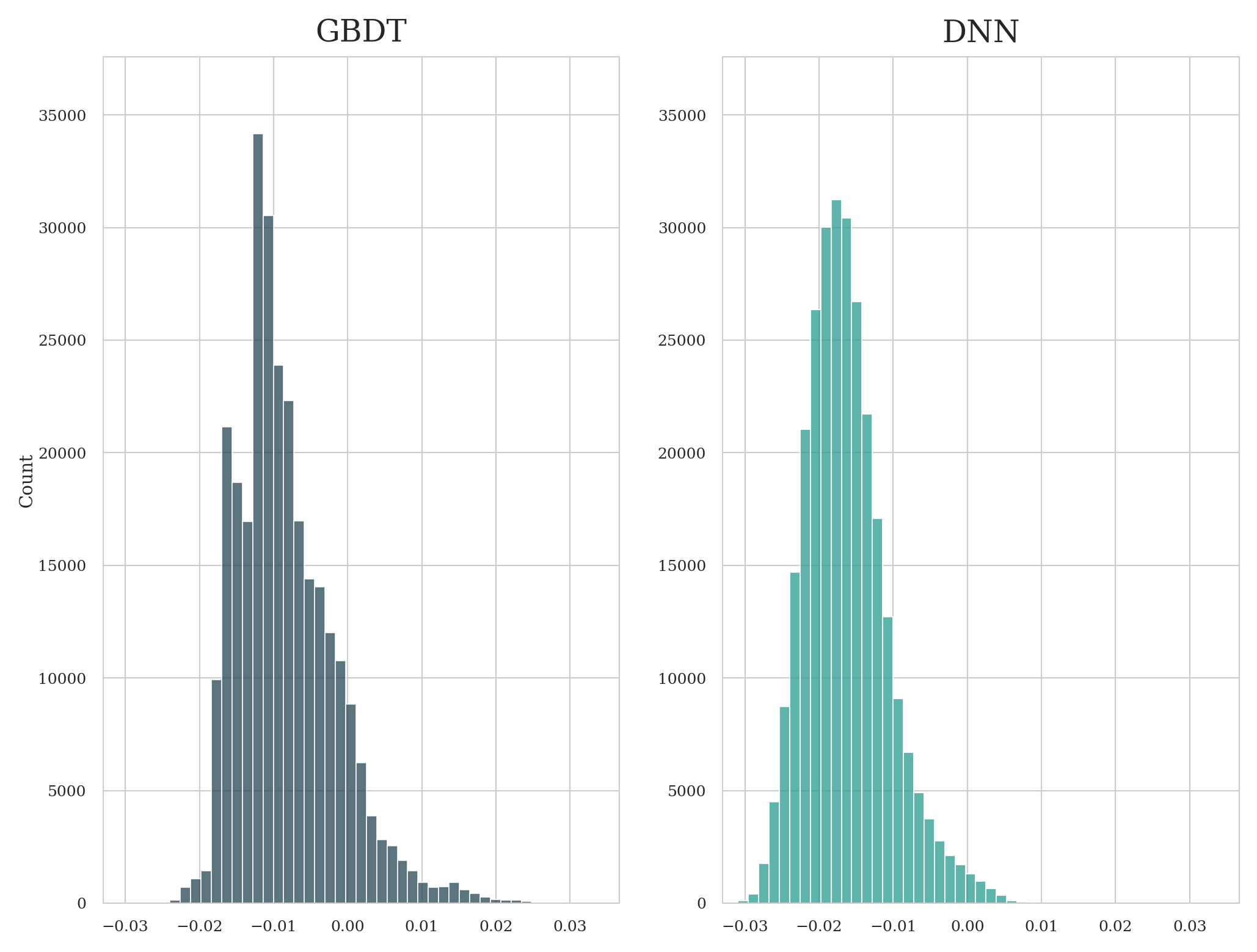}
        \caption{BMI}
    \end{subfigure}
    \begin{subfigure}[b]{0.45\textwidth}
        \includegraphics[width=\linewidth]{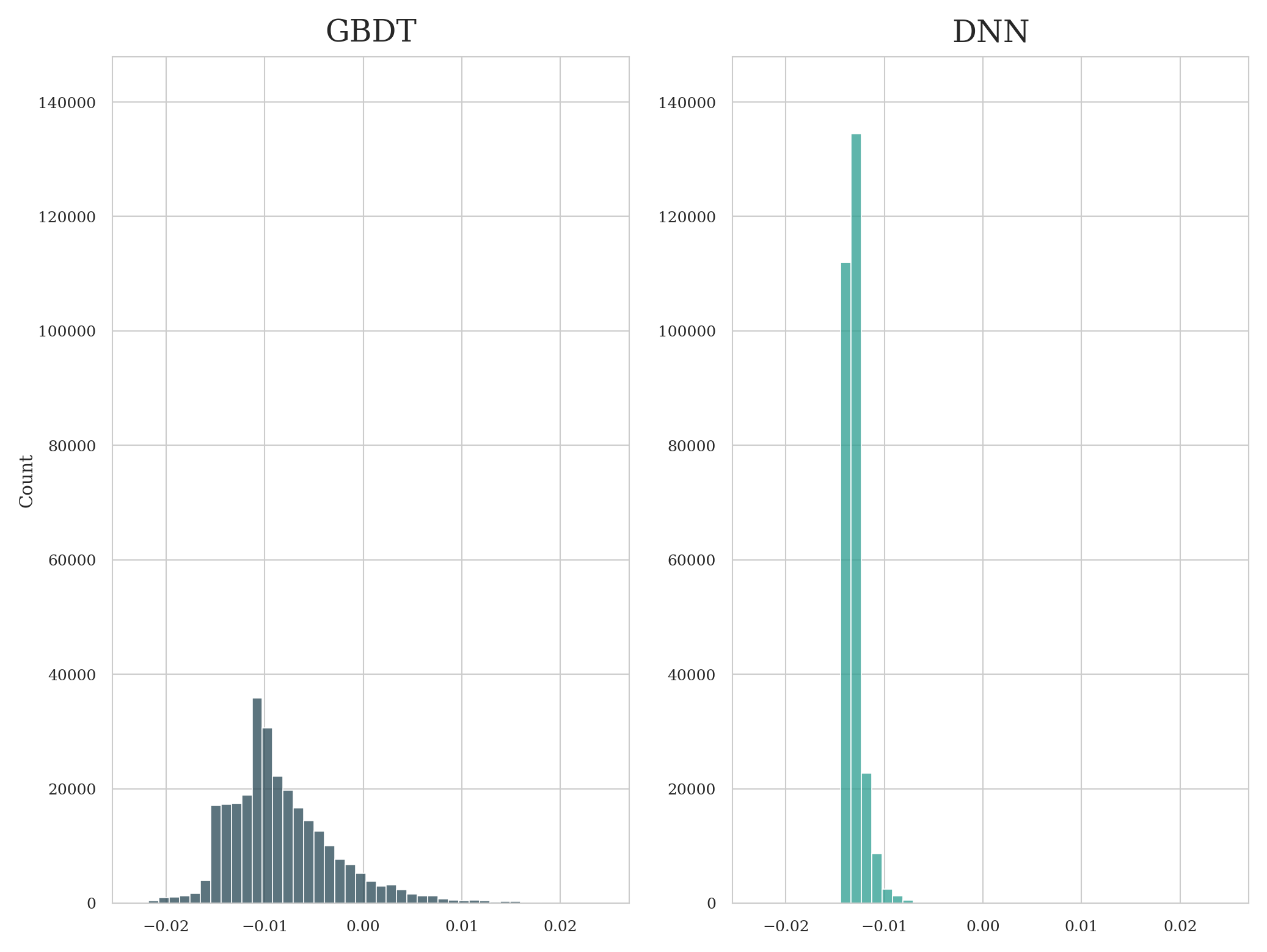}
        \caption{BMI - models with functional intercept}
    \end{subfigure}

    \begin{subfigure}[b]{0.45\textwidth}
        \includegraphics[width=\linewidth]{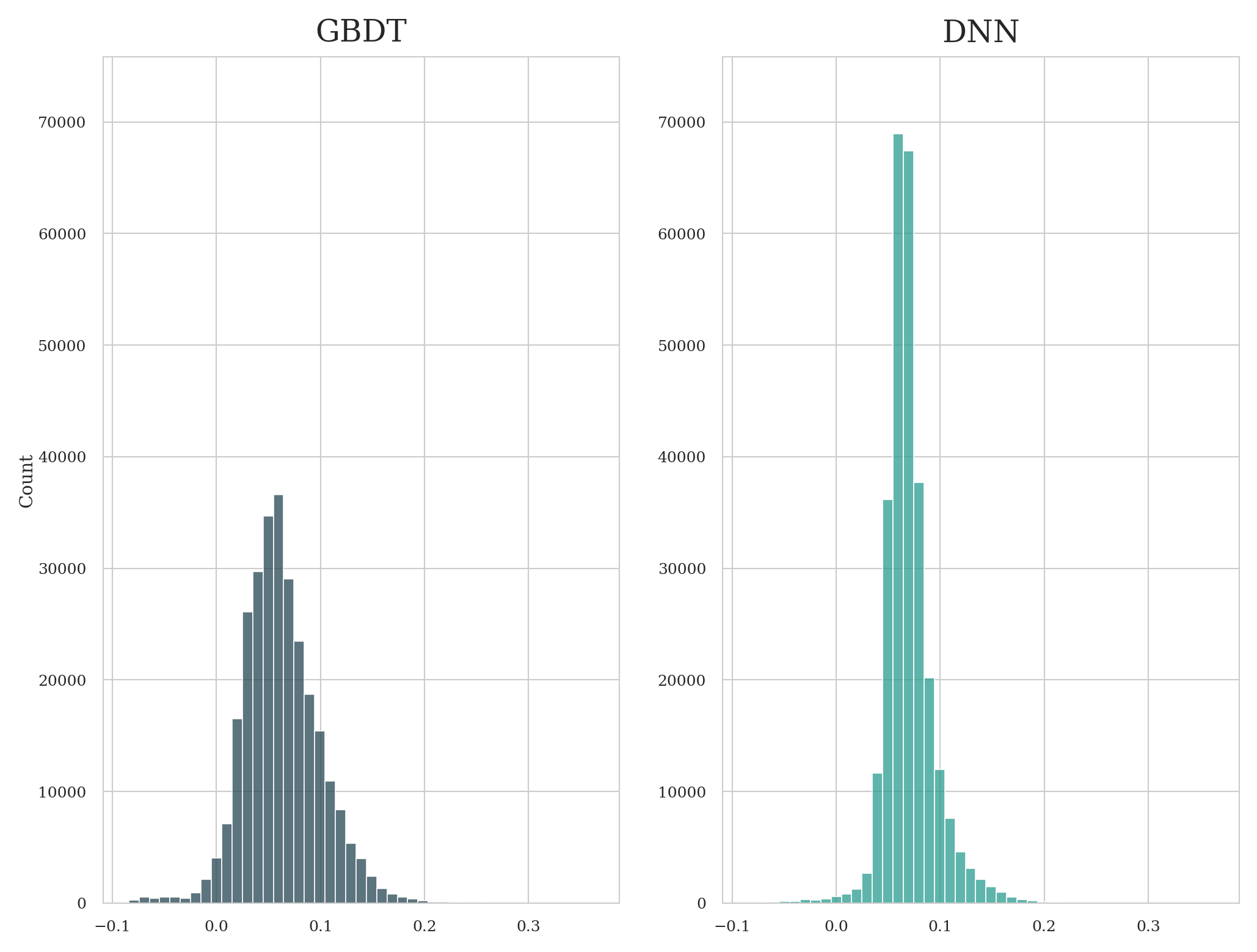}
        \caption{Number of chronic conditions}
    \end{subfigure}
    \begin{subfigure}[b]{0.45\textwidth}
        \includegraphics[width=\linewidth]{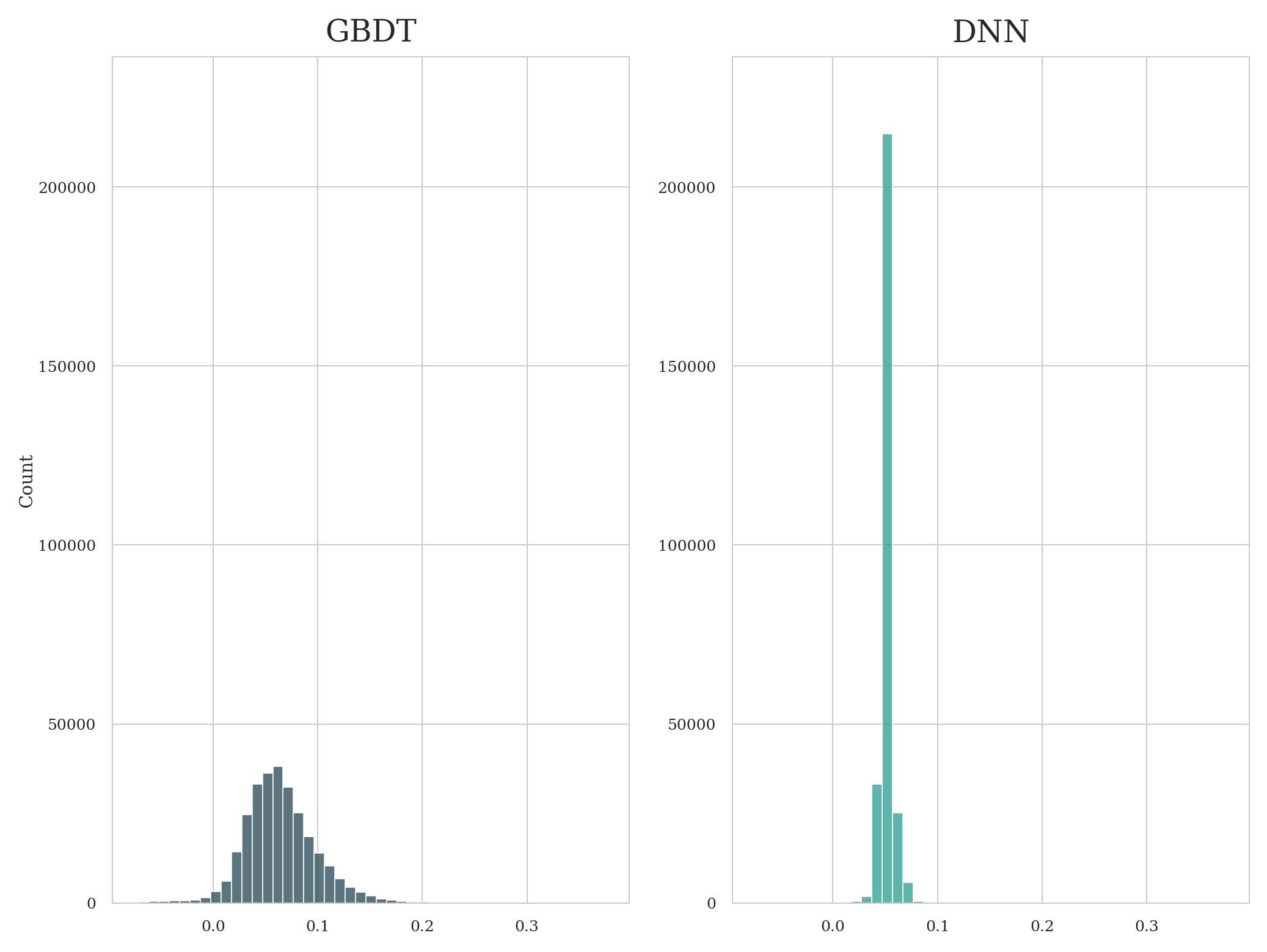}
        \caption{Number of chronic conditions - models with functional intercept}
    \end{subfigure}

    \begin{subfigure}[b]{0.45\textwidth}
        \includegraphics[width=\linewidth]{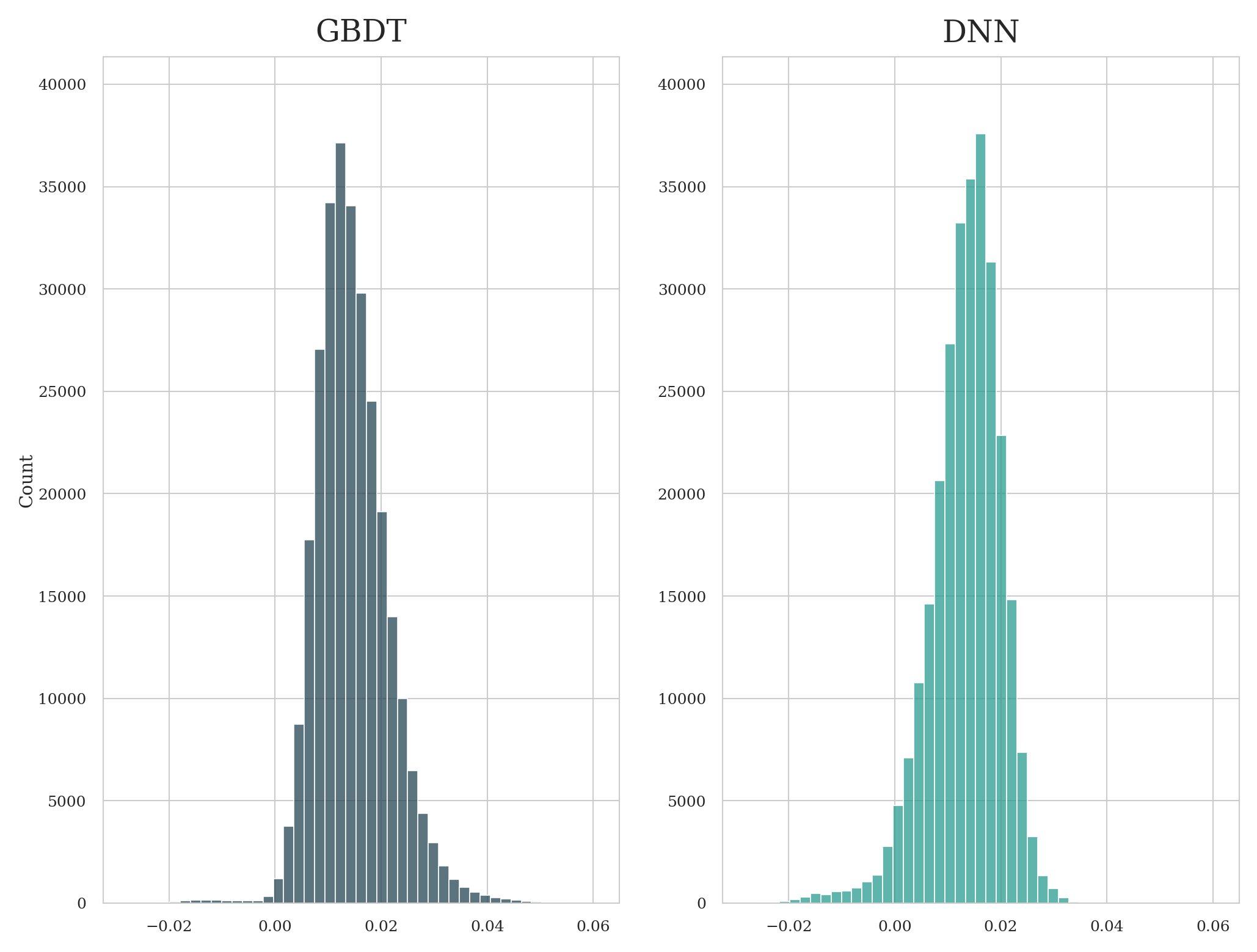}
        \caption{Number of doctor visits}
    \end{subfigure}
    \begin{subfigure}[b]{0.45\textwidth}
        \includegraphics[width=\linewidth]{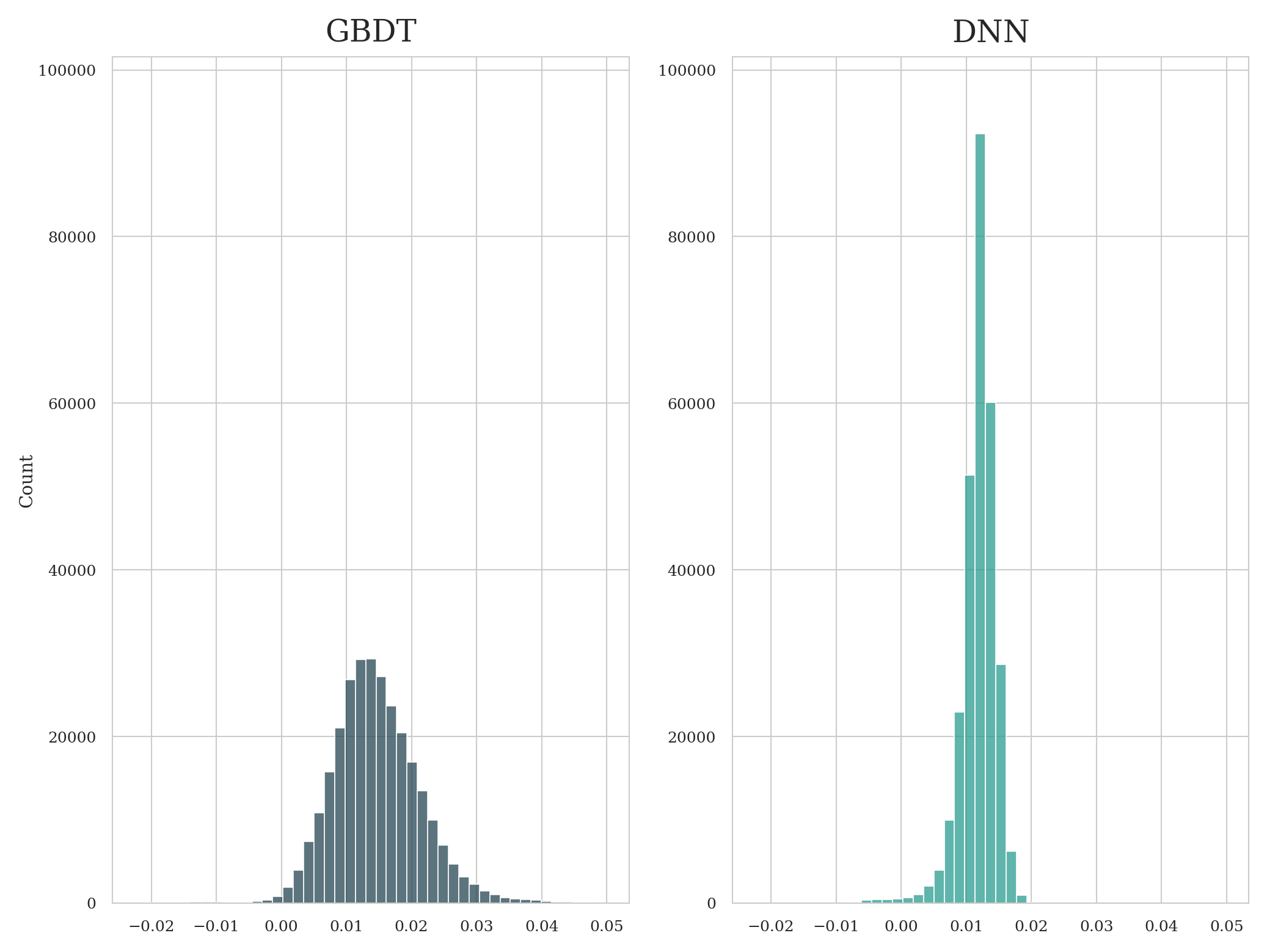}
        \caption{Number of doctor visits - models with functional intercept}
    \end{subfigure}

\end{figure}

\begin{figure}[htbp]\ContinuedFloat

    \begin{subfigure}[b]{0.45\textwidth}
        \includegraphics[width=\linewidth]{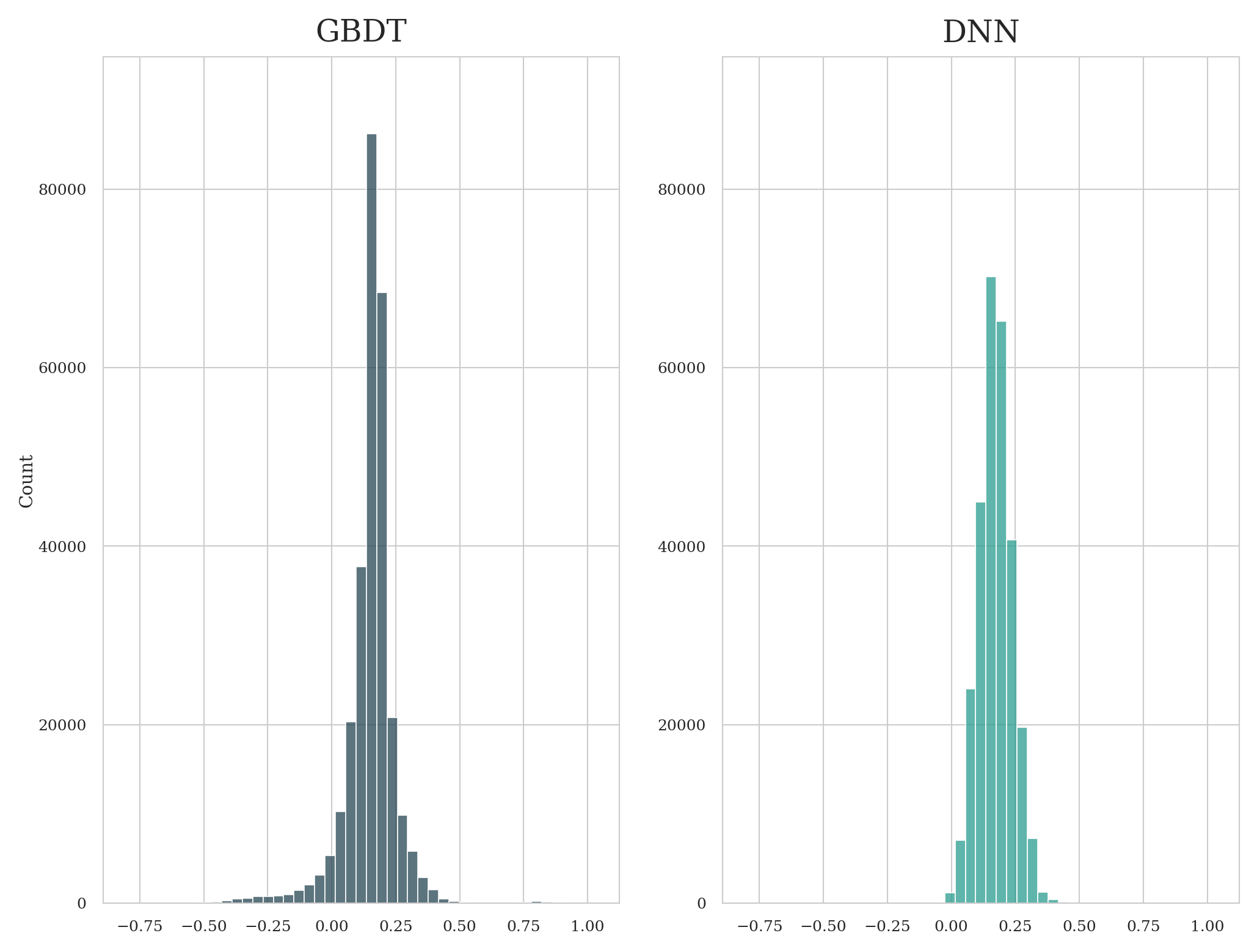}
        \caption{Fine motor skills}
    \end{subfigure}
    \begin{subfigure}[b]{0.45\textwidth}
        \includegraphics[width=\linewidth]{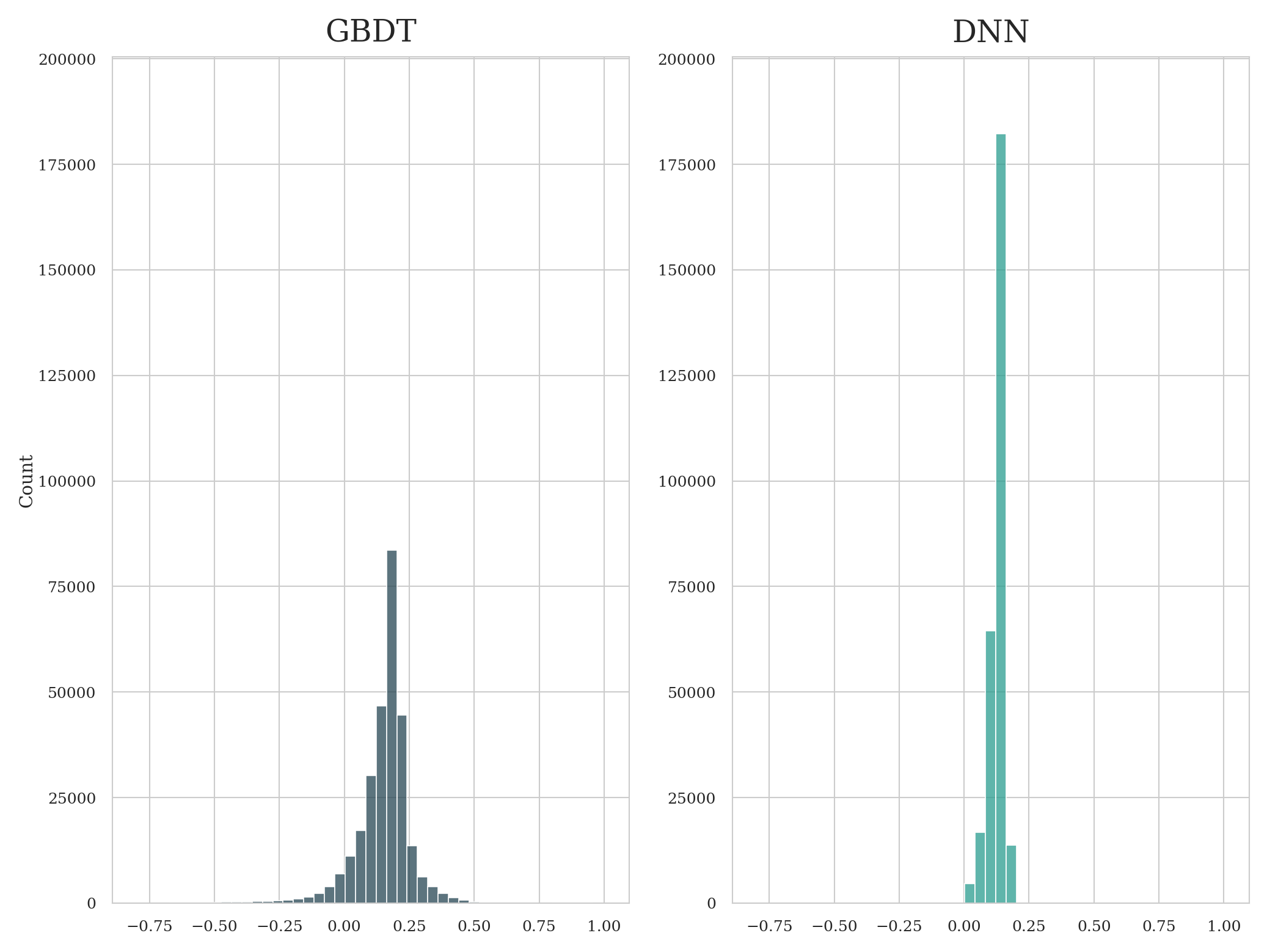}
        \caption{Fine motor skills - models with functional intercept}
    \end{subfigure}

    \begin{subfigure}[b]{0.45\textwidth}
        \includegraphics[width=\linewidth]{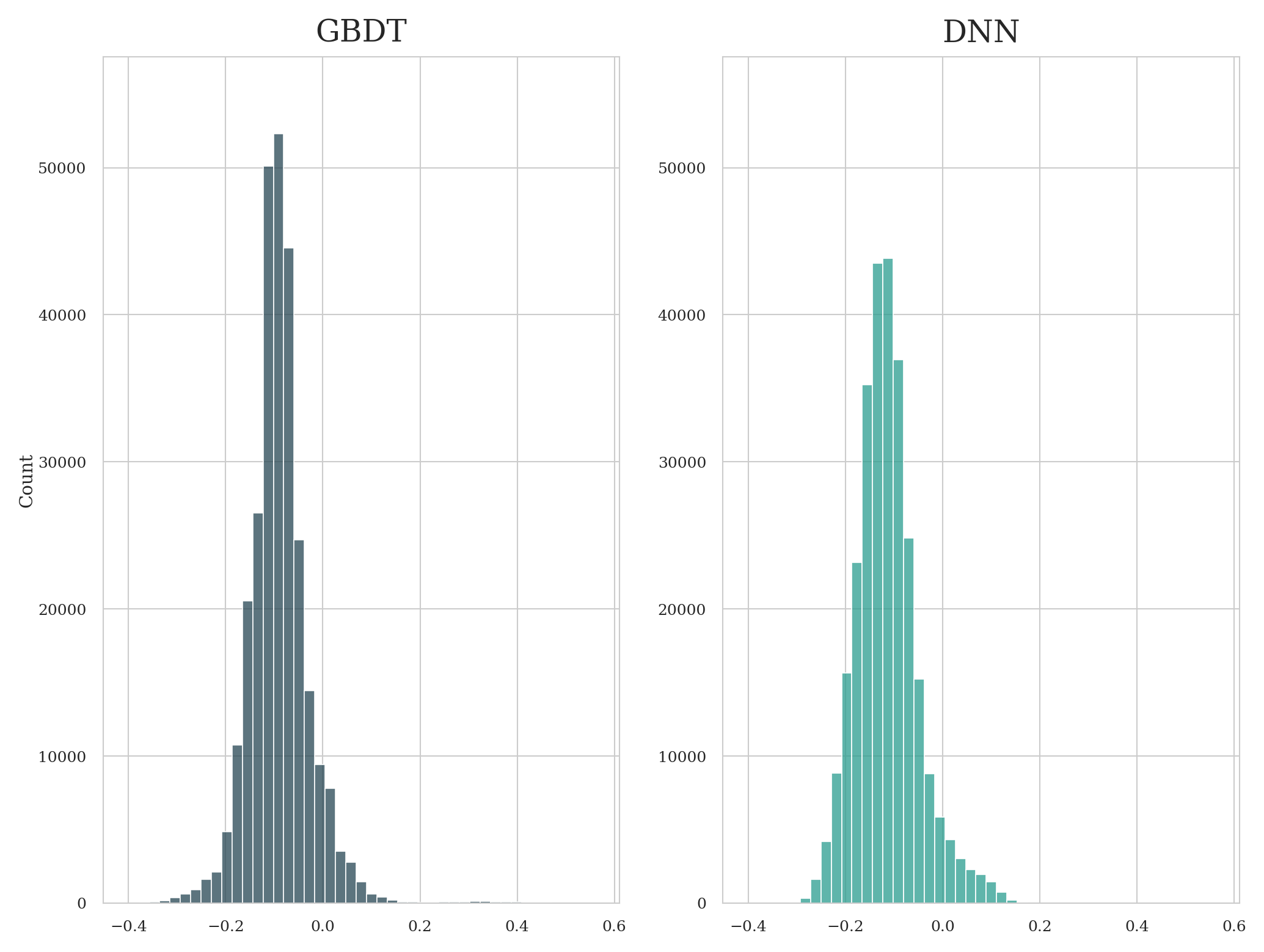}
        \caption{Gross motor skills}
    \end{subfigure}
    \begin{subfigure}[b]{0.45\textwidth}
        \includegraphics[width=\linewidth]{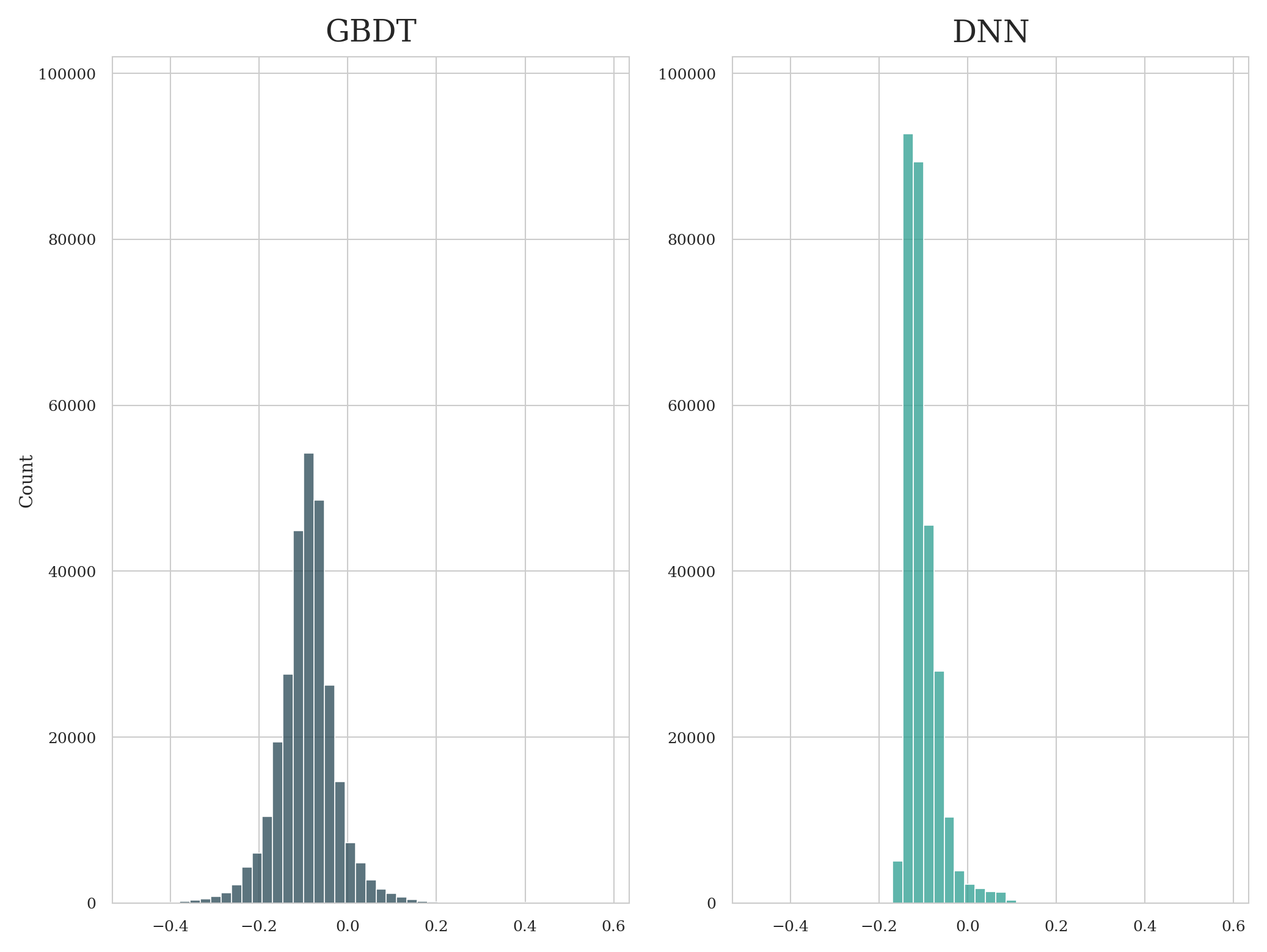}
        \caption{Gross motor skills - models with functional intercept}
    \end{subfigure}

    \begin{subfigure}[b]{0.45\textwidth}
        \includegraphics[width=\linewidth]{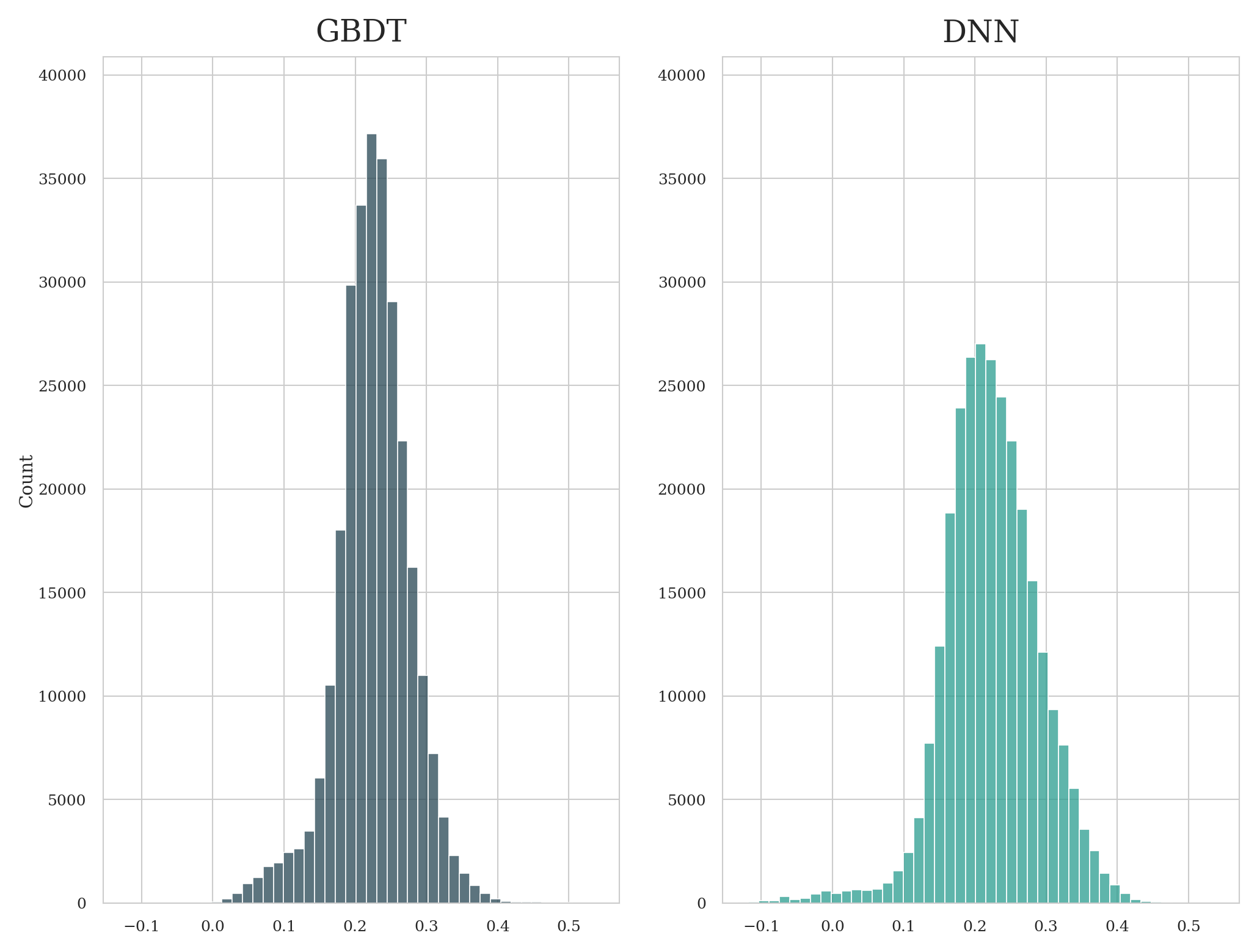}
        \caption{Large muscle skills}
    \end{subfigure}
    \begin{subfigure}[b]{0.45\textwidth}
        \includegraphics[width=\linewidth]{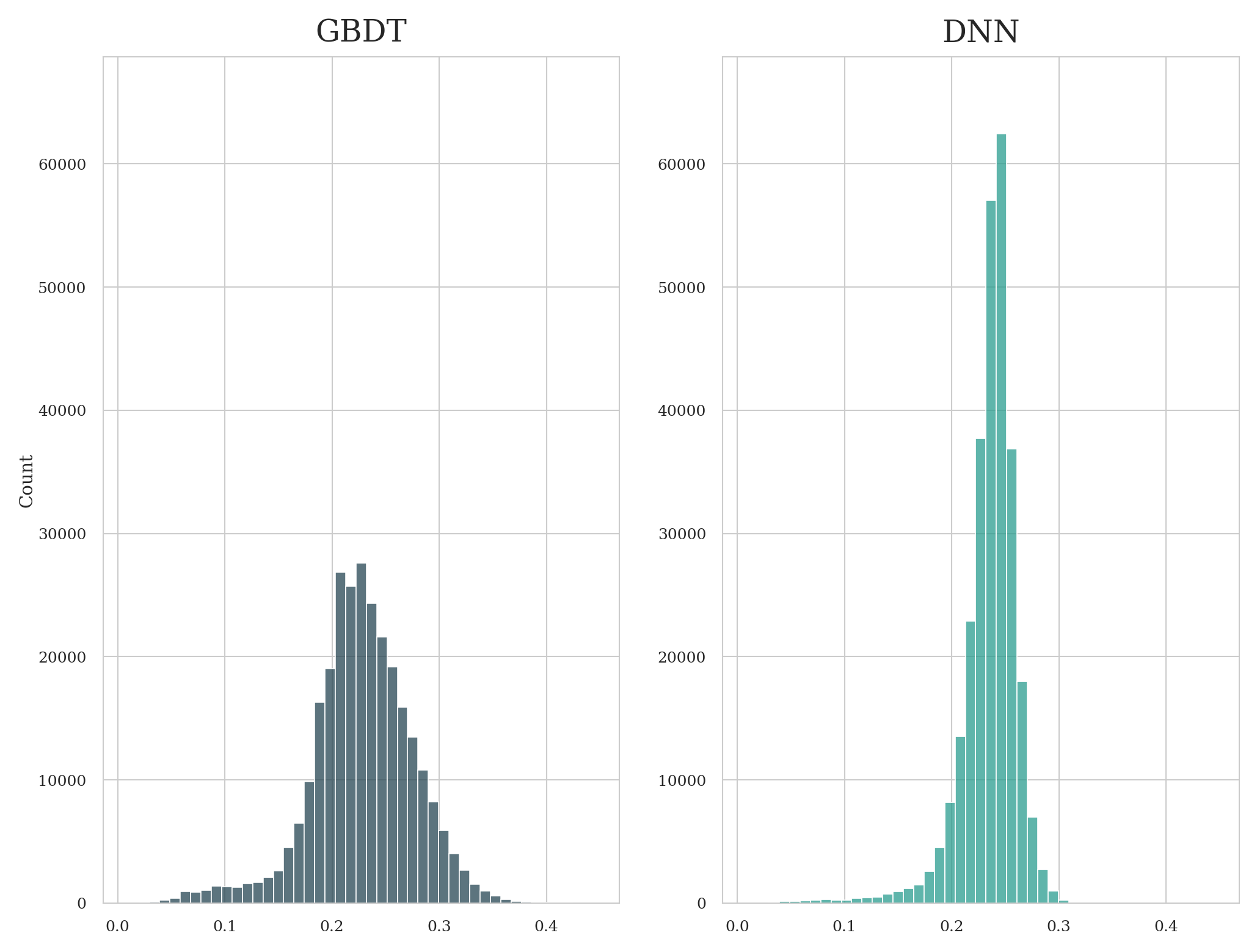}
        \caption{Large muscle skills - models with functional intercept}
    \end{subfigure}

\end{figure}

\begin{figure}[htbp]\ContinuedFloat

    \begin{subfigure}[b]{0.45\textwidth}
        \includegraphics[width=\linewidth]{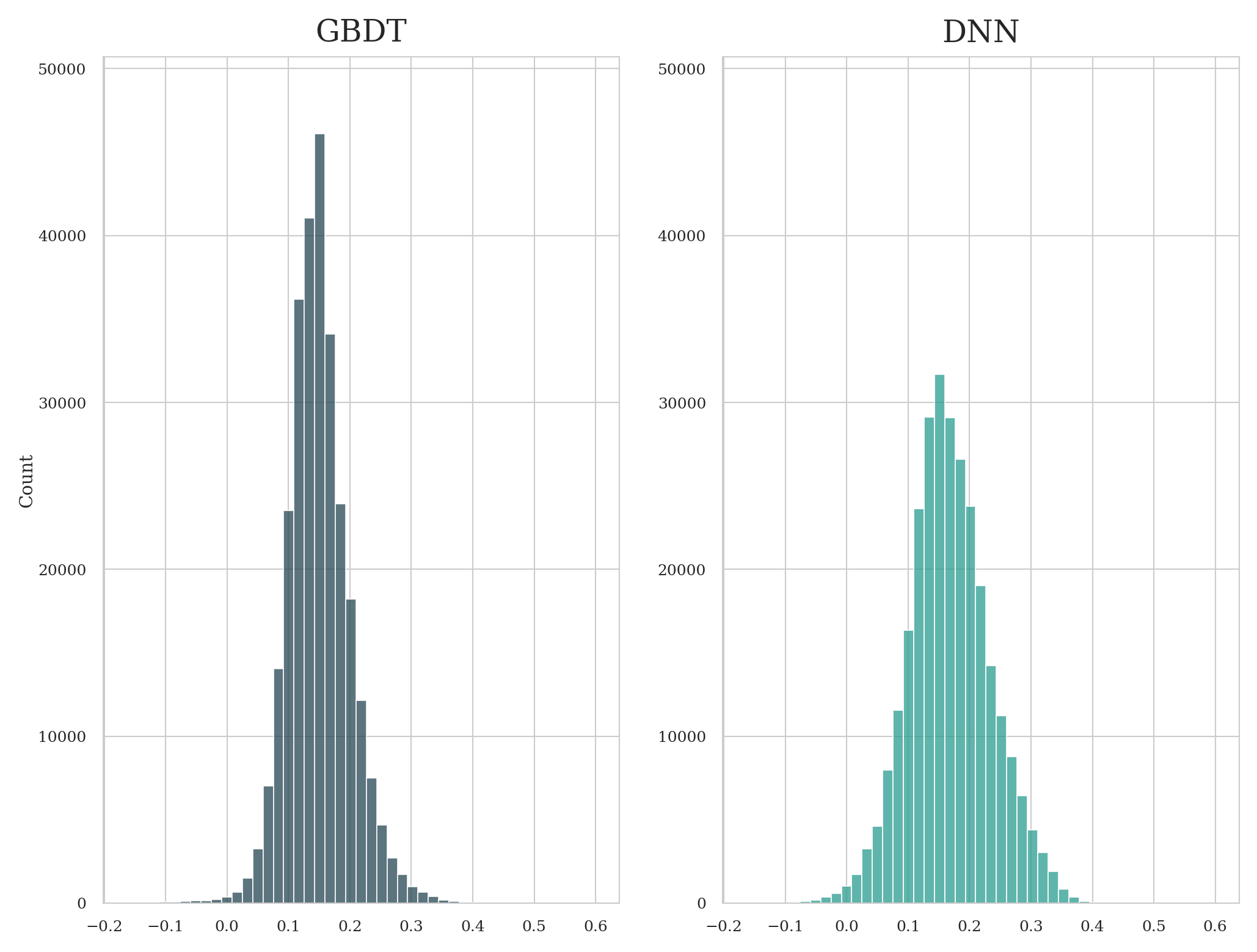}
        \caption{Mobility index}
    \end{subfigure}
    \begin{subfigure}[b]{0.45\textwidth}
        \includegraphics[width=\linewidth]{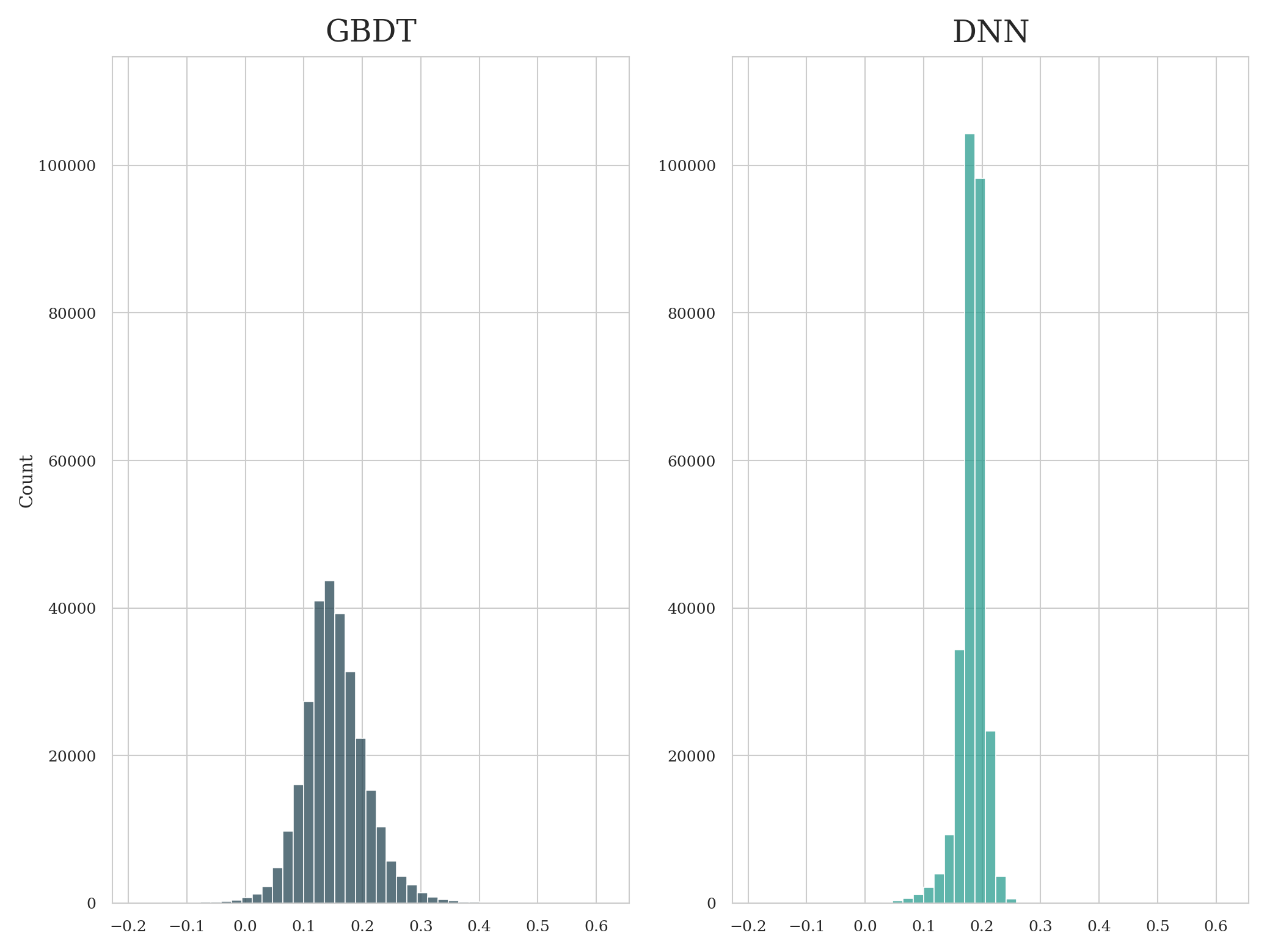}
        \caption{Mobility index - models with functional intercept}
    \end{subfigure}
    
    \begin{subfigure}[b]{0.45\textwidth}
        \includegraphics[width=\linewidth]{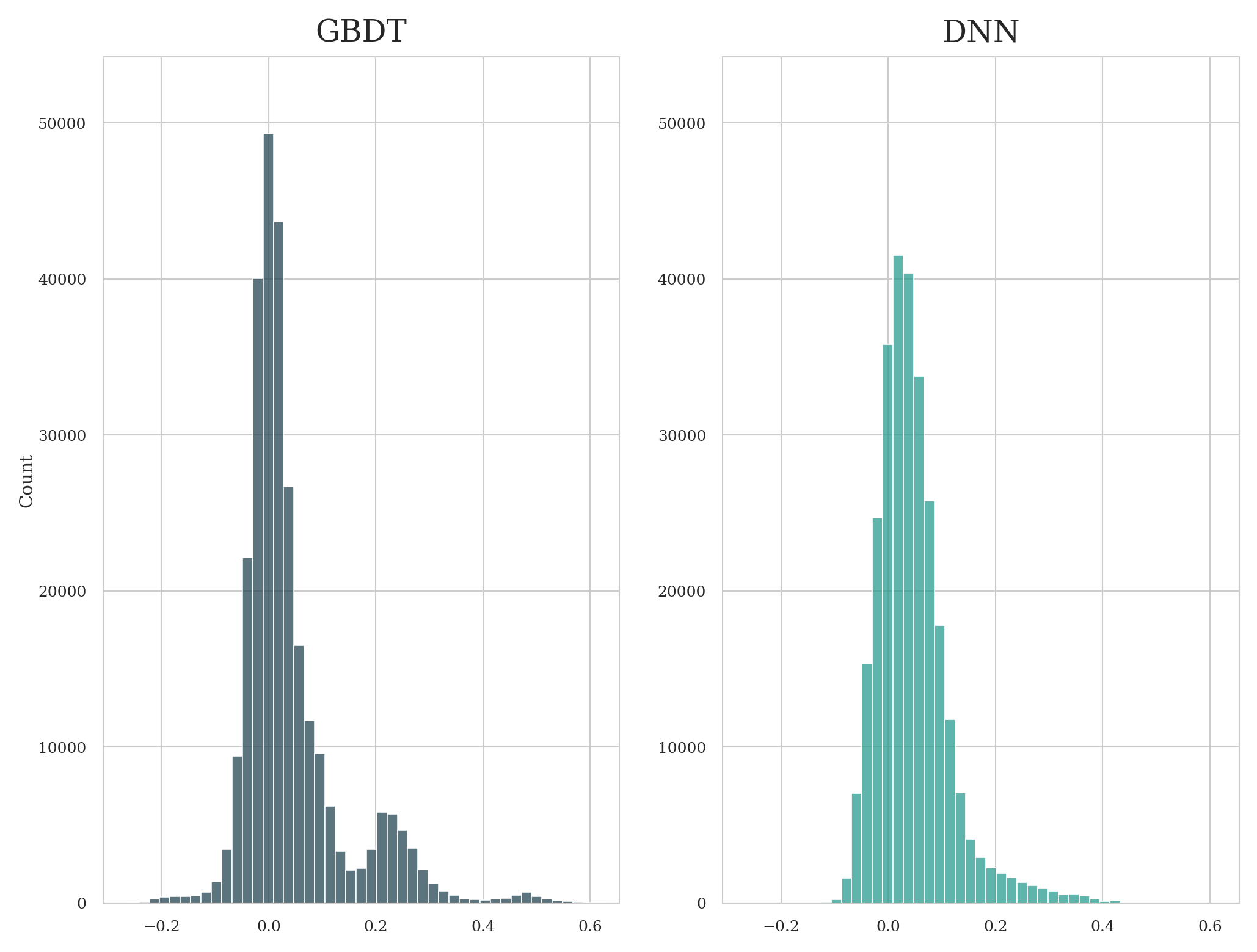}
        \caption{Daily activities index}
    \end{subfigure}
    \begin{subfigure}[b]{0.45\textwidth}
        \includegraphics[width=\linewidth]{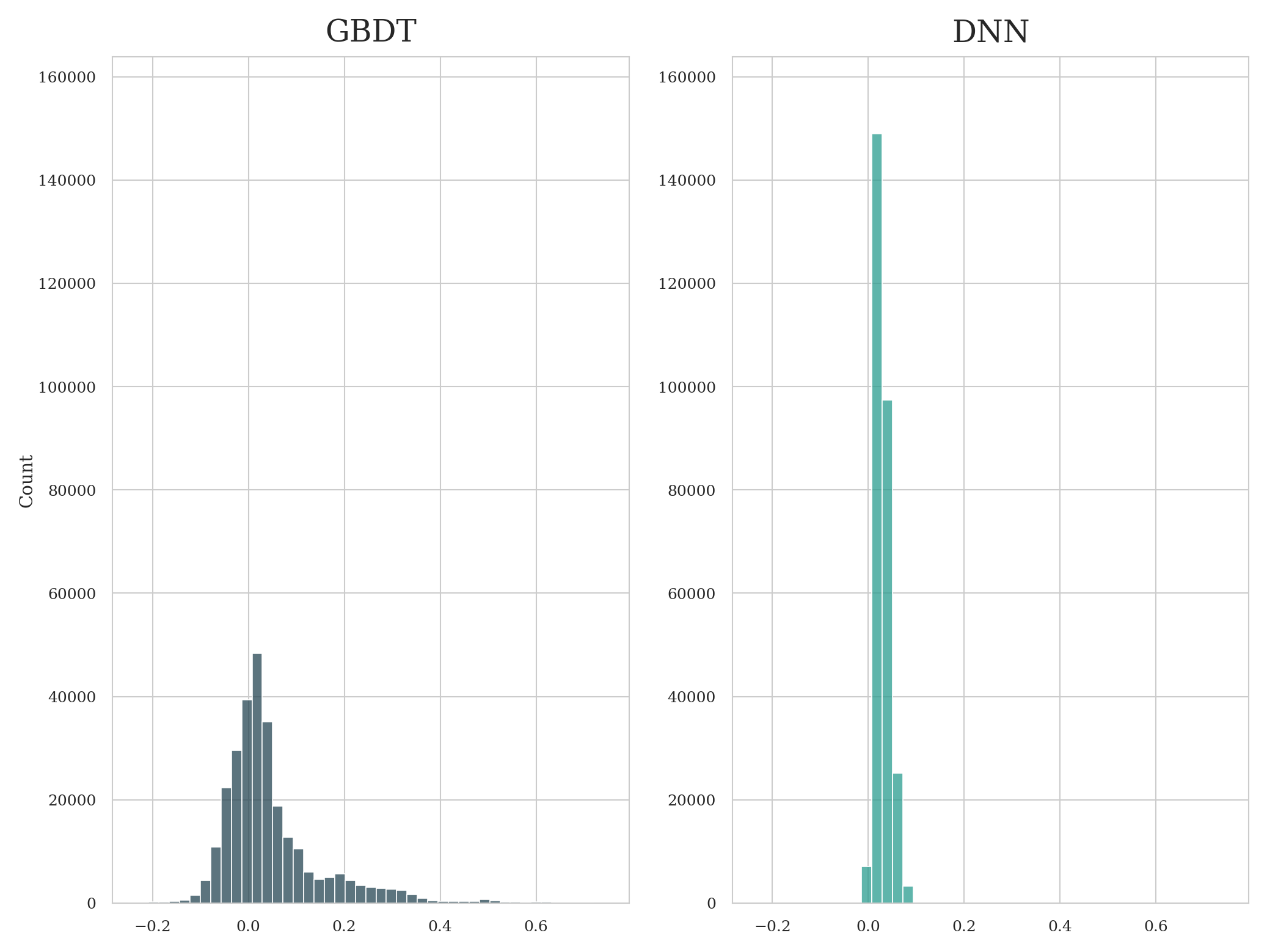}
        \caption{Daily activities index - models with functional intercept}
    \end{subfigure}
    
    \begin{subfigure}[b]{0.45\textwidth}
        \includegraphics[width=\linewidth]{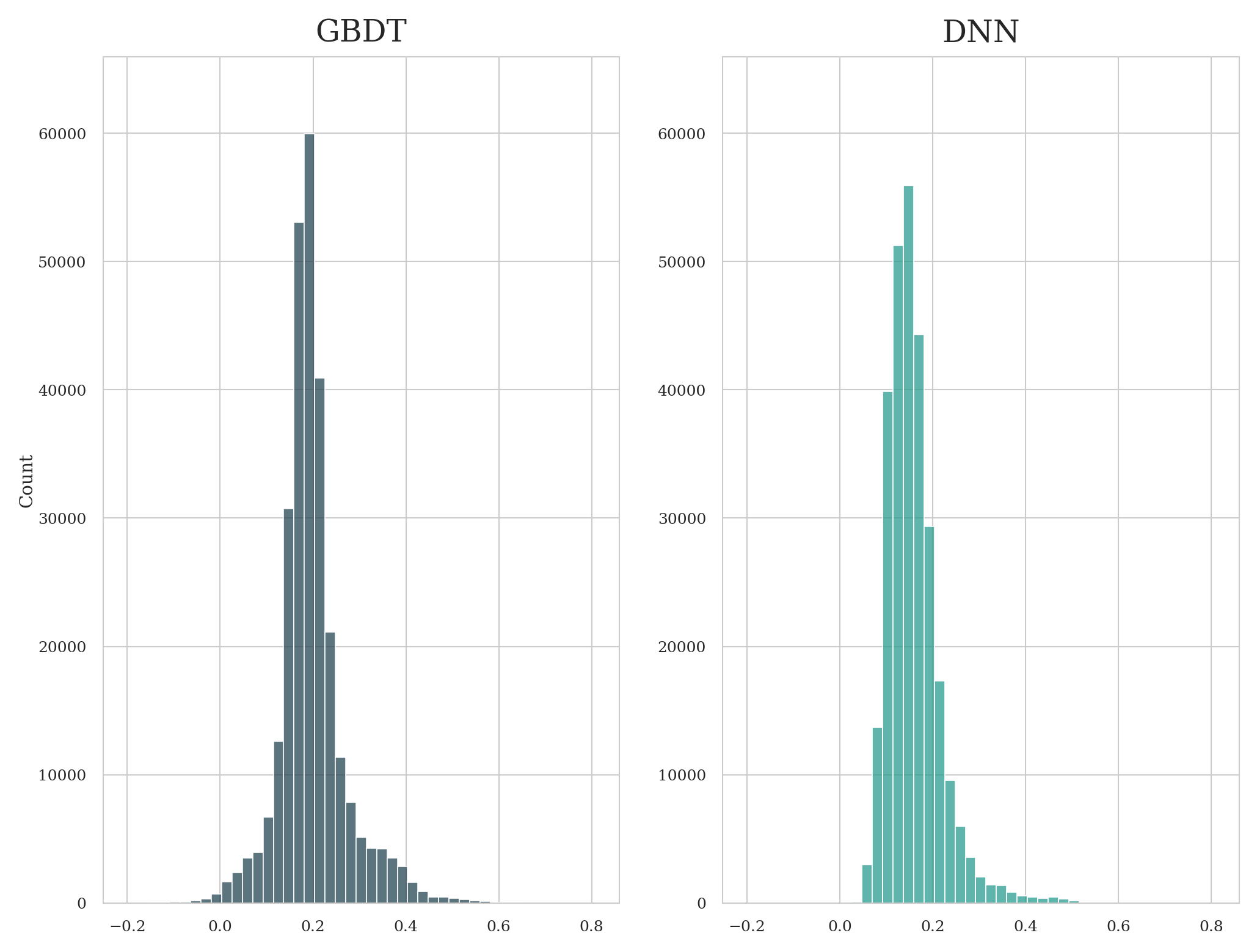}
        \caption{Instrumental activities index}
    \end{subfigure}
    \begin{subfigure}[b]{0.45\textwidth}
        \includegraphics[width=\linewidth]{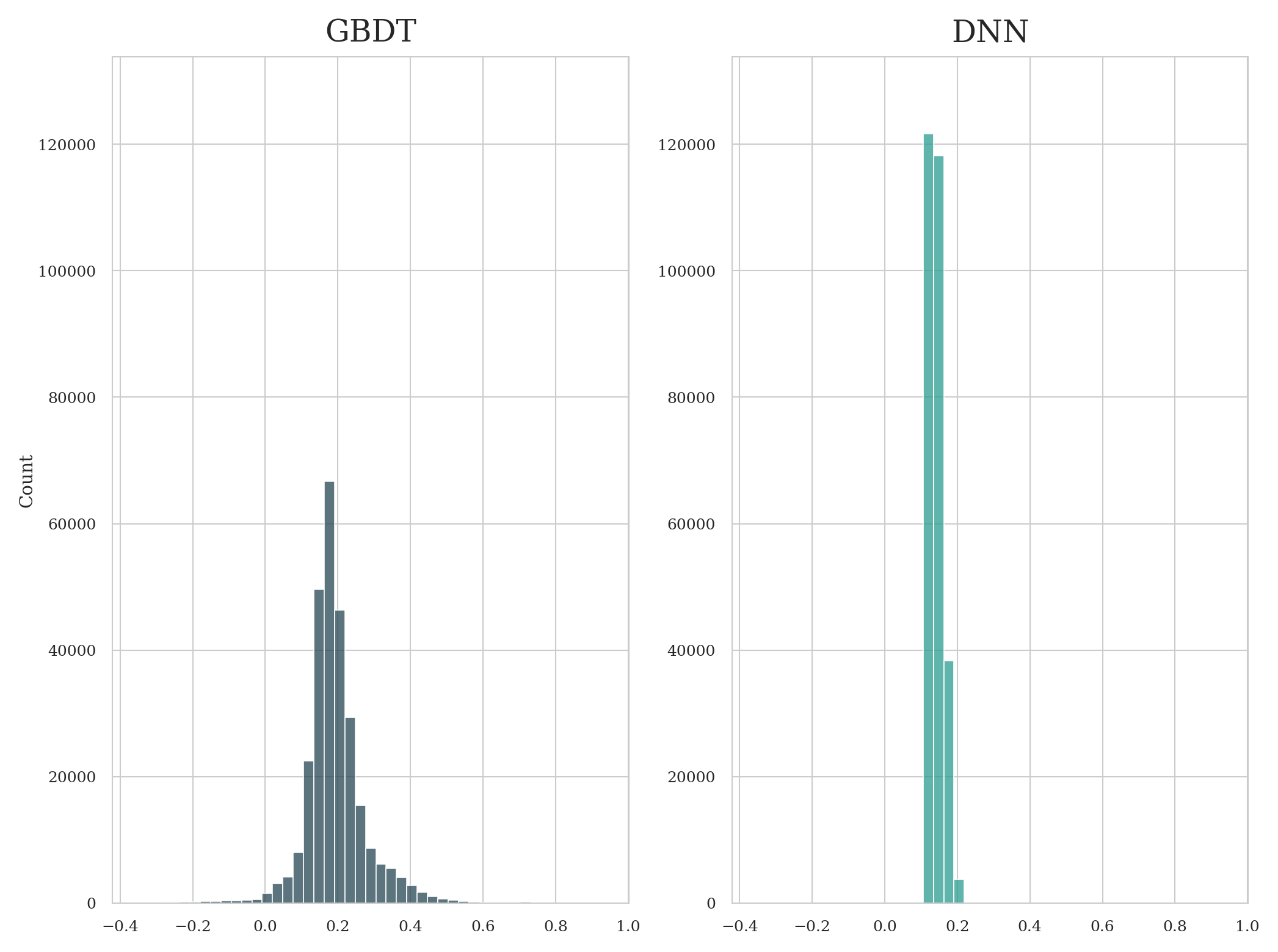}
        \caption{Instrumental activities index - models with functional intercept}
    \end{subfigure}

\end{figure}

\begin{figure}[htbp]\ContinuedFloat

    \begin{subfigure}[b]{0.45\textwidth}
        \includegraphics[width=\linewidth]{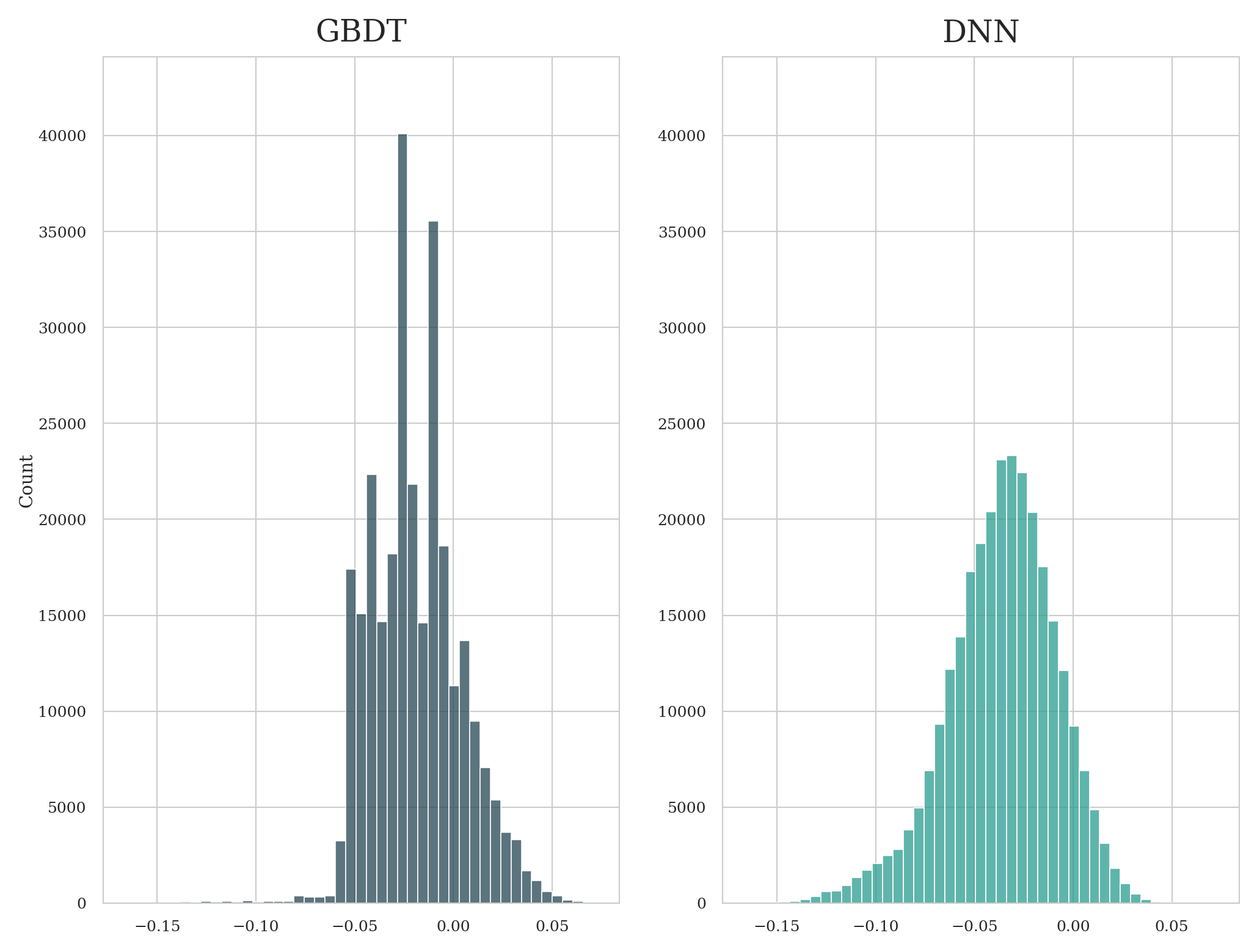}
        \caption{Recall 1}
    \end{subfigure}
    \begin{subfigure}[b]{0.45\textwidth}
        \includegraphics[width=\linewidth]{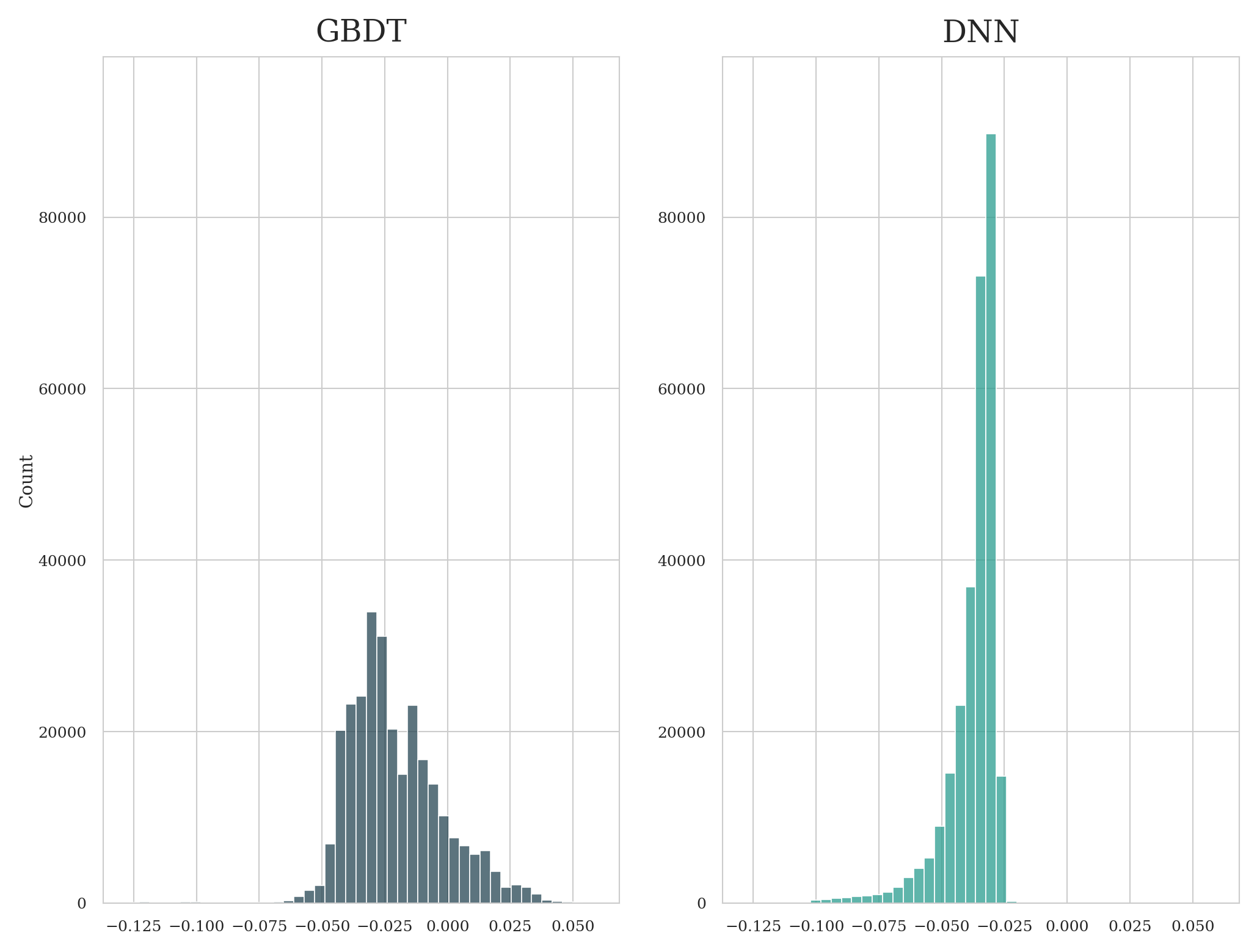}
        \caption{Recall 1 - models with functional intercept}
    \end{subfigure}

    \begin{subfigure}[b]{0.45\textwidth}
        \includegraphics[width=\linewidth]{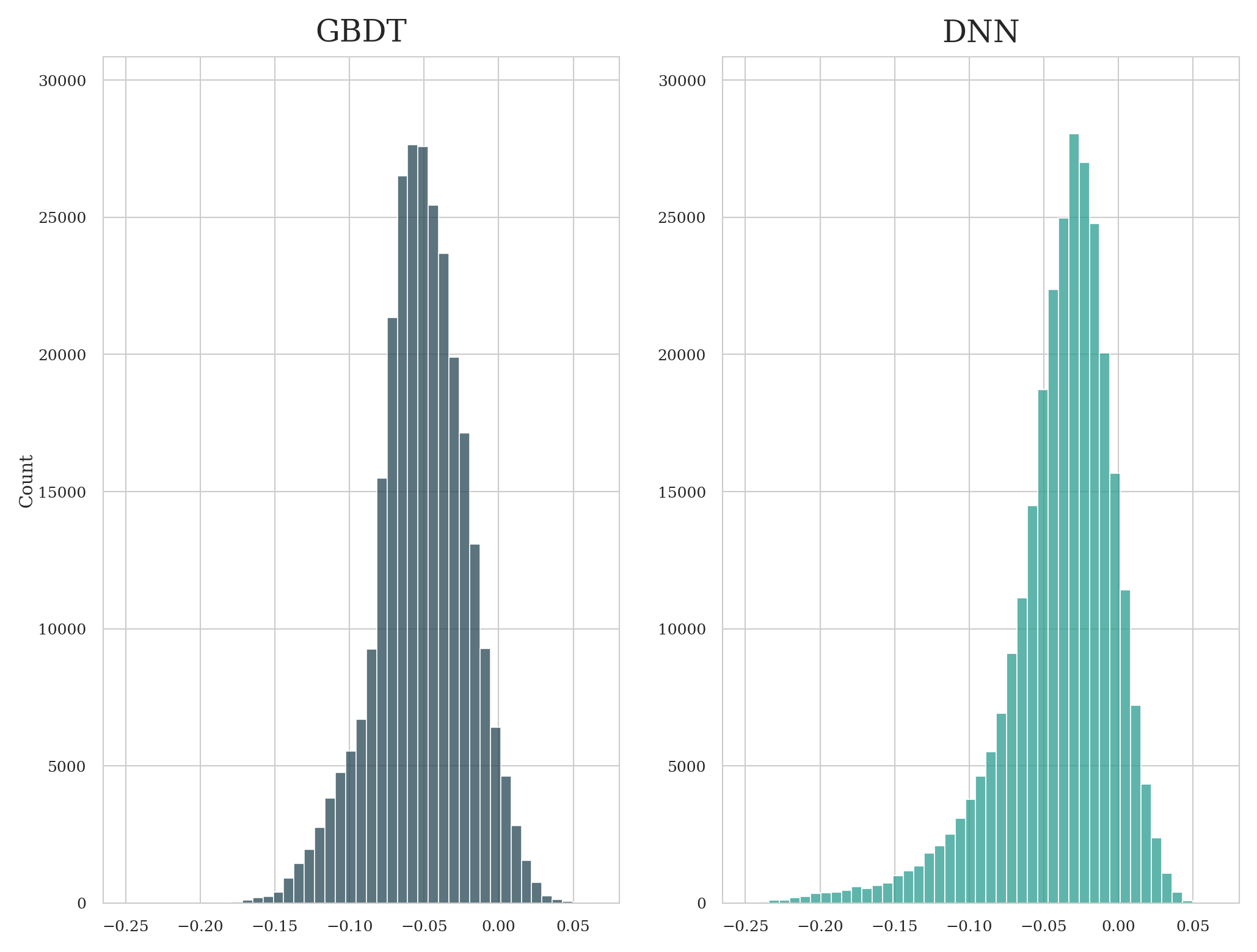}
        \caption{Recall 2}
    \end{subfigure}
    \begin{subfigure}[b]{0.45\textwidth}
        \includegraphics[width=\linewidth]{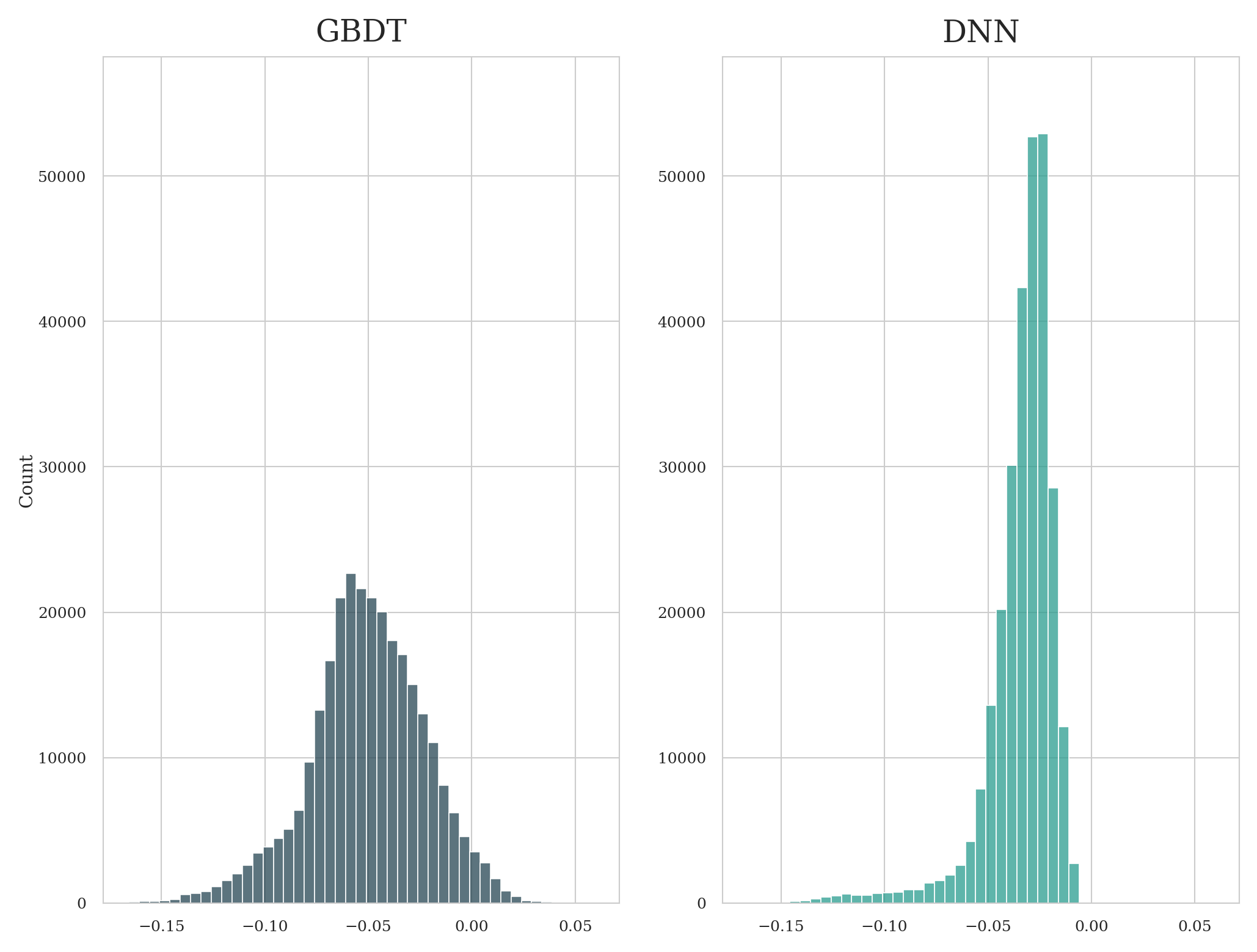}
        \caption{Recall 2 - models with functional intercept}
    \end{subfigure}

\end{figure}

\begin{figure}[htbp]\ContinuedFloat

    \begin{subfigure}[b]{0.45\textwidth}
        \includegraphics[width=\linewidth]{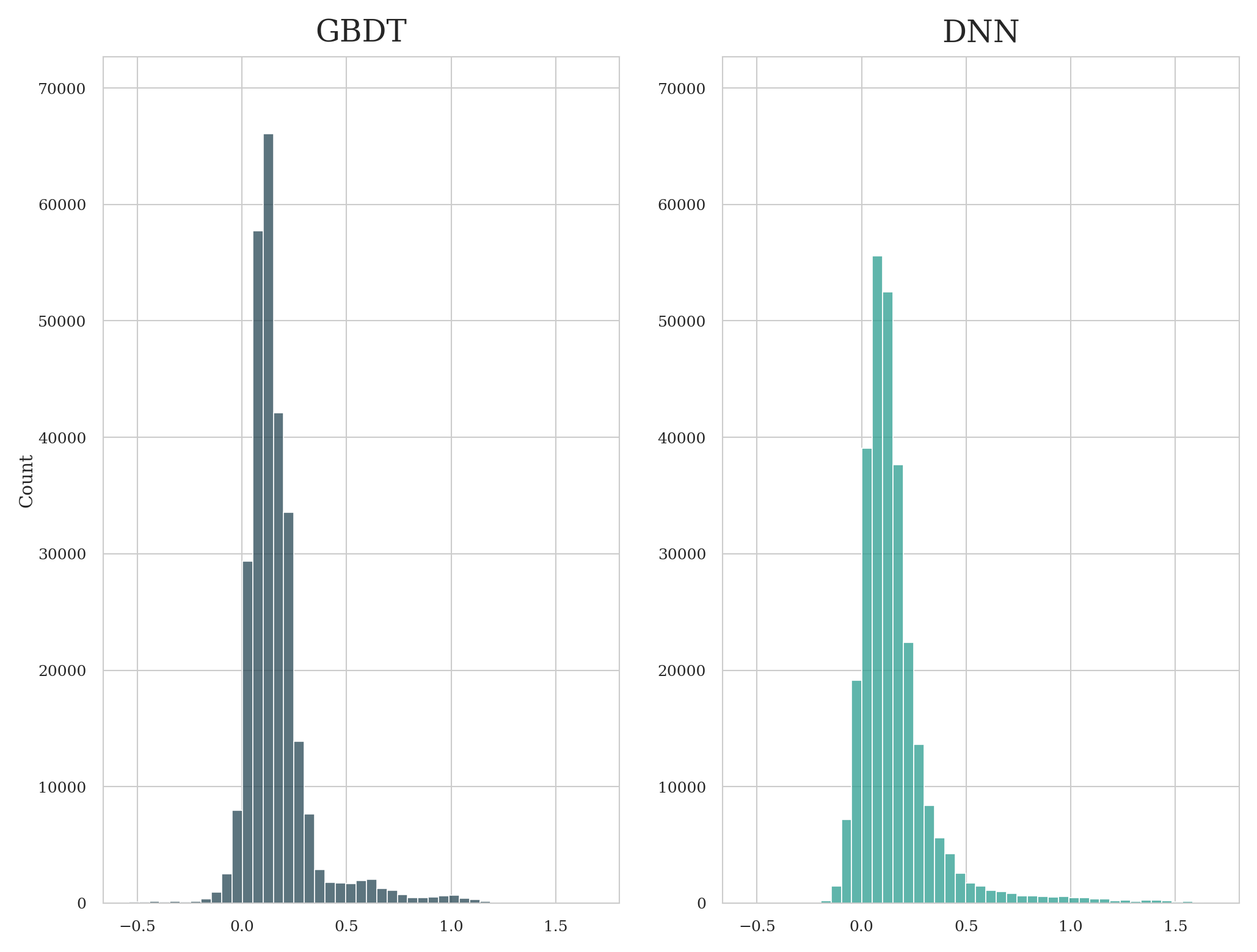}
        \caption{Hospitalised last year}
    \end{subfigure}
    \begin{subfigure}[b]{0.45\textwidth}
        \includegraphics[width=\linewidth]{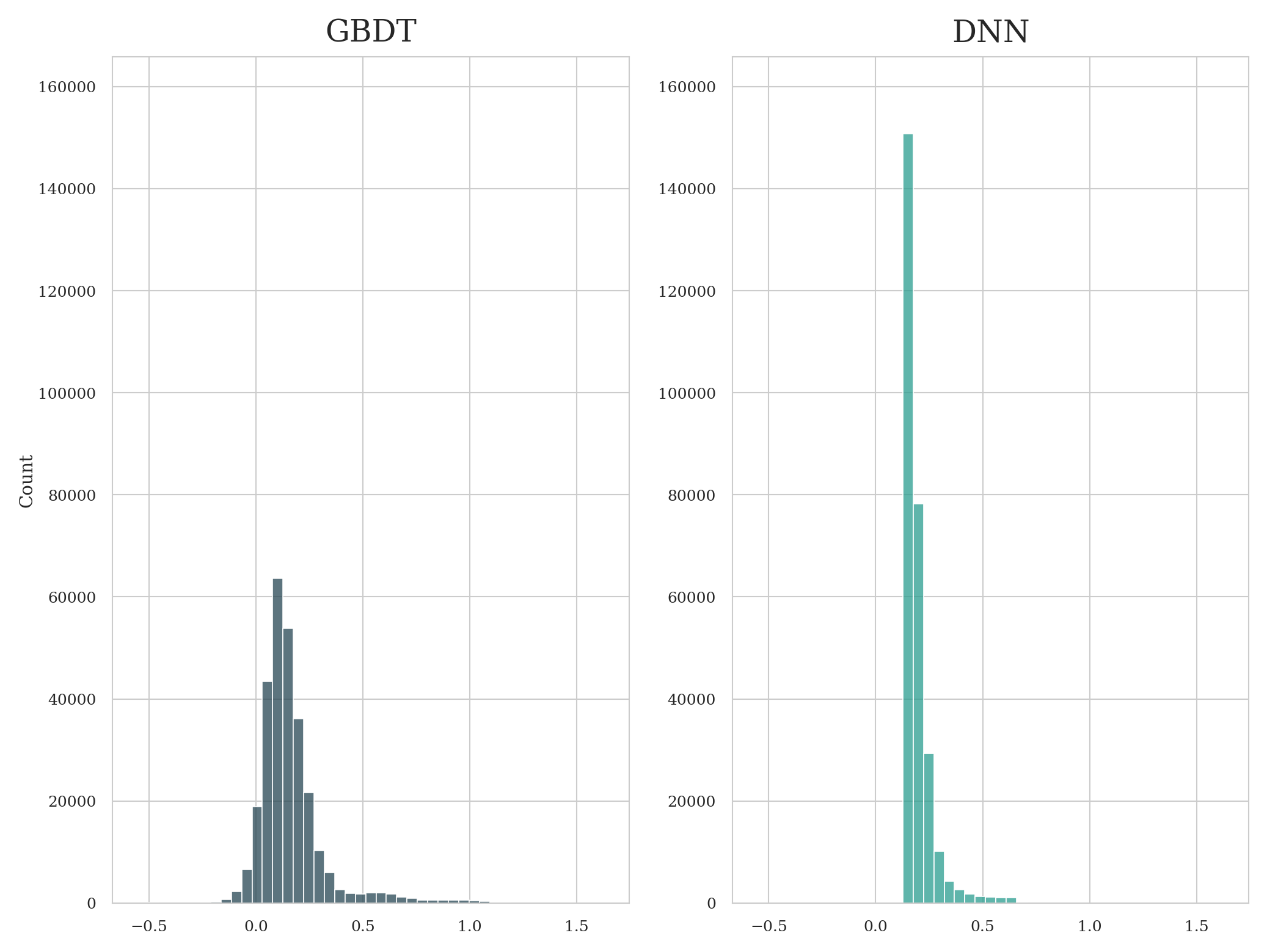}
        \caption{Hospitalised last year - models with functional intercept}
    \end{subfigure}
    
    \begin{subfigure}[b]{0.45\textwidth}
        \includegraphics[width=\linewidth]{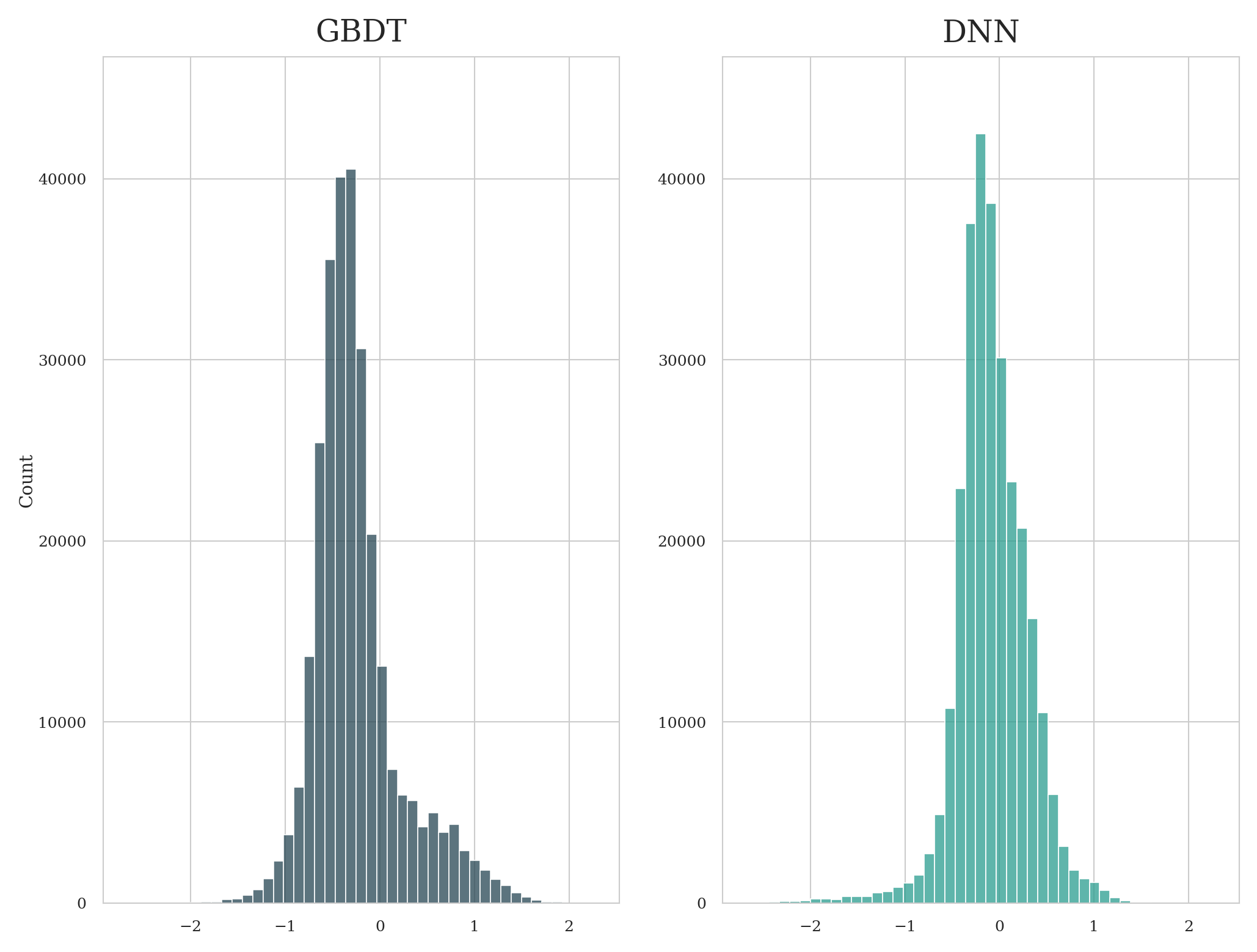}
        \caption{Nursing home last year (permanently)}
    \end{subfigure}
    \begin{subfigure}[b]{0.45\textwidth}
        \includegraphics[width=\linewidth]{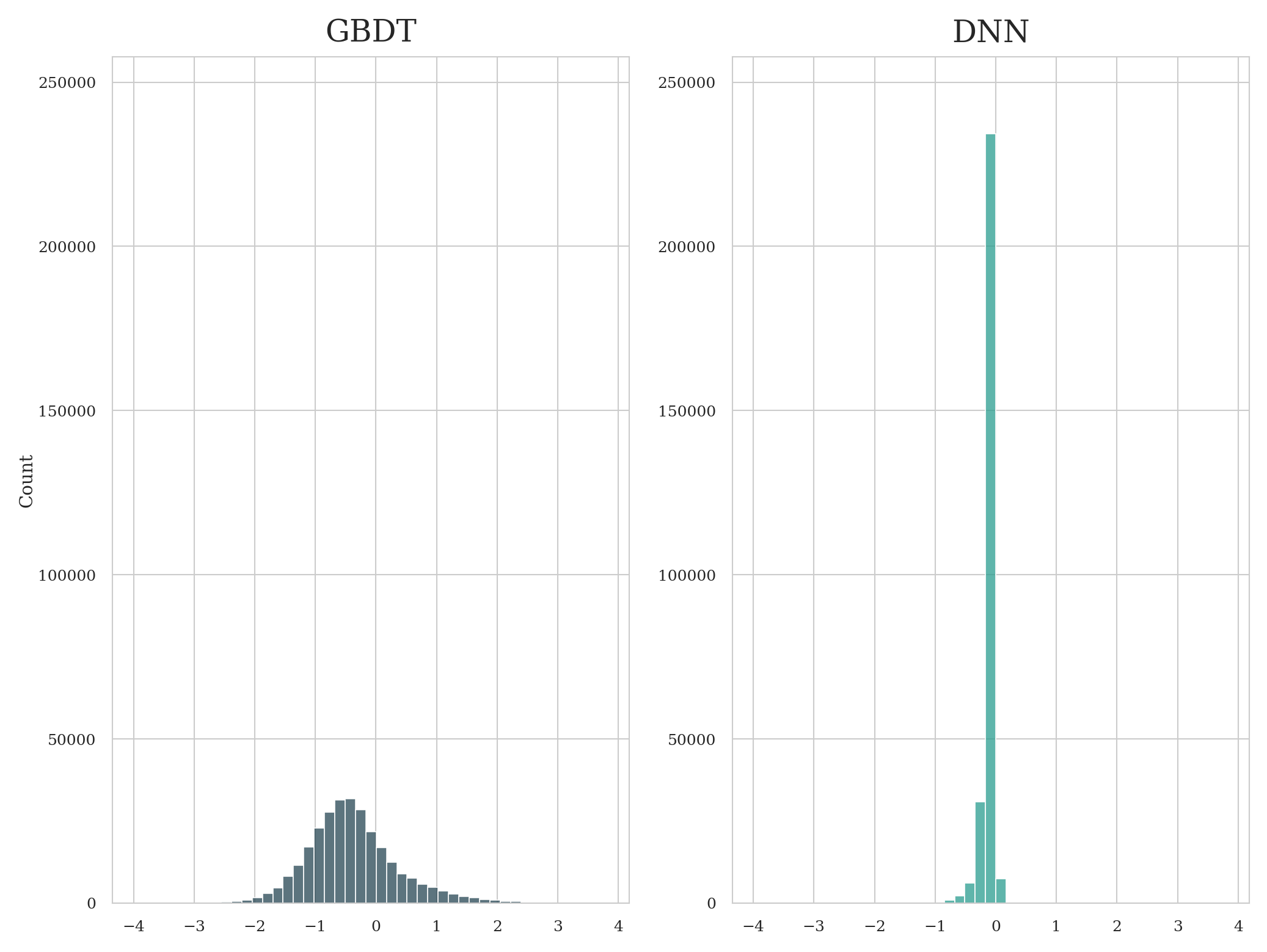}
        \caption{Nursing home last year (permanently) - models with functional intercept}
    \end{subfigure}

    \begin{subfigure}[b]{0.45\textwidth}
        \includegraphics[width=\linewidth]{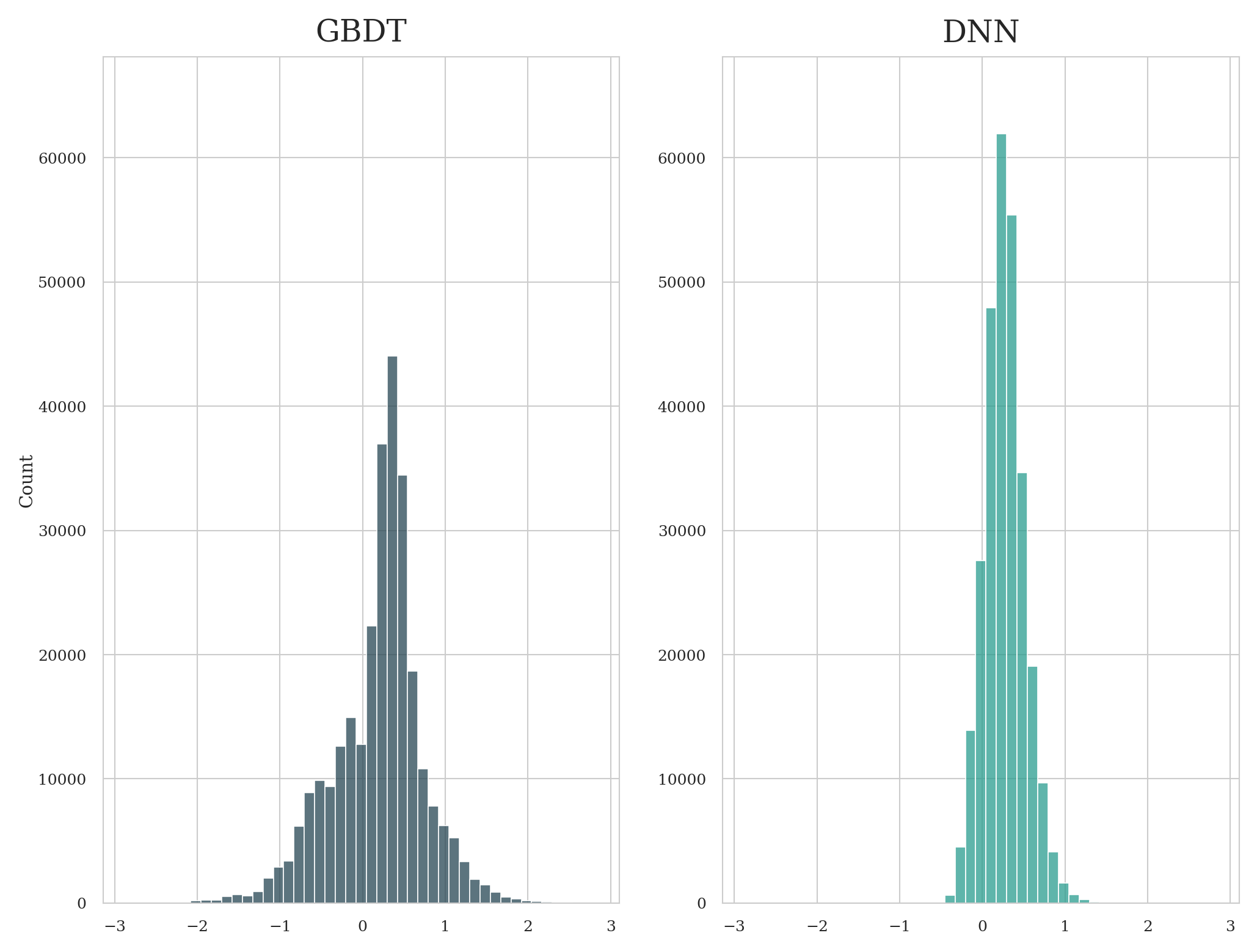}
        \caption{Nursing home last year (temporarily)}
    \end{subfigure}
    \begin{subfigure}[b]{0.45\textwidth}
        \includegraphics[width=\linewidth]{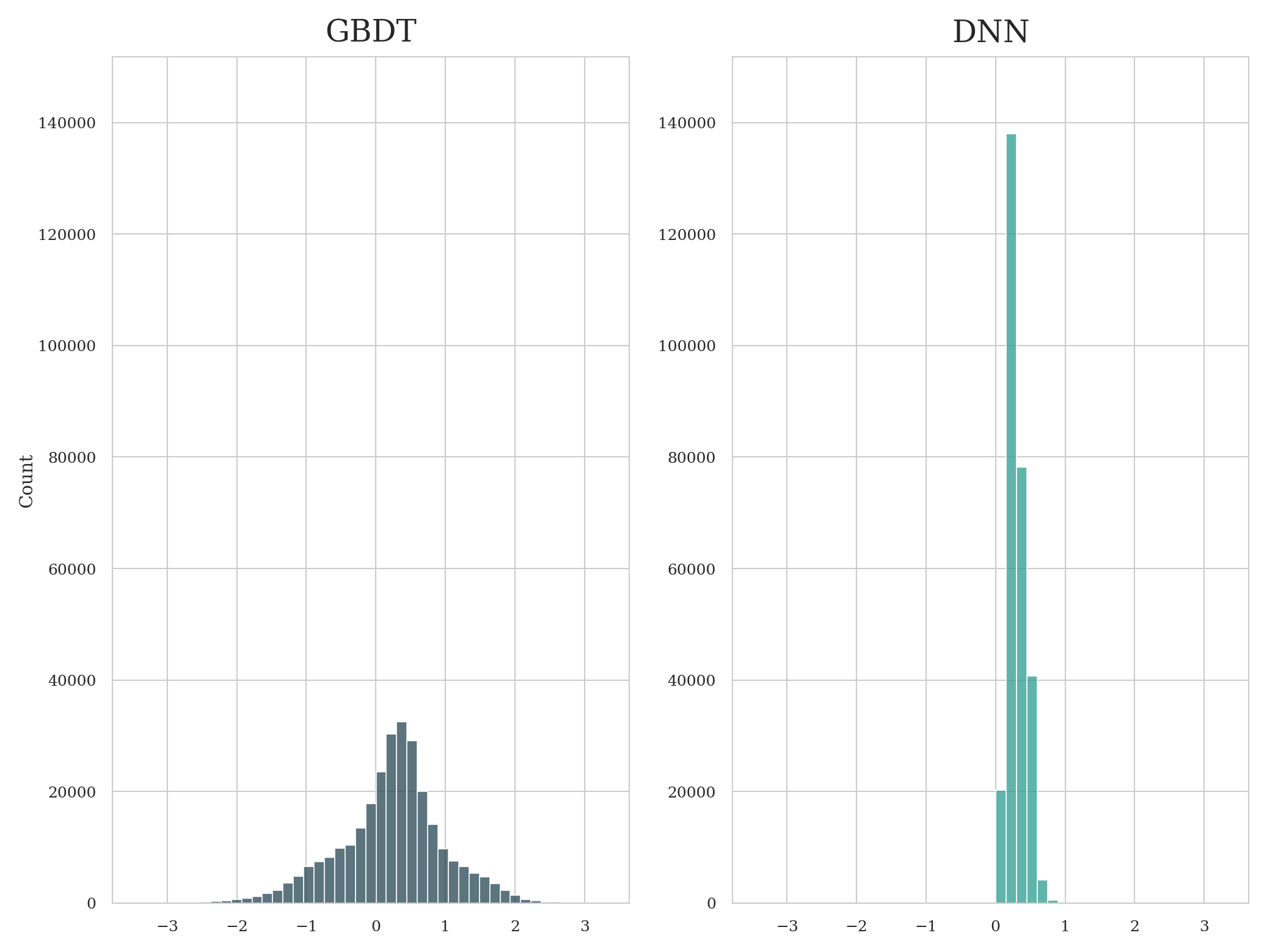}
        \caption{Nursing home last year (temporarily) - models with functional intercept}
    \end{subfigure}

\end{figure}

\begin{figure}[htbp]\ContinuedFloat

    \begin{subfigure}[b]{0.45\textwidth}
        \includegraphics[width=\linewidth]{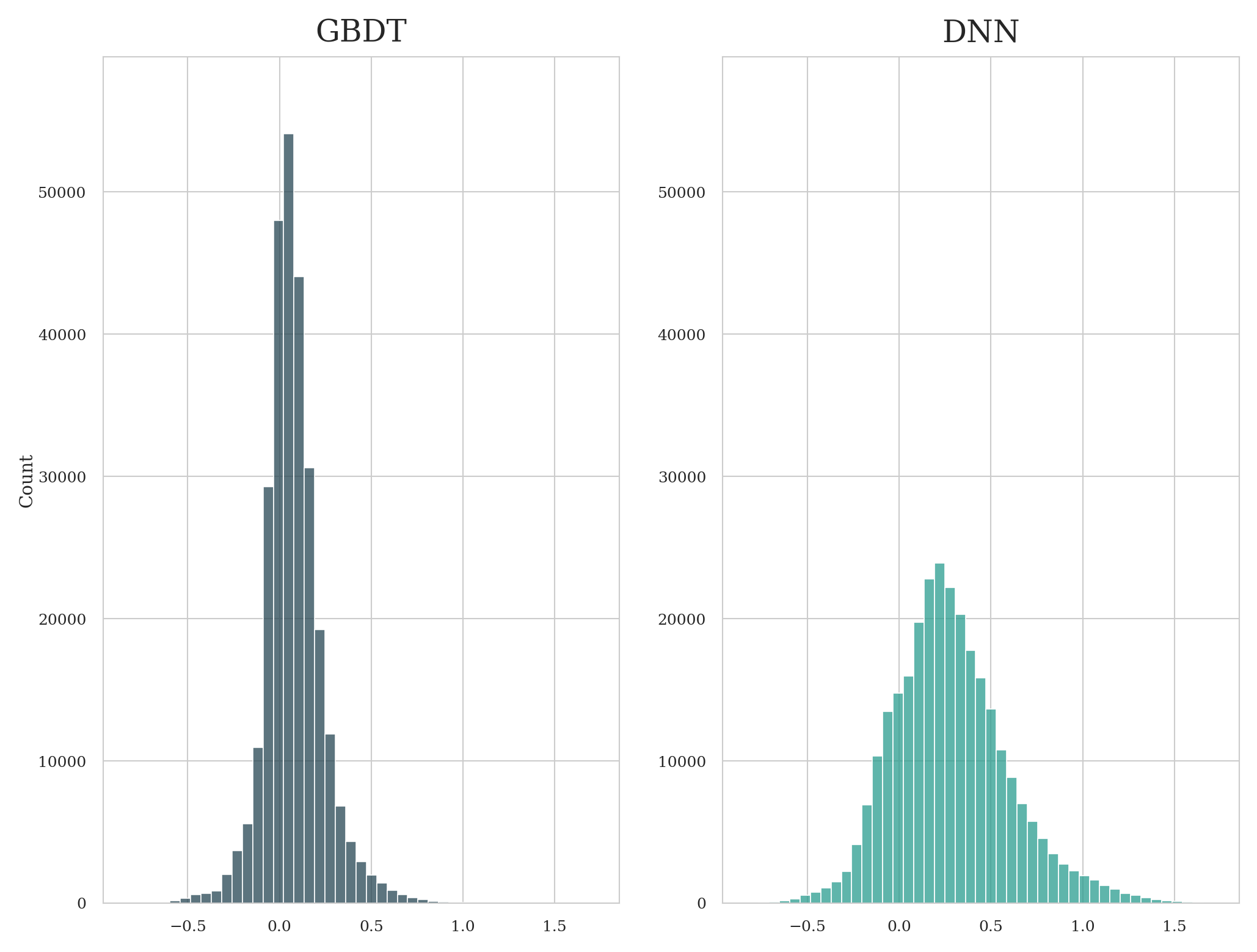}
        \caption{Self-perceived health - very good}
    \end{subfigure}
    \begin{subfigure}[b]{0.45\textwidth}
        \includegraphics[width=\linewidth]{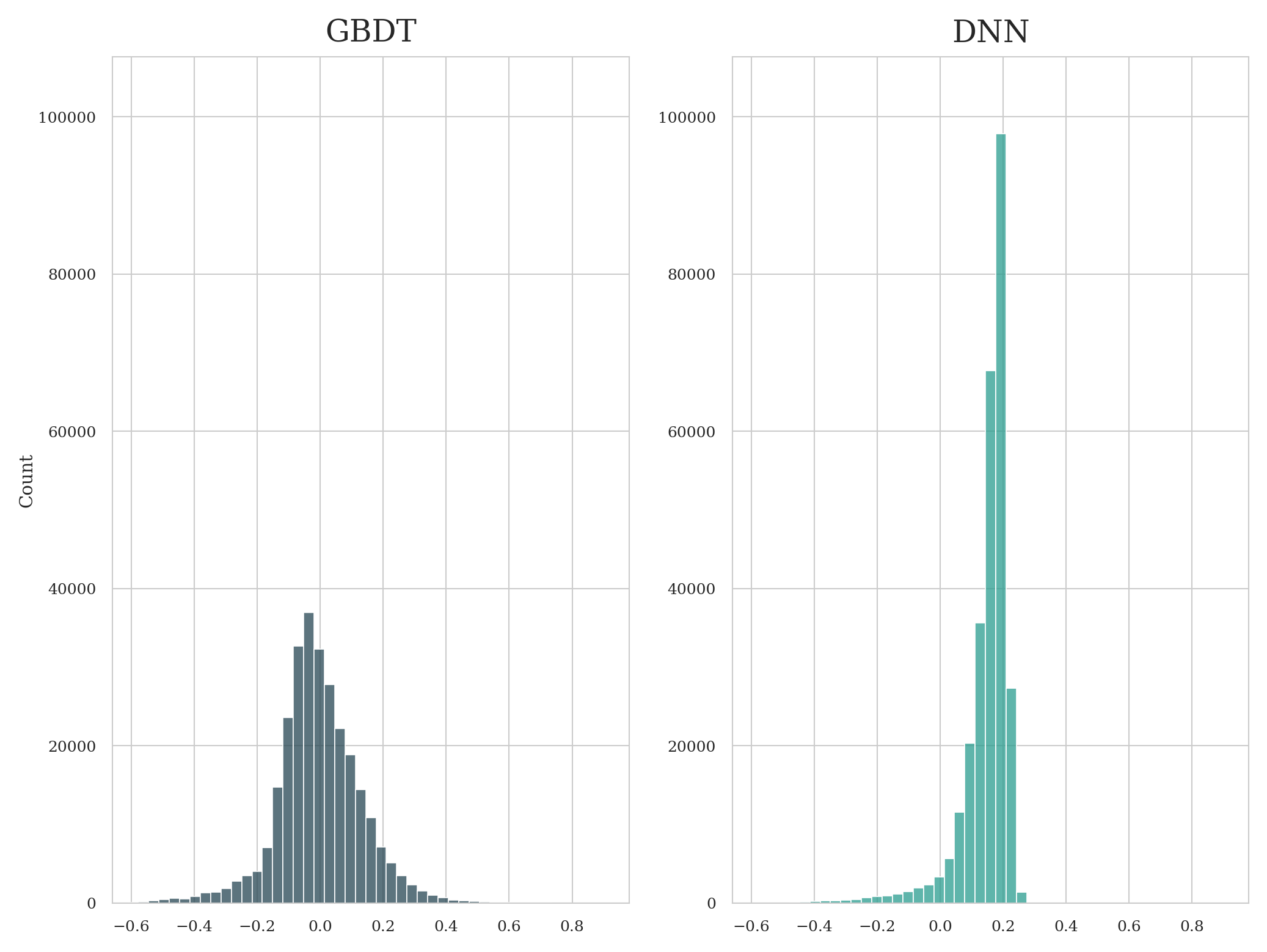}
        \caption{Self-perceived health - very good - models with functional intercept}
    \end{subfigure}

    \begin{subfigure}[b]{0.45\textwidth}
        \includegraphics[width=\linewidth]{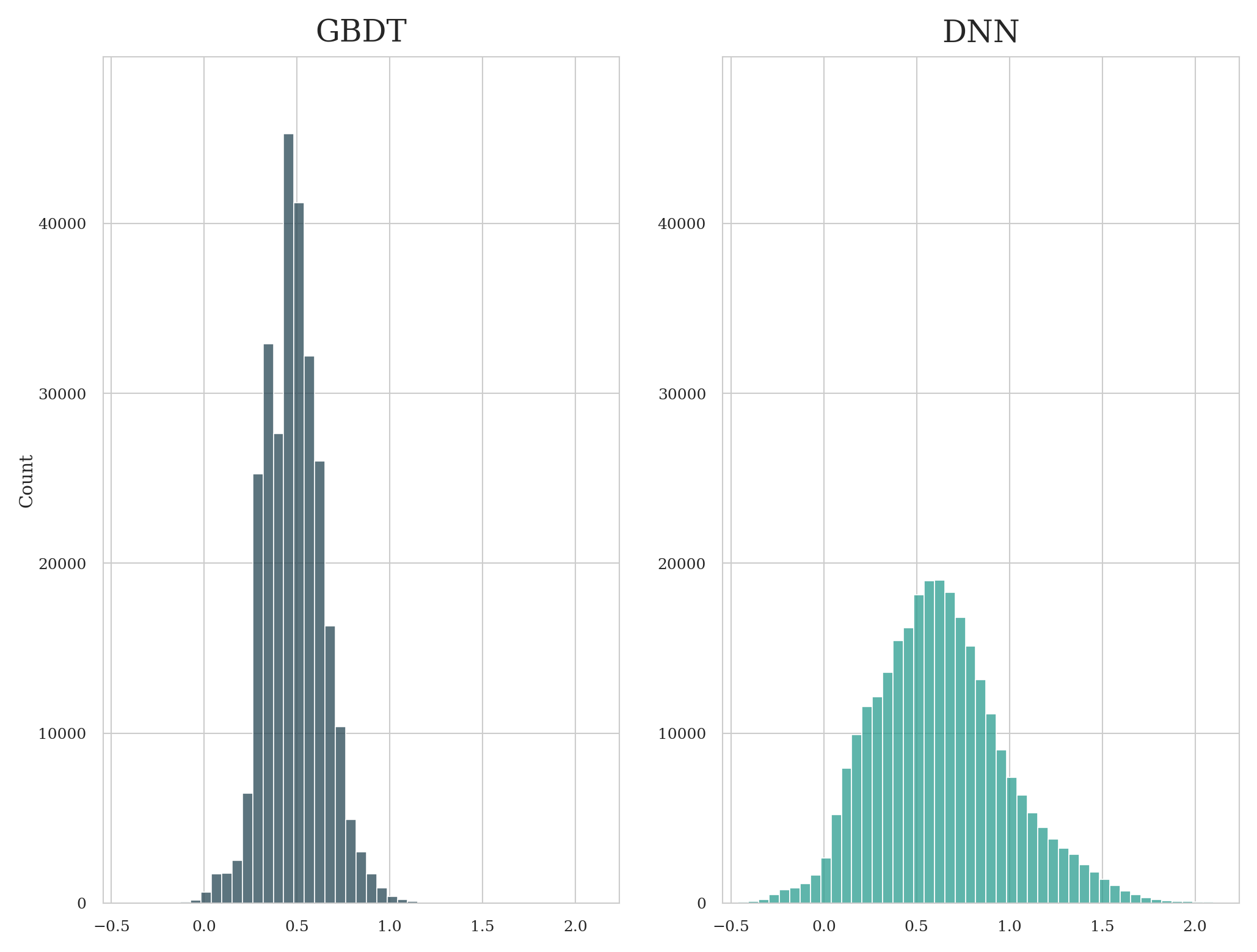}
        \caption{Self-perceived health - good}
    \end{subfigure}
    \begin{subfigure}[b]{0.45\textwidth}
        \includegraphics[width=\linewidth]{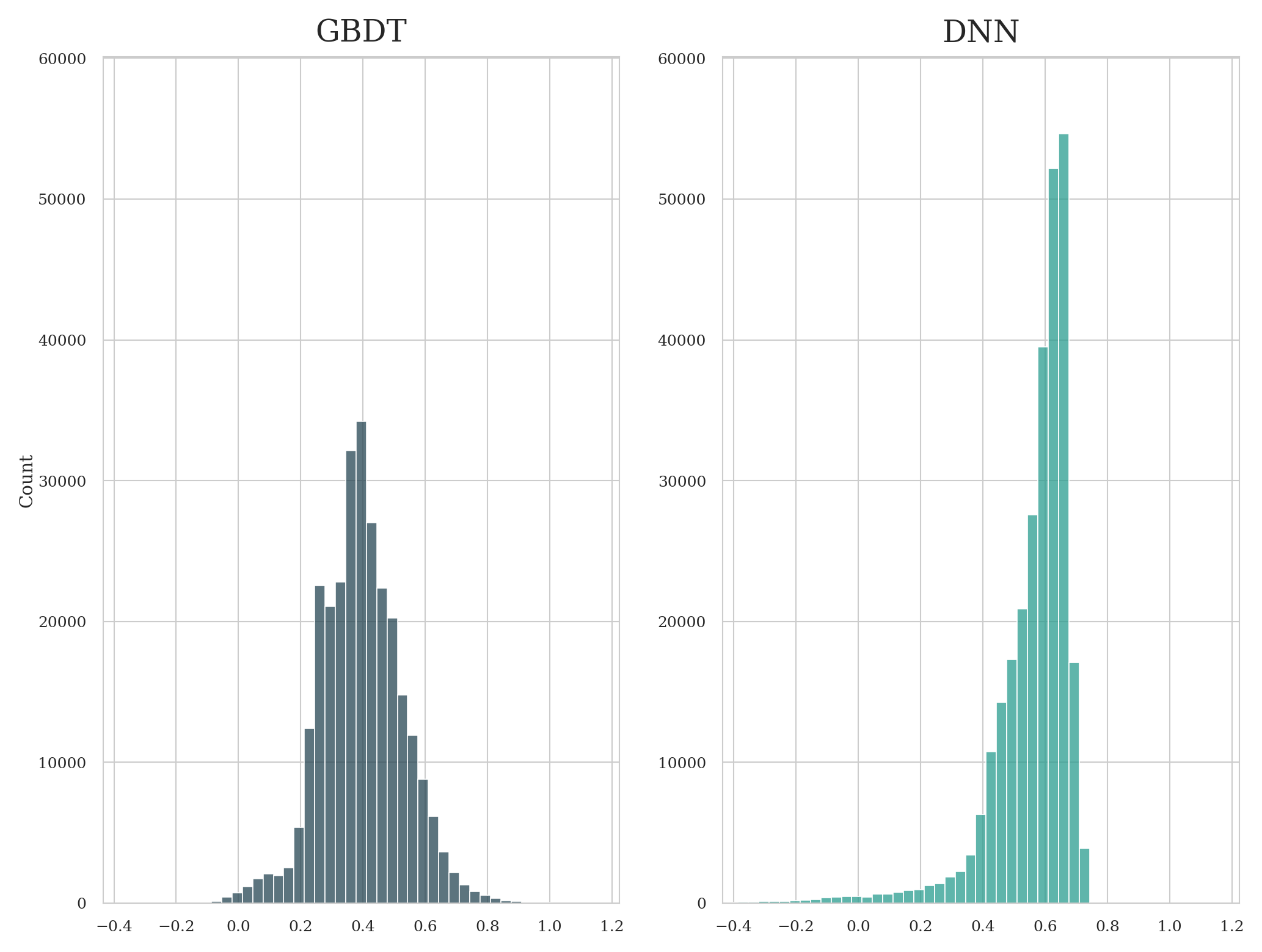}
        \caption{Self-perceived health - good - models with functional intercept}
    \end{subfigure}

    \begin{subfigure}[b]{0.45\textwidth}
        \includegraphics[width=\linewidth]{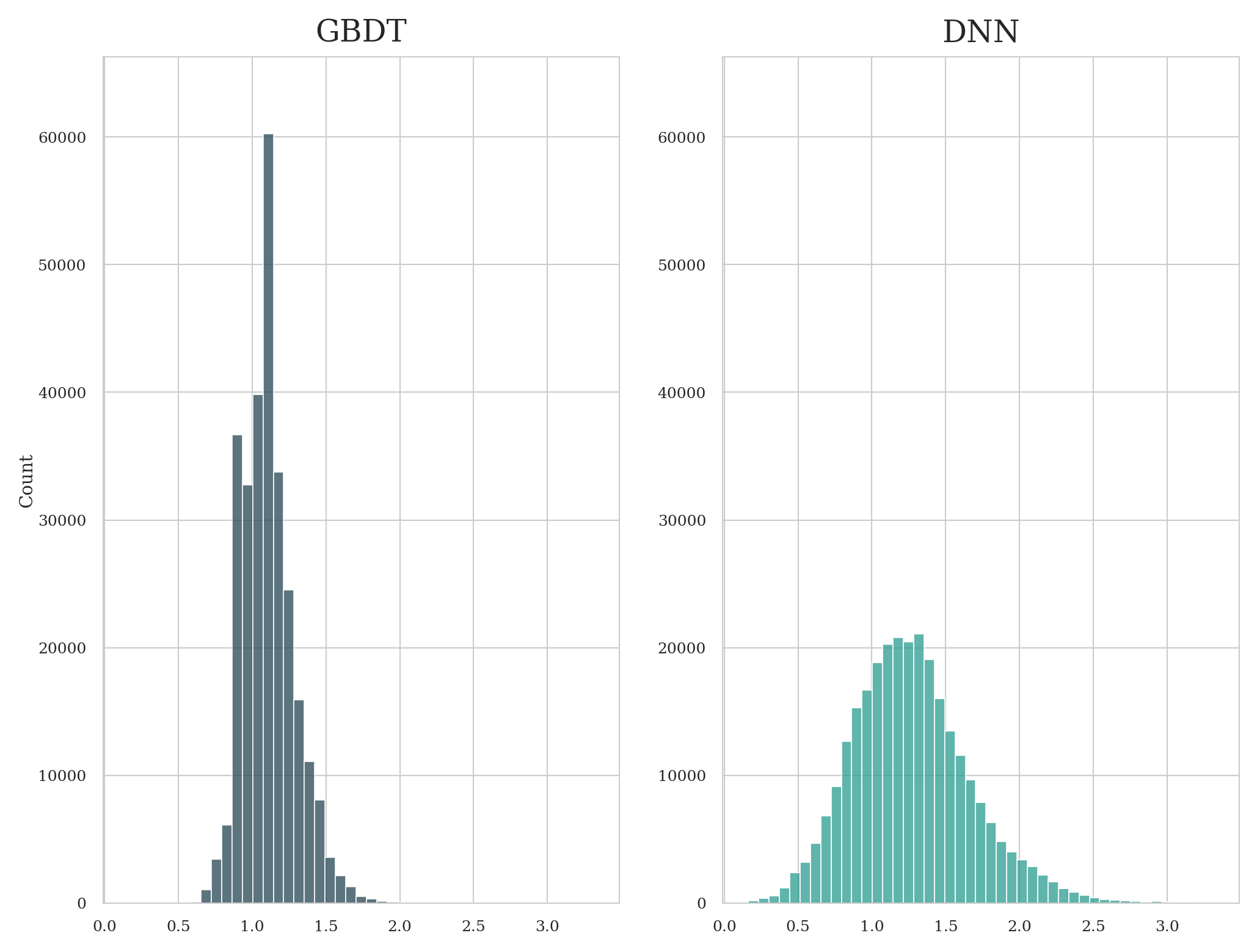}
        \caption{Self-perceived health - fair}
    \end{subfigure}
    \begin{subfigure}[b]{0.45\textwidth}
        \includegraphics[width=\linewidth]{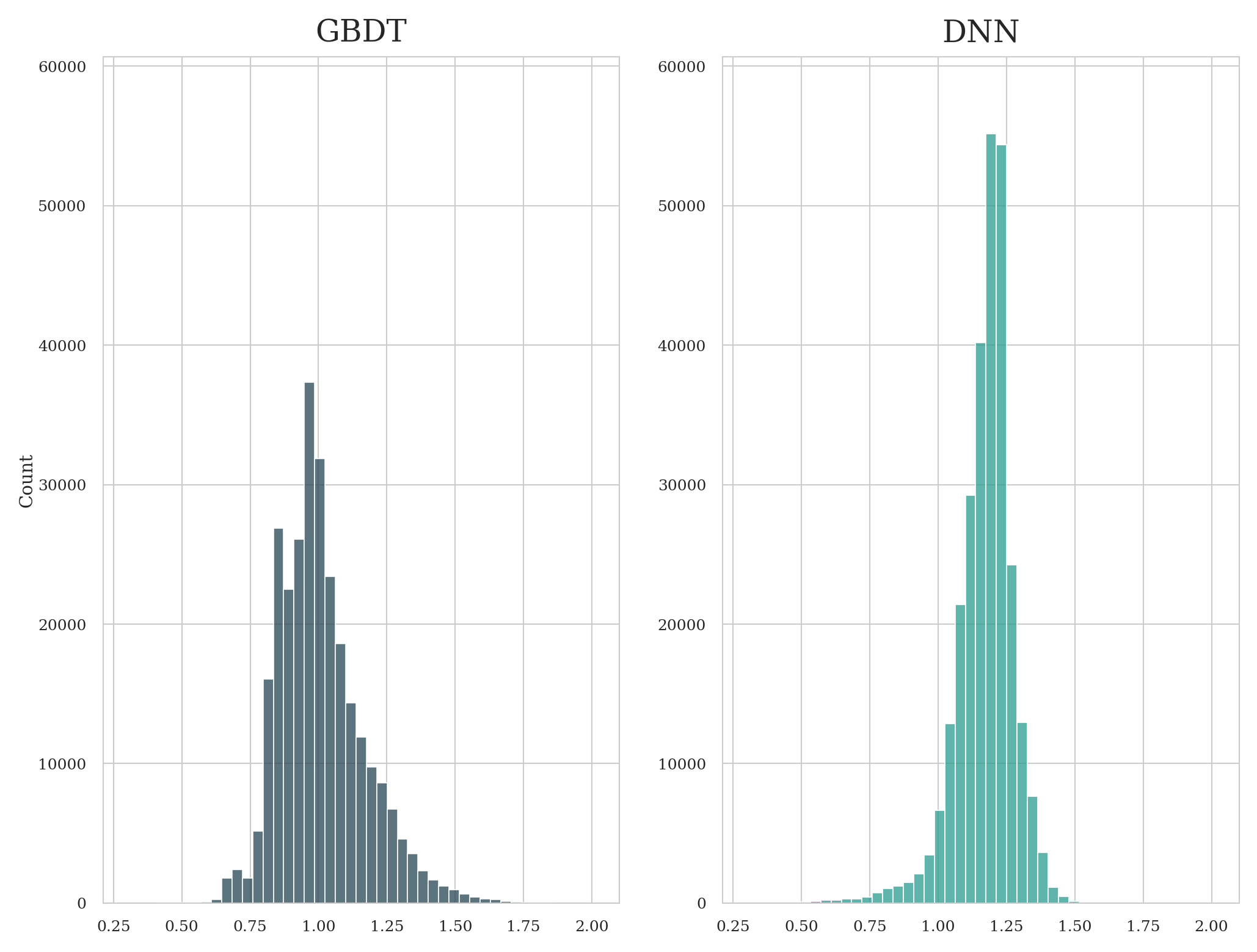}
        \caption{Self-perceived health - fair - models with functional intercept}
    \end{subfigure}

\end{figure}

\begin{figure}[htbp]\ContinuedFloat

    \begin{subfigure}[b]{0.45\textwidth}
        \includegraphics[width=\linewidth]{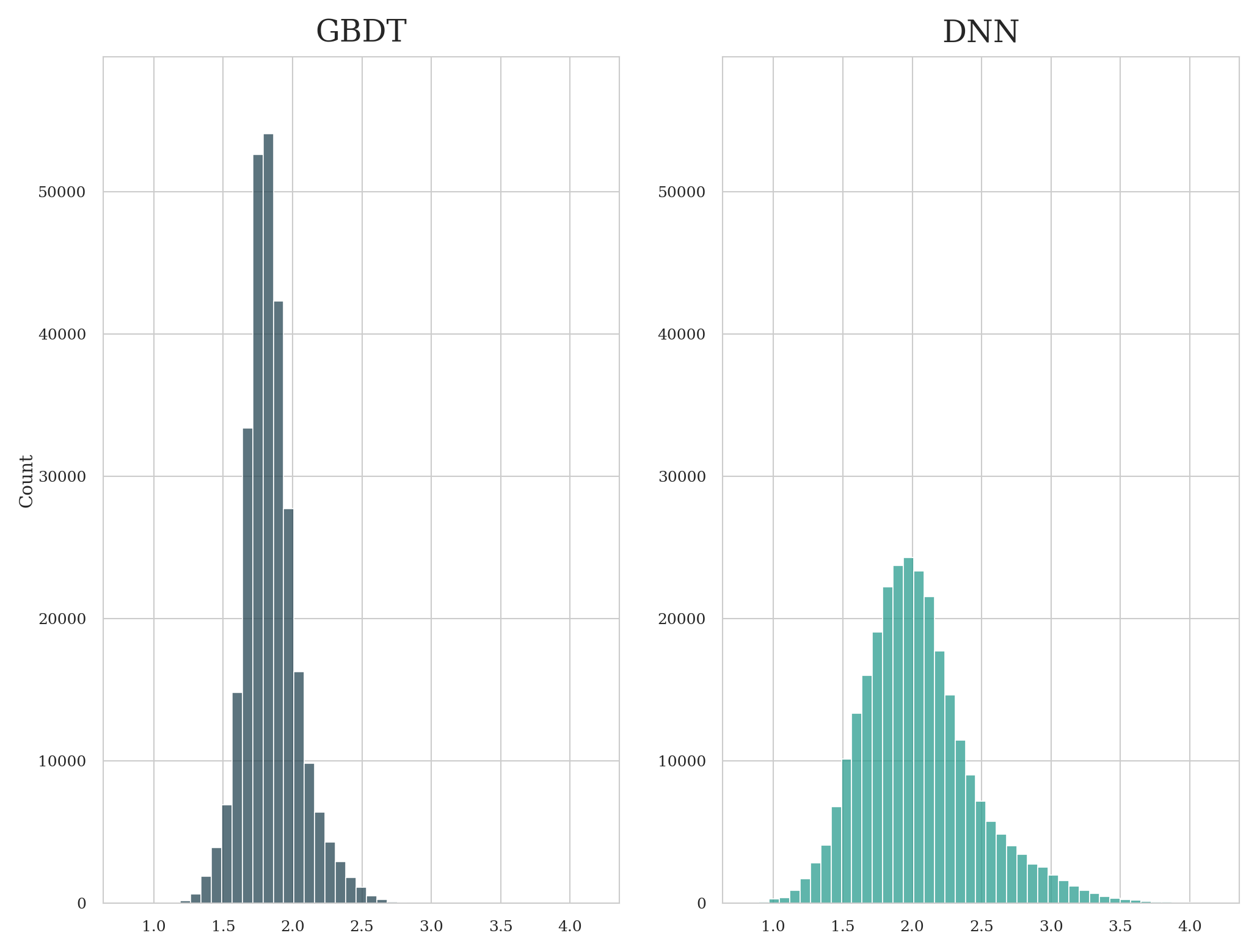}
        \caption{Self-perceived health - poor}
    \end{subfigure}
    \begin{subfigure}[b]{0.45\textwidth}
        \includegraphics[width=\linewidth]{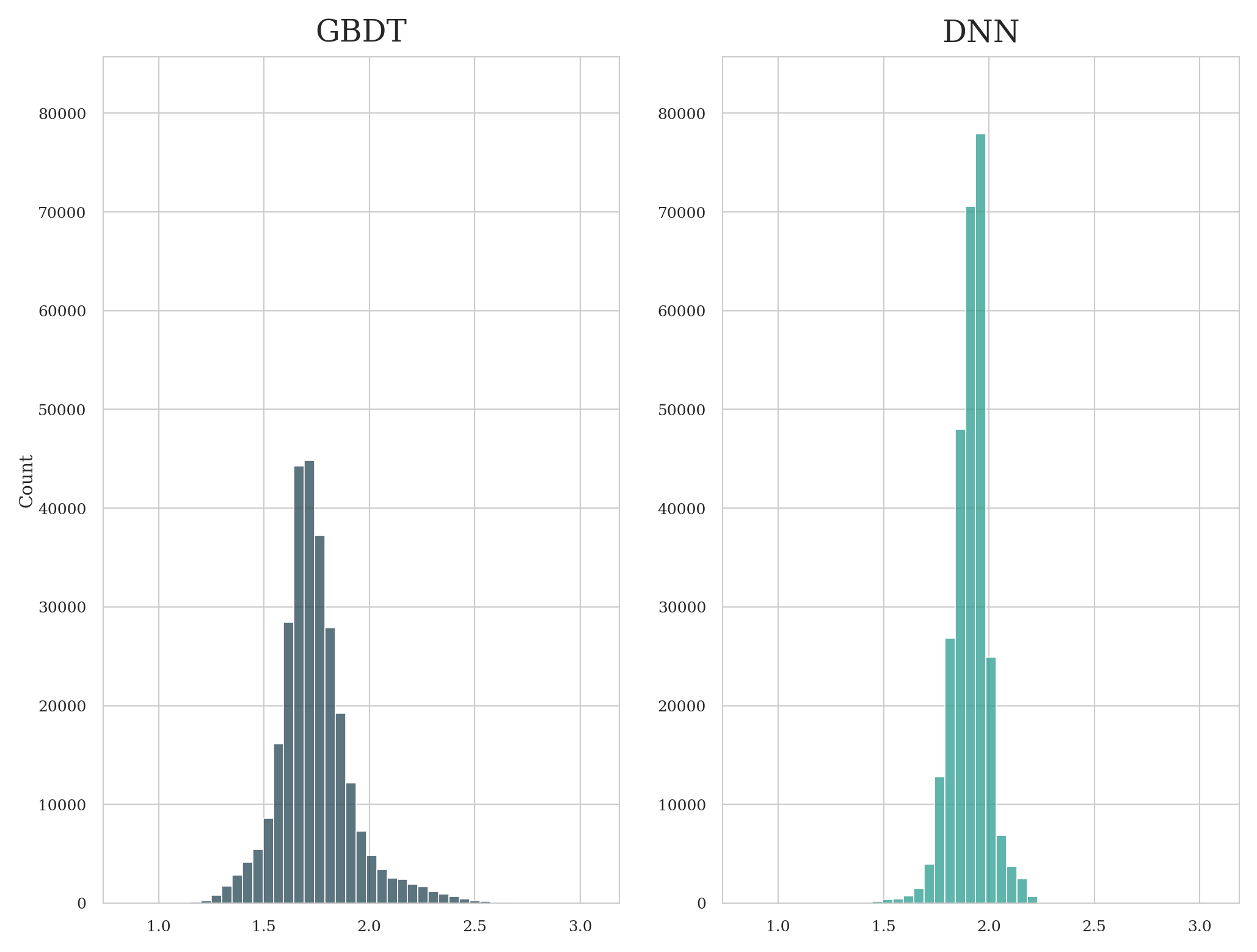}
        \caption{Self-perceived health - poor - models with functional intercept}
    \end{subfigure}

    \caption{Histograms of functional slopes for FS-GBDT and FS-DNN (left side) and FIS-GBDT and FIS-DNN (right side).}
    \label{fig:ind_spec_const_all}
\end{figure}

\setcounter{table}{0}
\setcounter{figure}{0}
\FloatBarrier
\section{Hyperparameter search}\label{app:hyperparameter search}

Table \ref{tab:search space} summarises the hyperparameter search space and hyperparameters tuned for GBDT and DNN.

\begin{table}[htbp]
\centering
\caption{Hyperparameter search space and applicability.}
\label{tab:search space}
\tiny
\begin{tabular}{llll}
\toprule
\textbf{Parameter} & \textbf{Search space} & \textbf{Distribution} & \textbf{Applies to} \\
\midrule
Best iteration / epoch & Max 3000 iterations / 200 epochs & -- & GBDT/DNN \\
lambda\_l1 & $[10^{-8}, 1]$ & Log-uniform & GBDT/DNN \\
lambda\_l2 & $[10^{-8}, 1]$ & Log-uniform & GBDT/DNN \\
num\_leaves & $[2, 256]$ & Discrete uniform & GBDT \\
feature\_fraction & $[0.4, 1]$ & Uniform & GBDT \\
bagging\_fraction & $[0.4, 1]$ & Uniform & GBDT \\
bagging\_freq & $[1, 7]$ & Discrete uniform & GBDT \\
min\_data\_in\_leaf & $[1, 200]$ & Discrete uniform & GBDT \\
max\_bin & $[64, 511]$ & Discrete uniform & GBDT \\
min\_sum\_hessian\_in\_leaf & $[10^{-8}, 10]$ & Log-uniform & GBDT \\
min\_gain\_to\_split & $[10^{-8}, 10]$ & Log-uniform & GBDT \\
batch\_size & \{256, 512\} & Categorical & DNN \\
learning\_rate & $[0.0001, 0.01]$ (fixed 0.05 or 0.1 for GBDT) & Log-uniform & GBDT/DNN \\
dropout & $[0.0, 0.9]$ & Uniform & DNN \\
act\_func & \{ReLU, Sigmoid, Tanh\} & Categorical & DNN \\
batch\_norm & \{True, False\} & Categorical & DNN \\
layer\_sizes & \{[32], [64], [128], [32,32], [64,64], [128,128], [64,128], [128,64], [64,128,64]\} & Categorical & DNN \\
\bottomrule
\end{tabular}
\end{table}

Table \ref{tab:synthetic hyp search} provides an overview of the hyperparameter search range and optimal values for all models for the synthetic dataset of Section \ref{synthetic}.

\begin{table}[htbp]
\centering
\caption{Hyperparameter search for the synthetic dataset of Section \ref{synthetic}.}
\label{tab:synthetic hyp search}
\tiny
\begin{tabular}{lrr}
\toprule
 & FI-RUMBoost & FI-DNN \\
\midrule
Val. MCEL & 1.347 & 1.346 \\
Time [s] & 232 & 2788 \\
Best it.\slash ep. & 340 & 14 \\
lambda\_l1 & 0.000 & 0.012898 \\
lambda\_l2 & 0.000 & 0.000004 \\
num\_leaves & 94 & - \\
feature\_fraction & 0.437 & - \\
bagging\_fraction & 0.491 & - \\
bagging\_freq & 1 & - \\
min\_data\_in\_leaf & 57 & - \\
max\_bin & 85 & - \\
min\_sum\_hessian\_in\_leaf & 0.108 & - \\
min\_gain\_to\_split & 6.118 & - \\
batch\_size & - & 512 \\
learning\_rate & 0.1* & 0.002289 \\
dropout & - & 0.193 \\
act\_func & - & sigmoid \\
batch\_norm & - & True \\
layer\_sizes & - & [128, 64] \\
\bottomrule
&  & *fixed
\end{tabular}
\end{table}

Table \ref{tab:swissmetro hyp search} provides an overview of the hyperparameter search range and optimal values for all models for the Swissmetro dataset of Section \ref{swissmetro}.

\begin{table}[htbp]
\centering
\caption{Hyperparameter search for the Swissmetro dataset of Section \ref{swissmetro}. OL means Ordinal Logit.}
\label{tab:swissmetro hyp search}
\tiny
\begin{tabular}{lrrrrrrrrrr}
\toprule
 & FIS-GBDT & FS-GBDT & FI-RUMBoost & RUMBoost &  FIS-DNN & FS-DNN & FI-DNN & OL & GBDT & DNN \\
\midrule
Val. MCEL & 0.639 & 0.695 & 0.633 & 0.768 & 0.607 & 0.648 & 0.676 & 0.835 & 0.610 & 0.630 \\
Time [s] & 807 & 770 & 219 & 129 & 684 & 523 & 852 & 1093 & 1037 & 344\\
Best it.\slash ep. & 797 & 785 & 399 & 960 & 58 & 37 & 130 & 162 & 3000 & 38 \\
lambda\_l1 & 0.000 & 0.000 & 0.053 & 0.000 & 0.002 & 0.000 & 0.000 & 0.000 & 0.000 & 0.000 \\
lambda\_l2 & 0.000 & 0.003 & 0.000 & 0.000 & 0.005 & 0.000 & 0.000 & 0.000 & 0.000 & 0.000 \\
num\_leaves & 150 & 77 & 244 & 165 & - & - & - & - & 249 & - \\
feature\_fra. & 0.892 & 0.768 & 0.677 & 0.403 & - & - & - & - & 0.501 & - \\
bagging\_fra. & 1.000 & 0.879 & 0.443 & 0.999 & - & - & - & - & 0.478 & - \\
bagging\_fre. & 4 & 6 & 7 & 2 & - & - & - & - & 1 & - \\
min\_data. & 62 & 108 & 39 & 32 & - & - & - & - & 6 & - \\
max\_bin & 426 & 270 & 272 & 282 & - & - & - & - & 104 & - \\
min\_sum\_h. & 0.258 & 0.000 & 0.000 & 0.000 & - & - & - & - & 0.000 & - \\
min\_gain. & 0.003 & 0.002 & 0.000 & 0.001 & - & - & - & - & 0.652 & - \\
batch\_size & - & - & - & - & 256 & 512 & 256 & 512 & - & 512 \\
learning\_r. & 0.100* & 0.100* & 0.100* & 0.100* & 0.005 & 0.006 & 0.006 & 0.009 & 0.002 & 0.008 \\
dropout & - & - & - & - & 0.449 & 0.679 & 0.892 & 0.000 & - & 0.728 \\
act\_func & - & - & - & - & tanh & relu & sigmoid & - & - & tanh \\
batch\_norm & - & - & - & - & False & True & True & False & - & False \\
layer\_sizes & - & - & - & - & [64, 128] & [64, 128, 64] & [64, 128] & - & - & [128, 128] \\
\bottomrule
& & & & & & & & & & *fixed \\
\end{tabular}
\end{table}

Table \ref{tab:lpmc hyp search} provides an overview of the hyperparameter search range and optimal values for all models for the LPMC dataset of Section \ref{swissmetro}.

\begin{table}[htbp]
\centering
\caption{Hyperparameter search for the LPMC dataset of Section \ref{swissmetro}.}
\label{tab:lpmc hyp search}
\tiny
\begin{tabular}{lrrrrrrrrrr}
\toprule
 & FIS-GBDT & FS-GBDT & FI-RUMBoost & RUMBoost &  FIS-DNN & FS-DNN & FI-DNN & MNL & GBDT & DNN \\
\midrule
Val. MCEL & 0.668 & 0.677 & 0.645 & 0.809 & 0.657 & 0.670 & 0.664 & 0.835 & 0.795 & 0.794 \\
Time [s] & 1464 & 1169 & 264 & 296 & 5519 & 6027 & 5045 & 4718 & 2065 & 2508 \\
Best it.\slash ep. & 511 & 410 & 506 & 886 & 85 & 84 & 91 & 61 & 3000 & 66 \\
lambda\_l1 & 0.000 & 0.000 & 0.591 & 0.000 & 0.000 & 0.000 & 0.000 & 0.000 & 0.000 & 0.001 \\
lambda\_l2 & 0.001 & 0.000 & 0.076 & 0.000 & 0.001 & 0.068 & 0.000 & 0.000 & 0.011 & 0.273 \\
num\_leaves & 3 & 3 & 73 & 236 & - & - & - & - & 21 & - \\
feature\_fra. & 0.583 & 0.860 & 0.960 & 0.978 & - & - & - & - & 0.803 & - \\
bagging\_fra. & 0.759 & 0.720 & 0.592 & 0.634 & - & - & - & - & 0.436 & - \\
bagging\_fre. & 4 & 7 & 2 & 2 & - & - & - & - & 5 & - \\
min\_data. & 87 & 37 & 124 & 61 & - & - & - & - & 172 & - \\
max\_bin & 470 & 152 & 85 & 474 & - & - & - & - & 337 & - \\
min\_sum\_h. & 0.000 & 0.000 & 0.000 & 3.083 & - & - & - & - & 0.000 & - \\
min\_gain. & 0.000 & 0.870 & 2.373 & 0.001 & - & - & - & - & 0.000 & - \\
batch\_size & - & - & - & - & 256 & 256 & 256 & 256 & - & 256 \\
learning\_r. & 0.100* & 0.100* & 0.100* & 0.100* & 0.000 & 0.000 & 0.010 & 0.010 & 0.002 & 0.004 \\
dropout & - & - & - & - & 0.224 & 0.739 & 0.533 & 0.000 & - & 0.541 \\
act\_func & - & - & - & - & tanh & tanh & tanh & - & - & tanh \\
batch\_norm & - & - & - & - & True & True & False & False & - & False \\
layer\_sizes & - & - & - & - & [32, 32] & [64, 64] & [32] & - & - & [64, 128, 64] \\
\bottomrule
& & & & & & & & & & *fixed \\
\end{tabular}
\end{table}

Table \ref{tab:easyshare hyp search} provides an overview of the hyperparameter search for the case study of Section \ref{case study} 
while reporting all optimal values for all models.

\begin{table}[htbp]
    \centering
    \caption{Hyperparameter search for the case study of Section \ref{case study} with the easySHARE dataset.}
    \label{tab:easyshare hyp search}
    \tiny
\begin{tabular}{lrrrrrrrr}
\toprule
 & FIS-GBDT & FS-GBDT & FI-RUMBoost & RUMBoost &  FIS-DNN & FS-DNN & FI-DNN & MNL \\
\midrule
Val. MCEL & 0.253 & 0.253 & 0.253 & 0.261 & 0.253 & 0.253 & 0.254 & 0.262 \\
Time [s] & 4512 & 5683 & 7752 & 32617 & 9683 & 13214 & 12194 & 8795 \\
Best it.\slash ep. & 385 & 517 & 867 & 2259 & 25 & 29 & 20 & 30 \\
lambda\_l1 & 0.521 & 0.000 & 0.000 & 0.000 & 0.000 & 0.000 & 0.000 & 0.000 \\
lambda\_l2 & 0.000 & 0.000 & 0.001 & 0.000 & 0.000 & 0.000 & 0.036 & 0.000 \\
num\_leaves & 3 & 3 & 11 & 74 & - & - & - & - \\
feature\_fraction & 0.788 & 0.708 & 0.492 & 0.720 & - & - & - & - \\
bagging\_fraction & 0.998 & 0.998 & 0.879 & 0.863 & - & - & - & - \\
bagging\_freq & 2 & 1 & 1 & 3 & - & - & - & - \\
min\_data\_in\_leaf & 159 & 173 & 194 & 127 & - & - & - & - \\
max\_bin & 209 & 363 & 95 & 219 & - & - & - & - \\
min\_sum\_hessian\_in\_leaf & 0.000 & 0.098 & 2.234 & 0.000 & - & - & - & - \\
min\_gain\_to\_split & 0.000 & 3.631 & 5.169 & 0.000 & - & - & - & - \\
batch\_size & - & - & - & - & 256 & 256 & 256 & 256 \\
learning\_rate & 0.05* & 0.052* & 0.05* & 0.052* & 0.001 & 0.002 & 0.004 & 0.001 \\
dropout & - & - & - & - & 0.659 & 0.003 & 0.553 & 0 \\
act\_func & - & - & - & - & sigmoid & sigmoid & sigmoid & - \\
batch\_norm & - & - & - & - & True & False & False & False \\
layer\_sizes & - & - & - & - & [128, 64] & [32, 32] & [32, 32] & - \\
\bottomrule
& & & & & & & & *fixed
\end{tabular}
\end{table}

\FloatBarrier
\section{Ordinal threshold values}\label{app:thresholds}

Table \ref{thresholds table} summarises the ordinal threshold values for all models of Section \ref{case study}.

\begin{table}[htbp]
\centering
\caption{Ordinal threshold values for the case study of Section \ref{case study}.}
\label{thresholds table}
\tiny
\begin{tabular}{lrrrrrrrr}
\toprule
Thresholds & FIS-GBDT & FS-GBDT & FI-RUMBoost & RUMBoost & FIS-DNN & FS-DNN & FI-DNN & Ord. Logit \\
\midrule
1 & -1.019 & -1.003 & -1.865 & -1.585 & -1.438 & -1.265 & -1.190 & -1.509 \\
2 & 0.148 & 0.163 & -0.694 & -0.454 & -0.257 & -0.085 & -0.012 & -0.374 \\
3 & 1.033 & 1.052 & 0.203 & 0.614 & 0.614 & 0.803 & 0.874 & 0.458 \\
4 & 1.792 & 1.806 & 0.956 & 1.137 & 1.387 & 1.572 & 1.650 & 1.217 \\
5 & 2.528 & 2.541 & 1.688 & 1.846 & 2.131 & 2.314 & 2.396 & 1.934 \\
6 & 3.262 & 3.276 & 2.420 & 2.556 & 2.846 & 3.044 & 3.097 & 2.608 \\
7 & 3.999 & 4.013 & 3.149 & 3.268 & 3.556 & 3.764 & 3.811 & 3.300 \\
8 & 4.742 & 4.757 & 3.883 & 3.986 & 4.283 & 4.478 & 4.509 & 4.032 \\
9 & 5.560 & 5.571 & 4.682 & 4.775 & 5.084 & 5.290 & 5.288 & 4.822 \\
10 & 6.568 & 6.572 & 5.648 & 5.711 & 6.035 & 6.267 & 6.232 & 5.748 \\
11 & 7.755 & 7.789 & 6.900 & 6.978 & 7.260 & 7.510 & 7.463 & 6.975 \\
12 & 8.890 & 8.957 & 8.136 & 8.366 & 8.805 & 9.137 & 9.097 & 8.571 \\
\bottomrule
\end{tabular}
\end{table}

\bibliographystyle{elsarticle-harv} 
\bibliography{cas-refs}

\end{document}